\documentclass[10pt]{article}

\usepackage[textwidth=16.8cm, textheight=21.0cm, footnotesep=1.5em]{geometry}
\usepackage[utf8]{inputenc} 
\usepackage{amssymb,amsmath,mathtools,bbm}
\usepackage{float}
\usepackage[round]{natbib}
\usepackage[ruled,vlined]{algorithm2e}
\usepackage{booktabs}
\usepackage{siunitx}
\usepackage{caption}
\usepackage{hyperref}
\usepackage{cleveref}
\usepackage[parfill]{parskip} 
\usepackage{algcompatible}
\usepackage{enumitem}
\usepackage{xcolor}
\usepackage{graphicx}
\usepackage{placeins}
\usepackage{multirow}
\usepackage{rotating}
\usepackage[hang]{footmisc}

\newcommand{\pic}{\pi_{c}}
\newcommand{\bx}{\boldsymbol{x}}
\newcommand{\bxn}{\boldsymbol{x}_{n}}
\newcommand{\bxm}{\boldsymbol{x}_{m}}
\newcommand{\bxnT}{\boldsymbol{x}^{\top}_{n}}
\newcommand{\bTheta}{\boldsymbol{\Theta}}
\newcommand{\bmu}{\boldsymbol{\mu}}
\newcommand{\bmuc}{\boldsymbol{\mu}_{c}}
\newcommand{\bA}{\boldsymbol{\Lambda}}
\newcommand{\bAc}{\boldsymbol{\Lambda}_{c}}
\newcommand{\bAcT}{\boldsymbol{\Lambda}^{\top}_{c}}
\newcommand{\tbAc}{\tilde{\bA}_{c}}
\newcommand{\tbAcT}{\tilde{\bA}^{\top}_{c}}
\newcommand{\hbAc}{\hat{\bA}_{c}}
\newcommand{\hbAcT}{\hat{\bA}^{\top}_{c}}
\newcommand{\bD}{\boldsymbol{D}}
\newcommand{\bDc}{\boldsymbol{D}_{c}}

\newcommand{\bDci}{\boldsymbol{D}^{-1}_{c}}
\newcommand{\tbDci}{\tilde{\bD}^{-1}_{c}}
\newcommand{\bIdentity}{\boldsymbol{I}}
\newcommand{\bLc}{\boldsymbol{L}_{c}}
\newcommand{\bLci}{\boldsymbol{L}^{-1}_{c}}

\newcommand{\tbLci}{\tilde{\boldsymbol{L}}^{-1}_{c}}
\newcommand{\bUc}{\boldsymbol{U}_{c}}
\newcommand{\bUcT}{\boldsymbol{U}^{\top}_{c}}
\newcommand{\bVc}{\boldsymbol{V}_{c}}
\newcommand{\tbVc}{\tilde{\boldsymbol{V}}_{c}}

\newcommand{\bEc}{\boldsymbol{E}_{c}}

\newcommand{\bYc}{\boldsymbol{Y}_{c}}

\newcommand{\bz}{\boldsymbol{z}}

\newcommand{\hbz}{\hat{\bz}}
\newcommand{\hbzT}{\hat{\bz}^{\top}}

\newcommand{\bSigma}{\boldsymbol{\Sigma}}
\newcommand{\bSigmac}{\boldsymbol{\Sigma}_{c}}
\newcommand{\bv}{\boldsymbol{v}}
\newcommand{\bvT}{\boldsymbol{v}^{\top}}
\newcommand{\qnc}{q_{n}(c;\,\bTheta)}

\newcommand{\qnczt}{q_{n,c}(\bz;\,\tilde{\bTheta})}
\newcommand{\bKK}{\boldsymbol{\KK}}

\newcommand{\OO}{\mathcal{O}}
\newcommand{\RR}{\mathbb{R}}
\newcommand{\LL}{\mathcal{L}}

\newcommand{\SeSp}{\mathcal{S}}
\newcommand{\KK}{{\mathcal K}}
\newcommand{\II}{{\mathcal I}}
\newcommand{\FF}{{\mathcal F}}
\newcommand{\HH}{{\mathcal H}}
\newcommand{\ct}{\tilde{c}}
\newcommand{\con}{c_{n}}
\newcommand{\GGn}{\SeSp_{n}}
\newcommand{\GGc}{g_{c}}
\newcommand{\KKn}{\KK_{n}}
\newcommand{\KKnPrime}{\KK_{n'}}
\newcommand{\KKnopt}{\KK_{n}^{\mathrm{opt}}}
\newcommand{\tKKn}{\tilde{\KK}_{n}}

\newcommand{\bxEst}{\bx^{\mathrm{est}}}

\newcommand{\EE}[2]{\EEE_{#1}\big[#2\big]}
\newcommand{\EEE}{\mathbbm{E}}

\newcommand{\DKL}{D_{\mathrm{KL}}\left[p(\bx\,|\,c, \bTheta)\ ||\ p(\bx\,|\,\ct, \bTheta)\right]}
\newcommand{\rel}{D_{c\ct}}
\newcommand{\relt}{\tilde{D}_{c\ct}}

\newcommand{\qnct}{q_n\hspace{-1pt}(c;\,\tilde{\bTheta})\xspace}
\newcommand{\Ez}[1]{\mathbb{E}_{q_{n,c}}\left[#1\right]}
\newcommand{\dv}[2]{\frac{\partial #1}{\partial #2}}

\newcommand{\hbmuz}{\Ez{\hbz}}
\newcommand{\hbmuzT}{\Ez{\hbz}^{\top}}
\newcommand{\hEzz}{\Ez{\hbz \hbzT}}

\newcommand{\algBreak}{\\[1mm]}

\DeclareMathOperator*{\argmin}{argmin}
\DeclareMathOperator*{\argmax}{argmax}
\newcommand{\myargmax}[1]{ \underset{#1}{\argmax} }
\newcommand{\myargmin}[1]{ \underset{#1}{\argmin} }

\newcommand{\eqdef}{=\vcentcolon}
\newcommand{\defeq}{\vcentcolon=}

\newcommand{\myparagraph}[1]{\vspace{0em}\noindent{}\textbf{#1:}\hspace{0em}}

\newcommand{\varMFA}{v-MFA}
\newcommand{\fullMFA}{em-MFA}
\newcommand{\fullMFAminus}{\fullMFA$^{\dag}$}
\newcommand{\kmeans}{\emph{k}-means}
\newcommand{\kmeansfa}{\kmeans+\emph{FA}}
\newcommand{\torchmfa}{\textit{torch-mfa}}
\newcommand{\warmup}{\emph{warm-up}}

\newcommand{\algline}{{\color{gray}\hrulefill}}
\newcommand{\furl}[1]{\footnote{\url{#1}}}
\newcommand{\crefnobrackets}[1]{Eq. \ref{#1}}

\DeclareMathOperator{\trace}{tr}

\DeclareMathOperator{\var}{var}
\DeclareMathOperator{\diag}{diag}

\DeclareMathAlphabet{\mathsfit}{T1}{\sfdefault}{\mddefault}{\sldefault}

\definecolor{brightred}{rgb}{1.0,0.1,0.1}

\Crefname{appendix}{Appendix}{Appendices}
\crefname{appendix}{Appendix}{Appendices}
\Crefname{figure}{Fig.}{Figs.}
\crefname{figure}{Fig.}{Figs.}
\Crefname{section}{Sec.}{Secs.}
\crefname{section}{Sec.}{Secs.}
\Crefname{equation}{Eq.}{Eqs.}
\crefname{equation}{Eq.}{Eqs.}
\Crefname{table}{Tab.}{Tabs.}
\crefname{table}{Tab.}{Tabs.}
\Crefname{algorithm}{Alg.}{Algs.}
\crefname{algorithm}{Alg.}{Algs.}

\renewcommand{\COMMENT}[2][.15\linewidth]{
\leavevmode\hfill\makebox[#1][l]{~#2}}

\makeatletter
\newcommand{\algorithmfootnote}[2][\footnotesize]{
  \let\old@algocf@finish\@algocf@finish
  \def\@algocf@finish{\old@algocf@finish
    \leavevmode\rlap{\begin{minipage}{\linewidth}
    #1#2
    \end{minipage}}
  }
}
\makeatother

\begin{document}

\title{
  Sublinear Variational Optimization of Gaussian Mixture Models with Millions to Billions of Parameters \\[0.5em]
}

\author{Sebastian Salwig$^{1,2,\dagger}$, Till Kahlke$^{2,\dagger,*}$, Florian Hirschberger$^1$, Dennis Forster$^3$ and J\"org L\"ucke$^2$\\[0.5em]
  $^\dagger${\small These authors share first authorship on this work.
    }
}
\date{\raggedright {\small $^1$Machine Learning Lab, DMPA, Faculty VI\\
    \phantom{$^{1}$}Carl von Ossietzky University Oldenburg\\
    \phantom{$^{1}$}26129 Oldenburg, Germany\\[1em]
    $^2$Artificial Intelligence Lab, Department of Computer Science,\\
    \phantom{$^{2}$}Faculty of Mathematics, Computer Science and Physics\\
    \phantom{$^{2}$}Innsbruck Research Center for Artificial Intelligence\\
    \phantom{$^{2}$}University of Innsbruck\\
    \phantom{$^{2}$}6020 Innsbruck, Austria\\[1em]
    $^3$Data Analytics and AI, Faculty 3\\
    \phantom{$^{3}$}Frankfurt University of Applied Sciences\\
    \phantom{$^{3}$}60318 Frankfurt am Main, Germany \\}
}

\maketitle

\renewcommand*{\thefootnote}{\fnsymbol{footnote}}
\footnotetext[1]{Corresponding author \\[0.5em]  email addresses: \texttt{\{till.kahlke, joerg.luecke, sebastian.salwig\}@uibk.ac.at, florian.hirschberger@uol.de}, \\ \phantom{email addresses: \{}\texttt{dennis.forster@fb3.fra-uas.de} \\[0.5em] The source code is available at \url{https://github.com/variational-sublinear-clustering/vamm}.}
\renewcommand*{\thefootnote}{\arabic{footnote}}

\begin{abstract}%
  \noindent Gaussian Mixture Models (GMMs) range among the most frequently used models in machine learning.
  However, training large, general GMMs becomes computationally prohibitive for data sets that have many data points $N$ of high-dimensionality $D$.
  For GMMs with arbitrary covariances, we here derive a highly efficient variational approximation, which is then integrated with mixtures of factor analyzers (MFAs).
  For GMMs with $C$~components, our proposed algorithm substantially reduces runtime complexity from $\OO(NCD^2)$ per iteration to a complexity scaling linearly with $D$ and
  sublinearly with $NC$.
  In numerical experiments, we first validate that the complexity reduction results in a sublinear scaling for the entire GMM optimization process.
  Second, we show on large-scale benchmarks that the sublinear algorithm results in speed-ups of an order-of-magnitude compared to the state-of-the-art.
  Third, as a proof of concept, we finally train GMMs with over 10~billion parameters on about 100~million images, observing training times of less than nine hours on a single state-of-the-art~CPU.
  Finally, and fourth, we demonstrate the effectiveness of large-scale GMMs on the task of zero-shot image denoising, where sublinear training results in state-of-the-art denoising times while competitive denoising performance is maintained. \\[1em]
  \vfill{}
\end{abstract}

\section{Introduction}
\label{sec:Introduction}

In machine learning and data science, Gaussian mixture models (GMMs) are widely
used and well-established tools. They are a canonical approach to clustering \citep[e.g.,][]{McLachlan2000},
can provide valuable insight into data set structures \citep[e.g.,][]{Bishop2006}, or are used as an integral
part in conjunction with other approaches for a range of different tasks \citep[e.g.,][]{Zoran2011,TianEtAl2019,Bouguila2020,RobinScrucca2023}.

One reason for their widespread use is the ability of GMMs to flexibly approximate data densities
in potentially high-dimensional data spaces. Any task accomplished by a parametric data density model can, in principle, be addressed using a sufficiently large-scale and sufficiently optimized GMM. Such task generality is possible as GMMs are universal probability density estimators, i.e., they can approximate data densities arbitrarily well \citep[e.g.,][]{Parzen1962,EscobarWest1995,MazyaSchmidt1996,ZeeviMeir1997,LiBarron1999}.

In the density modeling context, flexible component parametrization and the number of components are crucial for approximation quality \citep[e.g.,][]{ZeeviMeir1997,LiBarron1999}.
However, if GMMs are applied to the large-scale data sets currently used in data science and machine learning, large numbers of components $C$ get combined with many data points $N$ of potentially high dimensions $D$.
In such settings, optimizing general GMMs quickly becomes computationally infeasible.
For instance, the runtime cost for executing a single iteration of conventional expectation maximization \citep[EM;][]{DempsterEtAl1977} scales with $\OO(NCD^2)$ for general GMMs, making optimization of large-scale models very time-consuming or impractical.
To address the limited scalability of conventional GMM optimization, several research directions have been pursued.
Each research line discussed in the following aims at reducing the computational complexity.

A common approach is to constrain the covariances by using diagonal covariance matrices \citep[][]{HirschbergerEtAl2022,ExarchakisEtAl2022}, which do not model correlations within components.
Diagonal covariance matrices reduce the cost of one EM iteration from $\mathcal{O}(NCD^2)$ to $\mathcal{O}(NCD)$.
However, such constraints make GMMs much less flexible, potentially impacting their ability to efficiently approximate data densities.
Another contribution \citep[][]{AsheriEtAl2021} also constrains covariances but with the main focus on allowing more components $C$ for the same amount of data.
This work does not focus on improving scalability compared to conventional GMM optimization.
Further approaches (that do address the quadratic scaling with $D$) include
geometrically-oriented approaches \citep[][]{Elkan2003,ChengEtAl1984,BeiGray1985},
random projections \citep[][]{ChanLeung2017}, or dimensionality reduction approaches \citep[][]{ BouveyronEtAl2007,Richardson2018,HertrichEtAl2021,Liu2022}.

Still another line of research to reduce the optimization cost of GMMs aims at reducing the number of data points.
Such reduced data sets are known as coresets \citep[e.g.,][]{LucicEtAl2018,HarPeledMazumdar2004,FeldmanEtAl2011} and replace the set of $N$ original data points by a smaller weighted set of $N'$ data points.
Using the weights from a corresponding coreset algorithm, computational efforts only
scale with $N'$.
Coresets have been used for GMMs with diagonal covariances \citep[][]{HirschbergerEtAl2022,ExarchakisEtAl2022}.
However, with more general covariances, a reduction to fewer data points is often undesirable, because in high dimensions sufficiently many data points per component are required
to appropriately estimate correlations.
Other methods that do not directly reduce the number of data points but aim at reducing the computational cost in $N$ include mini-batching \cite[e.g.,][]{Nguyen2020,Sculley2010}, training on separate subsets of the data set \cite[e.g.,][]{Liu2024}, or by hierarchical training schedules \citep[e.g.,][]{Richardson2018}.

\newpage

In contrast to approaches mentioned above, we here aim at decisively reducing computational complexity using variational techniques, while maintaining as flexible as possible GMMs.
The main contributions of this work can be summarized as follows:
\begin{enumerate}[label= (\Alph*)]
  \item We derive a truncated variational optimization method \citep[cf.][]{LuckeEggert2010,DrefsEtAl2022} applicable to GMMs with arbitrary covariance matrices.
  \item We introduce a highly efficient learning algorithm based on mixtures of factor analyzers \citep[MFAs, cf.][]{Ghahramani1996,McLachlanEtAl2003,Richardson2018}, enabling the application of GMMs with billions of parameters to very large-scale data sets.
\end{enumerate}

Additionally, we make use of advanced seeding approaches \citep[e.g.,][]{ArthurVassilvitskii2007, BachemEtAl2016a,BachemEtAl2016b,Fraenti2019}.
Seeding techniques improve optimization by selecting well-suited initial component centers.
But contributions (A) and (B), which jointly define the actual parameter optimization procedure (after initialization), will be the main focus of this work.
Our approach enables training of GMMs at scales that were previously considered computationally infeasible.

Regarding research contribution (A), variational optimization is generally used to reduce optimization complexity \citep[][]{NealHinton1998,JordanEtAl1999}.
Among the most common variational approaches are factored distributions (a.k.a.\ mean-field distributions) as families of variational distributions \citep[][and many more]{JordanEtAl1999,BleiJordan2006}, and Gaussians \citep[][]{OpperArchambeau2009,KingmaWelling2014}. Neither of these variational families is suitable for standard mixture models.\footnote{Only in the context of fully Bayesian approaches, factorization (of parameter distributions) has been used \citep[e.g.,][]{NasiosBors2006}.} But variational approaches have also repeatedly been applied to
standard Gaussian mixtures that are the focus of this work \citep[e.g.,][]{NealHinton1998,SheltonEtAl2017,ForsterEtAl2018}.
In the case of GMMs with diagonal covariances, it has recently been shown that variational optimization can reduce the complexity of one EM iteration from a linear scaling with $C$ to a scaling which is constant w.r.t.~$C$ \citep[][]{HirschbergerEtAl2022,ExarchakisEtAl2022}.
Such complexity reduction has enabled optimization of the, so far, largest scale GMMs with
up to \num{50000} components and millions of parameters.
However, existing procedures for reducing computational complexity have only been applied to GMMs without correlations within clusters \citep[][]{HirschbergerEtAl2022,ExarchakisEtAl2022}.
The previous derivation of approximate optimization rested on the assumption of Euclidean distances in data space and between component centers.
While this assumption can be motivated if the data points within a component {\em are} uncorrelated, it does not hold for the arbitrary covariances in general GMMs.
To address this limitation, we here derive a truncated variational optimization techniques that is directly applicable to GMMs with arbitrary covariance matrices.

Regarding research contribution (B), MFAs flexibly model correlations per component along lower-dimensional hyperplanes.
Each component aligns with a hyperplane of $H \leq D$ dimensions, which can be of arbitrary orientation, and which can be
different from component to component.
As a result, the complexity of an EM step is reduced from $\OO(NCD^2)$ to $\OO(NCDH)$.
For $H=D$, general GMMs are recovered.\footnote{The GMM is overparameterized in this case.}
For many types of data, however, $H$ can be much smaller than $D$.
Due to their reduced complexity in $D$, MFAs are applicable to high-dimensional data \citep[e.g.,][]{KockEtAl2022}, where they can accomplish tasks usually reserved for neural network approaches \citep[cf.][]{Richardson2018}.
However, with current optimization techniques, MFAs still scale approximately linear with $NC$, which remains their computational bottleneck.

By simultaneously reducing how the optimization complexity depends on $NC$ and how it depends on $D$, we here enable the optimization of as flexible as possible GMMs at as large scales as possible.
The resulting algorithm, that realizes a variational optimization for the flexibly parameterized MFAs, will be referred to as \varMFA{}.

The training principles and the derivation of \varMFA{} are described in \cref{sec:material_and_methods}, and its numerical evaluation is described in \cref{sec:numerical_experiments}.

\section{Methods}
\label{sec:material_and_methods}
We will first introduce the class of GMMs we consider, i.e., MFAs, and then derive a variational optimization for MFAs in \cref{sec:varEM,sec:VariationalEStep,sec:construction_Gn}.
Finally, the implementation of the complete, variational EM algorithm is described in \cref{sec:algorithmic_realization}.

\subsection{Variational Optimization of MFAs}
\label{sec:varEM}

In the Mixture of Factor Analyzers (MFA) model \citep[e.g.,][]{Ghahramani1996, Richardson2018}, a data point $\bx \in \RR^{D}$ is modeled by a hidden mixture
component $c \in \{1, \dots, C\}$ and a hidden factor $\bz \in \RR^{H}$. Each component with mixing proportion~$\pic$ (satisfying $\sum_c \pic = 1$) and mean $\bmuc \in \RR^{D}$ represents a factor analyzer \citep[cf.][]{McLachlan2000,McLachlanEtAl2003} with a factor
loading matrix $\bAc \in \RR^{D\times{}H}$.
The generative model is given by
\begin{align}
  p(c\,|\,\bTheta)         = \pic, \qquad
  p(\bz\,|\,\bTheta)       = \mathcal{N}(\bz; \boldsymbol{0}, \bIdentity), \qquad
  p(\bx\,|\,c, \bz,\bTheta) = \mathcal{N}(\bx; \bAc\bz+\bmuc, \bDc)
  \label{eq:mfa},
\end{align}
where $\mathcal{N}$ represents a multivariate Gaussian distribution, $\bDc \defeq \diag(\sigma_{c,1}^{2}, ..., \sigma_{c,D}^{2}) \in
  \RR^{D\times{}D}$ is a diagonal matrix containing independent noise,
and $\bTheta \defeq \{\pi_{1:C}, \bA_{1:C}, \bmu_{1:C}, \bD_{1:C} \}$ denotes all model
parameters.

The MFA model can also be considered as a GMM with low-rank covariance matrices
$\bSigmac = \bAc\bAcT+\bDc$. In this case, the generative model is expressed as
\begin{align}
  p(c\,|\,\bTheta)     = \pic
  , \qquad
  p(\bx\,|\,c,\bTheta) = \mathcal{N}(\bx; \bmuc, \bAc\bAcT+\bDc).
  \label{eq:mfa_as_gmm}
\end{align}
The reformulation results from marginalization over $\bz$ using Gaussian identities.

To fit the MFA model to a given data set $\bx_{1:N}$, we seek parameters
$\bTheta^{*}=\argmax_{\bTheta}\, \LL(\bTheta)$ that optimize the
log-likelihood given by
\begin{align}
  \mathcal{L}(\bx_{1:N}; \bTheta) & =
  \sum_{n=1}^{N}\log\Big( \sum_{c=1}^{C}\pic~\mathcal{N}(\bxn; \bmuc,
    \bAc\bAcT+\bDc) \Big).
  \label{eq:loglikelihood}
\end{align}
Direct log-likelihood optimization is usually difficult. Instead, efficient algorithms
often employ approaches such as Expectation Maximization (EM) or variational approximations
of EM \citep[][]{DempsterEtAl1977,NealHinton1998,JordanEtAl1999}.
Variational approaches optimize the free energy, which is a lower bound of the log-likelihood (also known as the evidence lower bound---ELBO).
The free energy is
iteratively optimized by computing expectation values of latent variables
in the E-step and updating the model parameters $\bTheta$ in the M-step.
In this study, we employ a variational version of the EM algorithm, which
uses truncated posterior distributions \citep[e.g.,][]{LuckeEggert2010,SheltonEtAl2017,DrefsEtAl2022} as its family of
variational distributions. Concretely, we use variational distributions $\qnct$ defined by
\begin{align}
  \forall n=1,...,N: \quad \qnct \defeq q(c; \bxn,\KKn,\tilde{\bTheta}) = \frac{p(c,\bxn \,|\,
    \tilde{\bTheta})}{\sum_{\tilde{c} \in \KKn} p(\tilde{c},\bxn \,|\,
    \tilde{\bTheta})}\,\delta(c\in\KKn),
  \label{eq:qnc}
\end{align}
where $\delta(c\in\KKn)$ is equal to 1 if $c\in\KKn$ and 0 otherwise (also known as Iverson bracket), and $\KKn$ denotes those component indices for which
the variational distribution $\qnct$ is non-zero.
Throughout this work, we will assume that the size of any set $\KKn$
is restricted to $C'\leq{}C$ indices, i.e., $|\KKn|=C'$ for all $n$.

If we use truncated distributions (\crefnobrackets{eq:qnc}), we obtain as variational free energy objective (see \cref{appendix:aintro} for details)
\begin{align}
  \FF(\bx_{1:N}; \bKK,\tilde{\bTheta},\bTheta) \! \defeq & \sum\limits_{n=1}^{\;N}\Big( \! \sum_{\crampedclap{\;\; c=1}}^C \qnct \log\big( \pic~\mathcal{N}(\bxn; \bmuc,
      \bAc\bAcT+\bDc) \big) + \HH\left[ q_n \right] \Big),
  \label{eq:FE}
\end{align}
where $\bKK$ is the collection of all index sets $\mathcal{K}_{1:N}$, and $\mathcal{H}\left[ q_n \right] = - \mathbb{E}_{q_n}[\log \qnct ]$ is the Shannon entropy. A detailed derivation is given in the appendix of \citet[][]{Lucke2019}.
The `hard' zeros used by the truncated distributions mean that we effectively only have to evaluate $C'$  non-zero summands per data point.
This reduces the computational complexity of one M-step from linearly scaling with $C$ to a linear scaling with $C'$.

The optimal values of the variational parameters $\tilde{\bTheta}$ are the current values of the model parameters $\bTheta$,
which can be shown in general \citep{Lucke2019} and, therefore, also applies for the MFA model.
For $\tilde{\bTheta}=\bTheta$, the free energy simplifies to (see \cref{appendix:aintro} for details)
\begin{align}
  \FF(\bKK,\bTheta) \defeq\FF(\bx_{1:N}; \bKK, \bTheta, \bTheta) = \sum\limits_{n=1}^{N} \log\!\Big(\sum_{\crampedclap{\;\; c\in\KKn}} p(c,\bxn\,|\,\bTheta) \Big).
  \label{eq:FEShort}
\end{align}
As could be deduced from \cref{eq:FEShort}, optimizing the MFA model using variational EM requires to repeatedly evaluate joint probabilities $p(c,\bxn \,|\, \bTheta)$ or, equivalently, log-joints.
In previous work \citep{HirschbergerEtAl2022,ExarchakisEtAl2022}, the use of GMMs with diagonal covariance matrices enabled the evaluation of a single log-joint in $\OO(D)$.
However, in this work, the goal is to model correlations, which substantially increases the computational complexity to $\OO(D^2)$ when all correlations are considered.
By modeling data distributions along lower-dimensional hyperplanes, the MFA model reduces the complexity to $\OO(DH)$ while preserving arbitrary correlations along its hyperplanes.
Further details on the computational techniques employed for the MFA model, as well as the derivations of the parameter updates, can be found in \cref{appendix:EfficientEvaluationMFAs,appendix:Mstep}, respectively. Further details on the relation of the variational objective \cref{eq:FE} to the log-likelihood objective \cref{eq:loglikelihood} can be found in \cref{appendix:error_bound}.
\subsection{Efficient Partial Variational E-steps}
\label{sec:VariationalEStep}

As with other variational approaches, the crucial challenge is to derive a computationally efficient E-step.
To do so for the MFA model, we follow here a strategy similar to previous work \citep{HirschbergerEtAl2022,ExarchakisEtAl2022}, which has derived efficient partial variational E-steps focusing on diagonal covariance matrices.

We start by noting that the joint probabilities defined by the MFA model play a central role in the optimization of the free energy. This role is underlined by the following proposition:

\myparagraph{Proposition 1}
Consider the joint probability $p(c,\bxn\,|\,\bTheta)$ defined by the MFA generative model in \cref{eq:mfa_as_gmm}, and the free energy $\FF$ for index sets $\bKK$.
If we replace
a component~\mbox{$c\in\KKn$} by a component $\ct\not\in\KKn$ (for an arbitrary $n$), then the free energy increases
if and only if \mbox{$p(\ct,\bxn\,|\,\bTheta)>p(c,\bxn\,|\,\bTheta)$}.

Proposition~1 is a special case of a general result for truncated variational distributions \citep{Lucke2019}.
It therefore applies to any mixture model and to MFA as a special case.
For completeness, we provide the proof for general mixture models in \cref{appendix:PropOne}.

Proposition~1 then leads to an equivalent definition of the optimal variational parameters~$\KKn$  as in previous work \citep{HirschbergerEtAl2022,ExarchakisEtAl2022}.  $\KKnopt$ is given by
\begin{align}
  \KKnopt & = \{c \mid p(c,\bxn \,|\, \bTheta) \text{ is among the $C'$ largest joints} \},
  \label{eq:Knopt}
\end{align}
i.e., the optimal sets $\KKnopt$ contain those $C'$ components with the highest joint probabilities (also see \citealp{SheltonEtAl2017, ForsterEtAl2018}).
However, finding the optimal components would require the evaluation of $C$ joints per data point, i.e., $NC$ joint evaluations in total.
Hence, finding the optimal $\KKn$ given by \cref{eq:Knopt} would require the same number of joint evaluations as a full E-step.
To also reduce the computational complexity of the E-step, we consequently seek a procedure
similar to \citet{HirschbergerEtAl2022} and \citet{ExarchakisEtAl2022}, that builds upon
partial variational E-steps, i.e., we seek an increase of the free energy instead of its maximization.
In virtue of Proposition~1, we can increase the free energy by identifying components with \textit{higher} joints rather than finding the $C'$ components with the \textit{highest} joints.
This can be accomplished efficiently by evaluating only a subset of all joint probabilities during each E-step, which reduces the computational load compared to a full E-step.
Concretely, similar to \citet{HirschbergerEtAl2022}, we introduce a set $\GGn$, referred to as the search space,\footnote{The search space is denoted as $\mathcal{G}_n$ in \citet{HirschbergerEtAl2022}.} that includes candidate components $\ct$, which may replace a $c\in\KKn$ to increase the free energy for each data point $\bxn$.
The size of $\GGn$ is upper-bounded, i.e., we demand for all $n$ that $|\GGn|\leq{}S$.
Here, $S$ will be chosen to be larger than $C'$ but significantly smaller than $C$ (i.e., $C'<S \ll C$). Instead of finding the \textit{optimal} sets $\KKnopt$, we
now partially optimize each $\KKn$ using the search space $\GGn$ of a given $\bxn$. Given $\GGn$, the updated $\KKn$ is defined by
\begin{align}
  \KKn = \{c \mid p(c,\bxn\,|\,\bTheta) \mbox{ is among the $C'$ largest joints for all } c \in \GGn\} .
  \label{eq:Kn}
\end{align}
The index set $\KKn$ of \cref{eq:Kn} can consequently be determined by evaluating all joints with $\bxn$ and all $c\in\GGn$ as arguments.

We will define each search space $\GGn$ to contain $\KKn$ as subset ($\KKn \subset \GGn$), which guarantees that the update of $\KKn$ according
to \cref{eq:Kn} never decreases the free energy $\FF(\bKK,\bTheta)$.
However, this condition on the search spaces $\GGn$ is not sufficient to warrant an efficient increase of the free energy.
For high efficiency, replacement candidates $\ct\in\GGn \setminus \KKn$ have to be likely to result in larger joints, i.e., it has to be sufficiently likely that $\ct\in\GGn$ with $\ct \notin \KKn$ and $c \in \KKn$ satisfy $p(\ct,\bxn\,|\,\bTheta) > p(c,\bxn\,|\,\bTheta)$.

A first trivial option to define $\GGn$ would be to use all $c \in \KKn$ as members of $\GGn$, and to then add component indices that are uniformly sampled from $\{1,\ldots,C\}$.
However, as $C$ increases, the probability of finding a new component $\ct$ with high joint $p(\ct,\bxn\,|\,\bTheta)$ becomes increasingly small.
Therefore, we seek an approach that suggests components $\ct$ that are more likely to increase the free energy than such a blind random search.

\subsection{Construction of the Search Space for Guided Search}
\label{sec:construction_Gn}

In order to allow for a more guided search, we will define the search spaces $\GGn$ to contain component indices likely to improve
the free energy while being still efficiently computable. We remain with a data point centric approach and use a bootstrapping strategy to
continuously improve the search spaces $\GGn$ together with the sets $\KKn$.
In line with previous research \citep{HirschbergerEtAl2022}, we define the search spaces $\GGn$ by making use of (data point independent) sets $g_c$, that contain replacement candidates for $c$.
However, in this work, the definition of these sets will be introduced in a novel manner to account for GMMs with arbitrary covariances.
The components $\ct\in\GGc$ can, for now, be thought of as components estimated to be `similar' to component $c$, but we will only precisely define the $g_c$ further below.
The intuition behind using $\GGc$ sets goes as follows: if a data point $\bx$ lies in the vicinity of a component $c$, and component $c$ is `similar' to another component $\ct$ then, by transitivity, $\bx$
is also likely to be in the vicinity of $\ct$.
Using the sets $g_c$ (and one component sampled uniformly at random), a search space $\GGn$ for a given data point is defined as the following union:
\begin{align}
  \GGn \defeq \bigcup_{c \in \KKn} \GGc\, \cup \{c'\} \quad \mathrm{with} \quad c' \sim \mathcal{U}\{1,C\}.
  \label{eq:GGnSet}
\end{align}
Note that without the random component, there is no theoretical guarantee that the optimization process can find the optimal variational parameters $\KKnopt$ according to \cref{eq:Knopt} even in the limit of infinitely many iterations.

To ensure that the size of the search space is upper bounded, we restrict the maximum size of the sets $g_c$ to $G$.
The maximum size of a search space $\GGn$ is thus given by $S = |\KKn|\,|\GGc| + 1 = C'G + 1$.
In general, however, the index sets $\GGc$ may intersect, leading to smaller $\GGn$.

\begin{figure}[tb]
  \centering
  \includegraphics[width=0.8\textwidth]{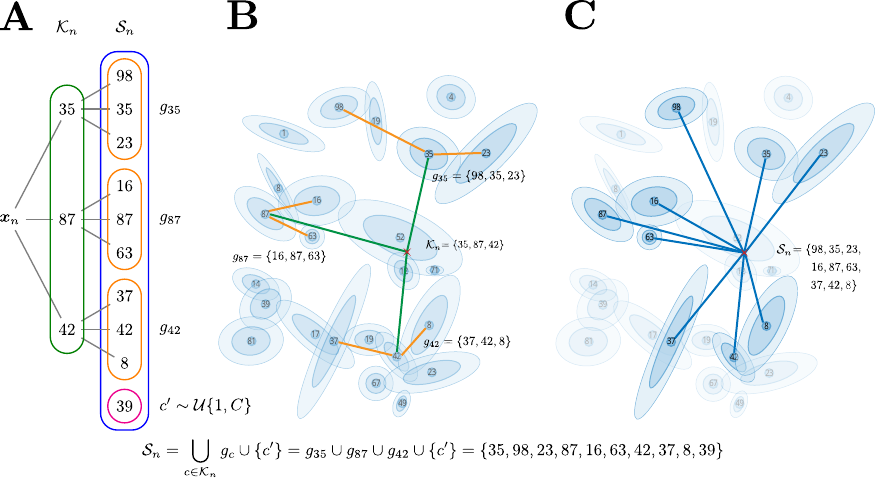}
  \caption{
    Construction of $\GGn$ for a data point $\bxn$. Subfigure \textbf{A} shows a schematic representation while \textbf{B} visualizes $\KKn$ and $\GGc$, and \textbf{C} visualizes $\GGn$ in data space.
    In this example, the set $\KKn = \{35, 87, 42\}$ contains $C'= 3$ component indices for $\bxn$, as indicated by the green box in \textbf{A} and green lines in \textbf{B}.
    Each of the three components has a set of neighborhoods $\GGc$ that includes the component $c$ itself and two additional component indices, i.e., $|\GGc| = G = 3$.
    Specifically, we have $g_{35} = \{98, 35, 23\}$, $g_{87} = \{16, 87, 63\}$ and $g_{42} = \{37, 42, 8\}$. These three sets are highlighted by orange boxes in \textbf{A} and orange lines in \textbf{B}.
    The union of the three sets $\GGc$ and a randomly added component $c' = 39$ forms the search space $\GGn = \{98, 35, 23, 16, 87, 63, 37, 42, 8, 39\}$, as indicated by the blue box in \textbf{A} and blue lines in \textbf{C}.
    None of the sets are optimal for the data point $\bxn$ yet. The optimization procedure will be explained in the following.
  }
  \label{fig:search_space}
\end{figure}

The construction of the search space by \cref{eq:GGnSet} can be conceptualized as a search tree structure.
Specifically, we build $\GGn$ by first considering all components in $\KKn$,
which contains the most relevant components identified so far for $\bxn$.
Next, for each component~\mbox{$c\in\KKn$}, we consider the set of components in $\GGc$, that provides replacement candidates for component~$c$.
\Cref{fig:search_space} offers an illustration of the sets $\GGc$ and the construction of $\GGn$. \Cref{fig:search_space}\textbf{A} shows a schematic representation, \Cref{fig:search_space}\textbf{B} visualize $\KKn$ and $\GGc$, and \Cref{fig:search_space}\textbf{C} illustrates $\GGn$ in data space. \Cref{fig:search_space} also demonstrates that $|\GGn|$ is larger than $|\KKn|$, but significantly smaller than $C$.

It remains to define the sets $g_c$.
In previous work \citep[][]{HirschbergerEtAl2022}, the sets $\GGc$ were updated based on Euclidean neighborhoods.
Concretely, a set $\GGc$ contained components $\ct$ with small Euclidean distances $||\bmu_c-\bmu_{\ct}||$ to the component $c$ (with $\bmu_c$ and $\bmu_{\ct}$ denoting component centers).
These component-to-component distances were then approximated using component-to-datapoint distances (for further details, see~\citealp{HirschbergerEtAl2022}).

\begin{figure}[tb]
  \centering
  \includegraphics[width=0.90\textwidth,trim={0 0 8cm 0},clip]{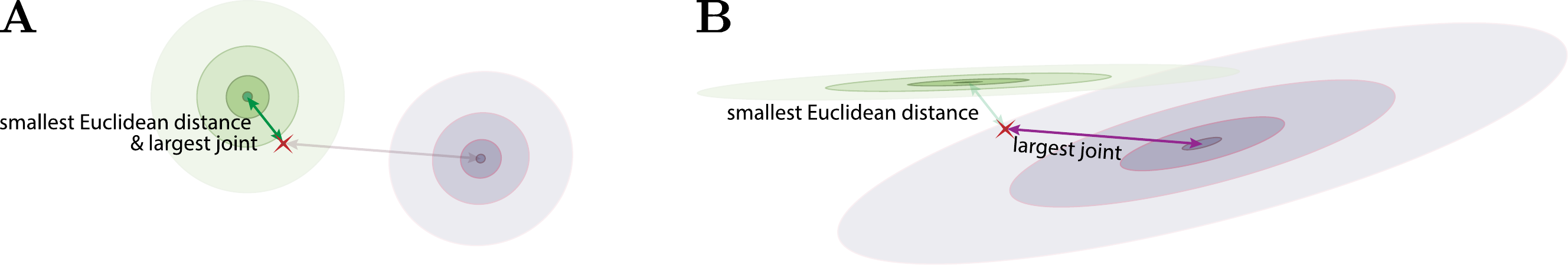}
  \caption{Comparison of Euclidean distances and joints. In \textbf{A} and \textbf{B}, the component centers are identical and the same data point is marked with a red cross. For isotropic, equally weighted components (\textbf{A}), the smallest Euclidean distance correspond to the largest joint between components and the data point. However, for general covariance matrices (\textbf{B}), this is no longer the case.}
  \label{fig:distance_vs_logjoints}
\end{figure}

In our case, however, covariance matrices are not restricted to the uncorrelated case.
The general and potentially strong correlations we seek to model can strongly change proximity relations among components as well as between data points and components, which makes previous Euclidean distance metrics \citep[][]{HirschbergerEtAl2022,ExarchakisEtAl2022} unsuitable for finding relevant components for a given component. \Cref{fig:distance_vs_logjoints} highlights that for general covariance matrices, small Euclidean distances no longer correspond to large joint values.

Instead of finding components similar to component $c$ using Euclidean distances, we measure component similarities via the Kullback-Leibler (KL) divergence.
The KL-divergence between the probability distributions of a component $c$ and component $\ct$ is given by
\begin{align}
  \DKL = \mathbb{E}_{p(\bx\,|\,c, \bTheta)}\left[\log \frac{p(\bx\,|\,c, \bTheta)}{p(\bx\,|\,\ct, \bTheta)} \right]
  \label{eq:kldivergence}.
\end{align}
Although there exists an analytically tractable form of the KL-divergence for Gaussian distributions, evaluating \cref{eq:kldivergence} introduces additional computational overhead.
Concretely, the divergence would have to be computed for all $(c, \ct)$ pairs, resulting in a complexity of $\OO(C^2DH)$ for the MFA model.
As we aim at handling large numbers of components efficiently, the quadratic scaling with $C$ quickly becomes a substantial computational bottleneck.
Thus, we aim to efficiently estimate the KL-divergence, while still capturing sufficient information about the KL-divergences for those $(c, \ct)$ pairs that are relevant for the optimization algorithm.
We stress that this highly efficient estimation of the KL-divergence for GMMs with general covariances is the crucial step to enable scalability of the whole variational algorithm.
While other measures are conceivable, the use of KL-divergences may in our context be a canonical choice.

To achieve an as efficient as possible estimation of \cref{eq:kldivergence}, we employ a finite sample approximation.
Instead of drawing samples from $p(\bx\,|\,c, \bTheta)$, we estimate the KL-divergence by using the data points currently assigned to component~$c$.
This gives rise to a partitioning of the data set into the following sets:
\begin{align}
  \II_c = \{n \mid c=c_n \} \quad \text{with}  \quad c_n = \myargmax{c'\in\KKn} \  p(c'\,|\,\bxn,\bTheta) = \myargmax{c'\in\KKn} \  p(c',\bxn\,|\,\bTheta). \label{eq:IIPartition_hard}
\end{align}
Given the sets $\II_c$,
the KL-divergence between the probability distributions of a component $c \in \KKn$ and a potential candidate $\ct$ to add to $\GGc$ can be estimated as follows (see \cref{appendix:comparison_to_KL} for details)
\begin{align}
  \DKL \approx & \frac{1}{N_{c\ct}} \sum_{n\in\II_c} \log \frac{p(\bxn\,|\,c, \bTheta)}{p(\bxn\,|\,\ct, \bTheta)} \delta(\ct \in \GGn) \eqdef \rel \label{eq:deltaF} \\
               & \text{where} \quad N_{c\ct}                                       = \sum_{n\in\II_c} \delta(\ct \in \GGn). \nonumber
\end{align}
If we disregard the condition $\delta(\ct \in \GGn)$, the function $\rel$ estimates the KL-divergence using \emph{all} data points in~$\II_c$.
The condition $\delta(\ct \in \GGn)$, however, reduces the number of samples to approximate the KL-divergence by considering only those data points for which component $\ct$ is present in their search spaces. This ensures that no additional computations of $p(\bxn\,|\, \ct, \bTheta)$ are required beyond those
that anyway have to be performed in a partial E-step (see \cref{sec:algorithmic_realization} for details).
As a result, $\rel$ are very efficiently computable, which is the primary motivation for employing this approximation.
Due to $\delta(\ct \in \GGn)$ in \cref{eq:deltaF}, the function $\rel$ is defined only for a restricted subset of components $\ct$.
For those $\ct$ that do not appear in any of the search spaces, (i.e., any $\ct \notin \bigcup_{n\in\II_c} \GGn$), we set $\rel$ to infinity, formally treating the component $\ct$ for $c$ as irrelevant.

\Cref{eq:deltaF} yields a ranking, where a lower value of $\rel$ indicates a greater similarity of component~$\ct$~to~$c$.
Consequently, we can define $g_c$ as the set of those components that result in the $G$ lowest values of $\rel$, i.e.,
\begin{align}
  \GGc = \{\ct \mid \rel \mbox{\ is among the $G$ lowest values} \}. \label{eq:GUpdate}
\end{align}

A potential drawback of using \cref{eq:deltaF} is that $\rel$ may only provide a coarse estimate of the KL-divergence, depending on the convergence state of the search spaces, the model parameters, etc.\ (we elaborate in \cref{appendix:comparison_to_KL}).
Importantly, however, the estimate still captures sufficient information about component similarities in the sense of KL-divergences.
Since we only require the values of $\rel$ to rank components $\ct$, their exact values are of secondary importance.

We note that a relation to KL-divergences was also discussed in \citet{HirschbergerEtAl2022}. However, the general distance introduced in \citet{HirschbergerEtAl2022} was approximated in such a way that, for isotropic GMMs, it recovers a measure based on Euclidean distances. Further details, along with numerical comparisons to our novel and generalized method, are provided in \cref{appendix:comparison_approx_analytical,appendix:comparison_approx_numerical}.
While the here defined method (based on \mbox{\crefnobrackets{eq:deltaF})} is especially advantageous in the general case with intra-component correlations, we also observed improvements for the uncorrelated case.

\subsection{Algorithmic Realization of \varMFA{}}
\label{sec:algorithmic_realization}

While \crefrange{sec:varEM}{sec:construction_Gn} described the theoretical foundation of the variational learning algorithm for MFA models, we
now focus on its algorithmic realization (which we will refer to as \varMFA{}).
Analogously to conventional EM optimization, \varMFA{} alternates between E-steps and M-steps to update variational parameters and model parameters, respectively.
The variational E-step is discussed below, and details on the M-step are provided in \cref{appendix:Mstep}.

\begin{figure}[tb]
  \centering
  \includegraphics[width=\textwidth]{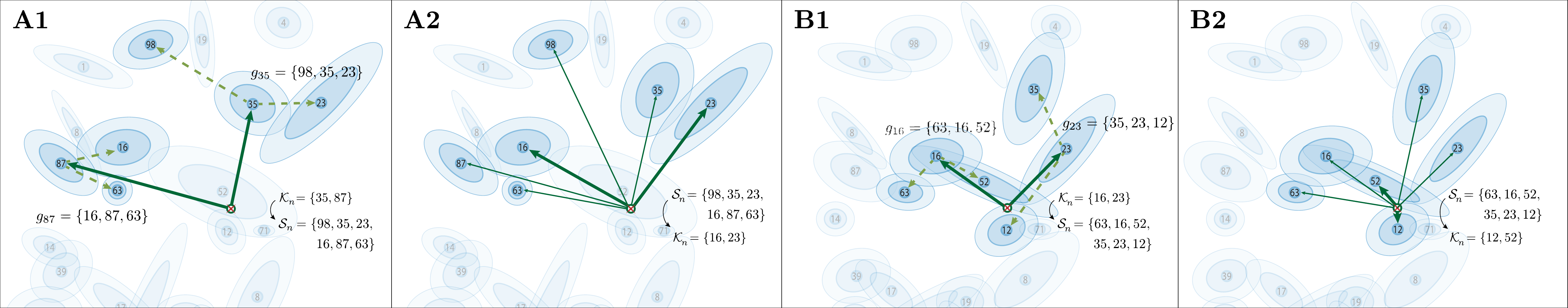}
  \caption{
    Update of $\KKn$ over two iterations.
    In \textbf{A1}, a red cross marks a data point $\bx_{n}$ with an initial suboptimal $\KKn = \{35, 87\}$ (for simplicity, we use only $C' = 2$ here) and $G=3$ replacement candidates for each of those components: $g_{35} = \{98, 35, 23\}$ and $g_{87} = \{16, 87, 63\}$. Also these replacement candidates are still suboptimal.
    Subsequently, in \textbf{A2}, the joints  $p(c, \bx_{n} \,|\, \bTheta)$ for all components in the search space $\GGn = g_{35} \cup g_{87} \cup \{c'\} = \{98, 35, 23, 16, 87, 63, 39\}$ are calculated, and the $C' = 2$ components with the largest joints are selected as new $\KKn = \{16, 23\}$.
    In the following steps of the algorithm, the replacement candidates $\GGc$ will be adjusted based on the current $\GGn$, and the model parameters will be updated in the M-step.
    In \textbf{B1} and \textbf{B2} the same procedure is repeated in the next iteration to find components $\KKn$ with larger joints.
    In this example, over two iterations, a total of 14~joints were calculated to find the best final $\KKn$, which is only a fraction of the total number of calculations required had we considered all $C$ components.
  }
  \label{fig:kn_update}
\end{figure}

\myparagraph{Variational E-step}
\Cref{alg:estep} shows the partial variational E-step, which comprises four blocks (variables are implicitly initialized with zeros or empty sets, respectively):
\begin{enumerate}[label=\textbf{(\arabic*)},noitemsep]
  \item For each data point, the search space $\GGn$ is constructed and a component, drawn from a discrete uniform distribution $\mathcal{U}\{1,C\}$, is added to $\GGn$ (it consequently applies that $|\GGn|\leq{}C'G+1$). Subsequently, the joints are evaluated for all $c \in \GGn$. The $\KKn$ sets are then optimized using these joints. \Cref{fig:kn_update} visualizes the iterative procedure for $\KKn$ optimization.
  \item The partition $\II_{1:C}$ of the data set is computed.
  \item The partition and the previously computed joints are used to determine the sets $g_{1:C}$, utilizing $\log p(\bxn\,|\,c, \bTheta) = \log p(c, \bxn\,|\,\bTheta) - \log p(c\,|\,\bTheta)$.
        $\rel=\relt/N_{c\ct}$ will be computed for all $\ct\in \bigcup_{n\in\II_c} \GGn$ excluding $\ct=c$, because $c$ will always be included in $\GGc$.
  \item  The variational distributions $\qnc$ are computed by normalizing of $p(c,\bxn\,|\,\bTheta)$.
\end{enumerate}
\begin{algorithm}[tb]
  \For{$n=1:N$}{
    $\GGn =\bigcup_{c\in\KKn}\GGc$;\algBreak
    $\GGn =\GGn \cup \{c\}$ with $c \sim \mathcal{U}\{1,C\}$;
    \ \\[-10pt]
    \nl\\[-4pt]
    \For{$c\in\GGn$}{
      compute joint $p(c,\bxn\,|\,\bTheta)$;
    }
    $\KKn \!=\! \{c \mid p(c,\bxn\,|\,\bTheta)$ is among the $C'$ largest joints for all $c \in \GGn\}$;\\
  }
  \algline \\
  \For{$n=1:N$}{
    $\con = \myargmax{c\in\KKn} \ p(c,\bxn\,|\,\bTheta)$;\\[-10pt]
    \nl\\[0pt]
    $\II_{\con} = \II_{\con} \cup \{n\}$;
  }
  \algline \\
  \For{$c=1:C$}{
    \For{$n\in\II_c$}{
      \For{$\ct\in\GGn \setminus \{c\}$}{
        $\relt = \relt + \log p(\bxn\,|\,c, \bTheta) - \log p(\bxn\,|\,\ct, \bTheta)$;\\[-7pt]
        \nl\\[-3pt]
        $N_{c\ct} = N_{c\ct} + 1$;\\
      }
    }
    $\GGc= \{\ct\,|\,(\relt / N_{c\ct})  \,$ is among the $G-1$ smallest values\\
    \phantom{$\GGc= \{$}of all computed $(\relt / N_{c\ct})\} \cup \{c\}$;
  }
  \algline\\

  \For{$n=1:N$}{
    \nl\\[-12pt]
    \For{$c\in\KKn$}{
      $\qnc = p(c,\bxn\,|\,\bTheta)/\sum_{\ct \in \KKn} p(\ct,\bxn\,|\,\bTheta)$ \;
    }
  }
  \caption{Variational E-step}
  \label{alg:estep}
\end{algorithm}
The computational complexity of a single variational E-step is $\OO(NSDH)$, dominated by the joint computations in block 1 of \cref{alg:estep} (assuming $N > C$).
Unlike conventional E-steps, which require evaluating $NC$ joint probabilities by considering combinations of all data points with all components, the variational E-step only loops over all data points and their respective search spaces. As a result, it requires at most $NS$ joint evaluations, which is significantly fewer when $S \ll C$.
A detailed specification of the complexity is provided in \cref{appendix:estep_complexity}.

\myparagraph{Initialization of model parameters} To preserve the algorithmic complexity of \cref{alg:estep}, initialization methods should
exhibit complexities that are comparable to or lower than
those of the optimization algorithm itself.
The \kmeans{}{\tt++} seeding method \citep{ArthurVassilvitskii2007} is not suitable in our context, for instance, because it scales with $\OO(NCD)$.
Therefore, we employ AFK-MC$^2$ \citep{BachemEtAl2016b} as an efficient initialization method for the means $\bmuc$. To initialize the diagonal variances $\bD_{1:C}$, we compute the variance
along each dimension across all data points as follows
\begin{align}
  \bDc = \bD_\mathrm{init} \quad \forall c \qquad \text{with } \bD_\mathrm{init} = \diag(\sigma^2_{1}, \cdots, \sigma^2_{D}) \text{ and } \sigma^2_{d} = \var( x_{1:N,d} ), \nonumber
\end{align}
where $x_{n,d}$ is the $d$-th entry of $\bxn$.
For the factor loading matrices $\bAc$, we use uniform random numbers between 0 and 1 to set their values.
Initialization methods proposed by \citet{Richardson2018} and \citet{McLachlanEtAl2003} are computationally too demanding for our specific setting.

\myparagraph{Initialization of variational parameters} The initialization of index sets $\mathcal{K}_{1:N}$ and $g_{1:C}$ proceeds as
follows: Data points designated as seeds for components, sampled by
AFK-MC$^2$, ensure that the corresponding component index is contained within the respective set
$\KKn$.
The set $g_{c}$ consistently includes index $c$. Subsequently, $\mathcal{K}_{1:N}$ and
$g_{1:C}$ are populated with further indices by sampling uniformly from $\{1,\ldots,C\}$ (while
ensuring uniqueness of the indices).
Following this, $\mathcal{K}_{1:N}$ and $g_{1:C}$ are optimized through initial partial E-steps, while
keeping the model parameters fixed. This approach has been previously shown
to yield favorable results \citep{HirschbergerEtAl2022}. We refer to this procedure as \warmup{}.

\myparagraph{Complete \varMFA{}} The implementation of the complete \varMFA{} algorithm is outlined in \cref{alg:varMFA}.
Following initialization, the algorithm performs variational E-steps in a \warmup{} phase until the free energy $\mathcal{F}(\boldsymbol{\mathcal{K}}, \bTheta)$ has converged.
Subsequently, the main training process alternates between variational E-steps and M-steps until $\mathcal{F}(\boldsymbol{\mathcal{K}}, \bTheta)$ has converged again.
As convergence criterion for both loops, we consider $\mathcal{F}(\boldsymbol{\mathcal{K}}, \bTheta)$ to have converged after iteration $t$ if
\begin{align}
  \label{eq:convergence}
  \frac{\FF^{(t)} - \FF^{(t-1)} }{\FF^{(t-1)}} < \varepsilon,
\end{align}
where $\FF^{(t)}$ is the free energy $\mathcal{F}(\boldsymbol{\mathcal{K}}, \bTheta)$ of \cref{eq:FEShort} after iteration $t$ \citep[cf.][]{HirschbergerEtAl2022}.
In all numerical experiments we choose the convergence threshold $\varepsilon = 10^{-4}$ unless specifically stated otherwise.

\begin{algorithm}[tb]
  initialize $\pi_{1:C}$, $\bmu_{1:C}$, $\bA_{1:C}$, $\bD_{1:C}$, $\mathcal{K}_{1:N}$ and $g_{1:C}$ (see text for details)\;
  \
  \Repeat{$\mathcal{F}(\boldsymbol{\mathcal{K}}, \bTheta)$ has converged}{
    \nl \emph{variational E-step:}
    update $\mathcal{K}_{1:N}$, $g_{1:C}$ using \cref{alg:estep};
  }
  \algline \\
  \Repeat{$\mathcal{F}(\boldsymbol{\mathcal{K}}, \bTheta)$ has converged}{
    \emph{variational E-step:}
    update $\mathcal{K}_{1:N}$, $g_{1:C}$ using \cref{alg:estep};
    \ \\[-6pt]
    \nl\\[-6pt]
    \emph{variational M-step:}
    update $\pi_{1:C}$, $\bmu_{1:C}$, $\bA_{1:C}$ and $\bD_{1:C}$ (see \cref{appendix:Mstep} for details)\;
  }
  \caption{\varMFA{}}
  \label{alg:varMFA}
\end{algorithm}

\section{Numerical Experiments}
\label{sec:numerical_experiments}

To assess the efficiency and effectiveness of the \varMFA{} algorithm, we conducted numerical experiments organized into four sections.
In \cref{sec:scalability}, we provide a comprehensive comparison of the scaling behavior of \varMFA{} with increasing component numbers $C$
and compared to conventional EM optimization of MFA models (which we will refer to as \fullMFA{}).
Throughout our evaluation, the \fullMFA{} algorithm serves as a canonical baseline.
Following the scaling analysis, we evaluate in \cref{sec:quality} the performance and quality of \varMFA{} relative to \fullMFA{}.
We also compare \varMFA{} to a \kmeans{} algorithm combined with factor analysis \mbox{\citep[cf.][]{Richardson2018}}, denoted as \kmeansfa{}.
In \cref{sec:largescale}, we extend the analysis to data set and model sizes where the application of \fullMFA{} is no longer feasible.
On such tasks the efficiency of \varMFA{} translates to its ability to train GMMs at scales that have previously not been reported.
Finally, in \cref{sec:denoising}, we benchmark the performance and efficiency of \varMFA{} on one specific downstream task: image denoising.
The task allows for a comparison of \varMFA{} to various other algorithms, including self-supervised deep neural networks (DNNs) and variational autoencoder-based approaches
as examples of deep generative models.
Further details on all numerical experiments are provided in \cref{appendix:further_details}, and further control experiments are reported in \cref{appendix:control_experiments}.

In \cref{sec:scalability,sec:quality}, we evaluated \varMFA{} on four well-established benchmarks for high-dimensional image data, namely CIFAR-10 \citep{Krizhevsky2009}, CelebA \citep{Liu2015}, EMNIST \citep{Cohen2017} and SVHN \citep{Netzer2011} (for details, see \cref{appendix:datasets}).
The data sets have frequently been applied in different contexts, and they were used previously in settings similar to the one of interest here \citep{Richardson2018,HirschbergerEtAl2022,ExarchakisEtAl2022}.
Image data is commonly used for benchmarking, and is often considered as an example for data points lying close to a data manifold with much lower dimension than the dimensionality of the data.
Images are consequently well suited to assess the effectiveness and efficiency of algorithms that optimize MFAs.
Moreover, the chosen data sets provide a sufficiently large number of data points to estimate components for the large-scale MFA models we are interested in.

In the numerical experiments, we evaluated the efficiency of \varMFA{} and corresponding baselines in different settings using two measures for computational costs: the total number of joint probability evaluations (treated as generalized distances) and the wall-clock runtime \citep[cf.][]{HirschbergerEtAl2022,ExarchakisEtAl2022}.

The first measure, the total number of joint evaluations, is independent of concrete implementation details and the hardware used.
The number of joint evaluations also excludes initialization, parameter updates within the M-steps and auxiliary computations.
The second measure, the wall-clock runtime, encompasses all computations, memory management, and other auxiliary routines.
Actual improvements in wall-clock runtime are practically relevant, although this measure depends on implementation details and the specification
of the used hardware.

For the models under consideration in \cref{sec:scalability,sec:quality,sec:largescale}, we assess the algorithms' effectiveness using the negative log-likelihood (NLL), which represents a canonical quality measure
for probabilistic models \citep[e.g.,][]{Theis2016}.
In experimental settings where the \mbox{\fullMFA{}} algorithm can be used as a reference, we employ the relative NLL given by
\begin{align}
  \label{eq:relnll}
  \text{rel. NLL} = \frac{\text{NLL}_\text{algo} - \text{NLL}_\text{\fullMFA{}}}{\text{NLL}_\text{\fullMFA{}}},
\end{align}
where $\text{NLL}_\text{algo}$ refers to the NLL of the respective algorithm (e.g., \varMFA{}), and where $\text{NLL}_{\text{\fullMFA{}}}$ is the NLL of the \fullMFA{} algorithm.
Distance-based metrics that do not account for correlations within components (such as the quantization error used for $k$-means)
are much less appropriate for our analysis.
In the denoising benchmark in \cref{sec:denoising}, we quantify denoising performance using peak signal-to-noise ratio (PSNR) and structural similarity index measure (SSIM).

\subsection{Scalability}
\label{sec:scalability}

The \varMFA{} algorithm exhibits a reduced complexity per iteration compared to \fullMFA{} (cf. \cref{sec:varEM}).
However, a reduction per iteration does not guarantee a reduced complexity for the overall training.
Increasingly many iterations could negate the efficiency gain per iteration.
Furthermore, \varMFA{} requires updating the additional variational parameters.

To investigate the scalability of \varMFA{} and \fullMFA{}, we measured their overall computational demands per data point as the number of mixture components~$C$ increases.
For effective training, increasing model parameters commonly requires a corresponding increase in the amount and complexity of the data.
Maintaining a constant data set size while increasing $C$ could simplify optimization, typically resulting in a reduced number of iterations.
To mitigate this effect, we simultaneously increase the data set size alongside~$C$.
We constructed these data sets by uniformly subsampling $\tilde{N}$ data points for each value of $C$, where $\tilde{N} = N \cdot C / C_\mathrm{max}$, maintaining a constant ratio of $\tilde{N}/C$.
For the largest value $C = C_\mathrm{max}$, the entire data set was used.
Our scalability evaluation differs in detail from previous evaluations \citep[e.g.,][]{HirschbergerEtAl2022}, but we also adjusted data set size with $C$ to measure the efficiency, mitigating secondary effects on the number of iterations.

We applied \varMFA{} and \fullMFA{} to all four benchmark data sets, with component numbers ranging from $C=100$ to $800$ in steps of $100$, while fixing the hyperplane dimensionality to $H=5$.
In addition to these hyperparameters, the v-MFA algorithm includes the variational hyperparameters $C'$ and $G$.
While a precise definition of these variational hyperparameters is provided in \cref{sec:material_and_methods}, $S=C'G+1$ defines the size of the search spaces, which control the upper limit of joint probabilities evaluated per iteration by the \varMFA{} (with \mbox{$S\leq C$}).
Consequently, these variational hyperparameters balance optimization quality and computational efficiency in the algorithm.

We used the same variational hyperparameters ($C'=3$ and $G=15$) across all data sets and component numbers. For the effect of different hyperparameters on the results, we refer to the experiments in \cref{sec:quality}.
The \varMFA{} and \fullMFA{} algorithms iterate (variational) EM updates until a convergence criterion is reached (see \crefnobrackets{eq:convergence}).

\begin{figure}[p]
  \includegraphics[width=0.98\textwidth]{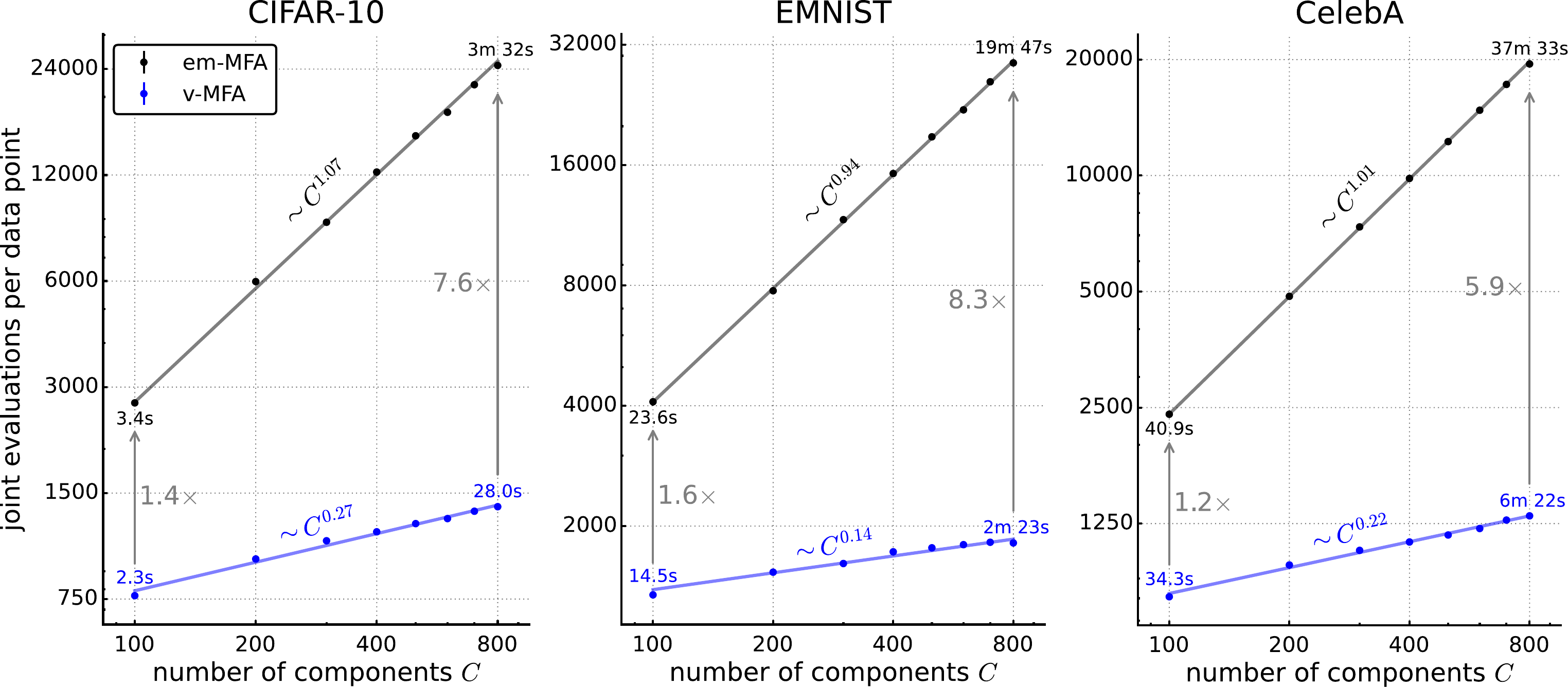}
  \vspace{0.8em} \\
  \begin{minipage}[c]{0.34\textwidth}
    \includegraphics[width=\textwidth]{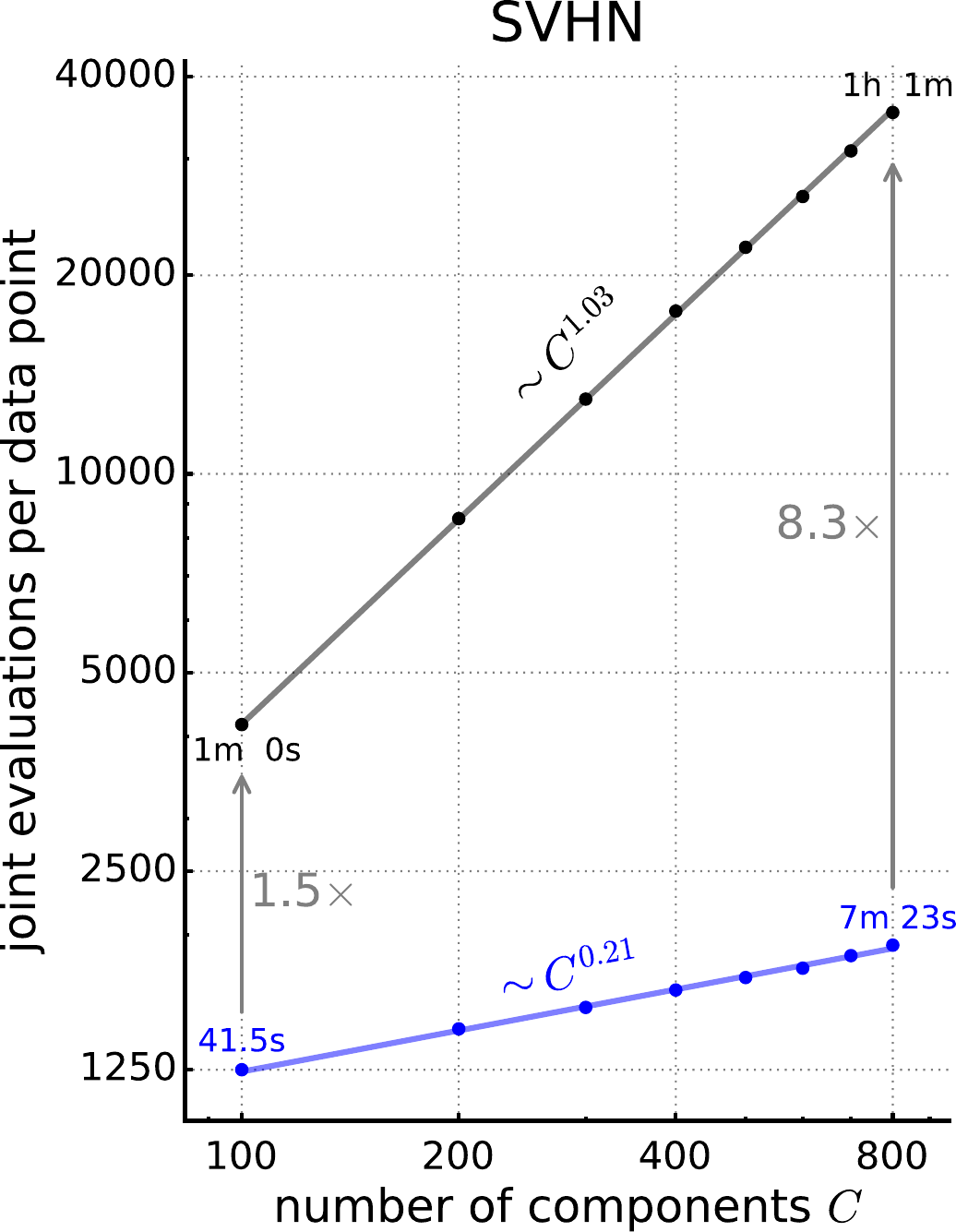}
  \end{minipage}
  \hspace{15pt}
  \begin{minipage}[c]{0.61\textwidth}
    \caption{Scaling behavior of \varMFA{} and \fullMFA{}.
      One measurement point shows, for a given GMM with $C$~components, the number of joint probabilities computed per data point until the convergence criterion is reached.
      The number of components~$C$ is increased from $C=100$ to $C=800$, and subplots show the results for the datasets CIFAR-10, EMNIST, CelebA and SVHN.
      Each measurement point is averaged over $40$ independent runs, with error bars indicating standard error of the mean (SEM). Note that the SEM values are very small and barely visible in the plot.
      The legend in the first subfigure applies to all four subfigures. Lines represent regressions used to determine the scaling exponent of $C$ (see main text), while annotations provide information on total wall-clock runtimes. Additionally, gray vertical arrows indicate the relative wall-clock speed-up of \varMFA{} (i.e., the wall-clock runtime required by \fullMFA{} divided by the wall-clock runtime required by \varMFA{}).
    } \label{fig:scaling}
  \end{minipage}
\end{figure}

The results of the scalability analysis are summarized in \cref{fig:scaling}.
To quantify the scalability with $C$, we fitted regressions of the form $f(C) = b \, C^a$.
This corresponds to linear functions $\log\left(f(C)\right) = \tilde{b} + a \, \log\left(C\right)$ in log-log plots. Hence, the slopes in \cref{fig:scaling} provide the scaling exponents for $C$.

For \fullMFA{}, each iteration requires precisely $NC$ joint evaluations, i.e., $C$ evaluations per data point.
The number of iterations tends to increase slightly with higher $C$, as also previously observed \citep[e.g.,][]{HirschbergerEtAl2022}.
This leads to a scaling exponent slightly above one for CIFAR-10, CelebA, and SVHN, indicating slightly superlinearly scaling with $C$.
Conversely, for EMNIST, the exponent falls slightly below one, as \fullMFA{} (and \varMFA{}) required fewer iterations to converge with increasing $C$.

In contrast to \fullMFA{}, the number of joint evaluations per data point and iteration is upper-bounded for \varMFA{}, i.e., it is limited by
the search space size $S=C'G +1 \leq{}C$ (cf. \cref{sec:construction_Gn}).
In practice, the actual number is often lower than this bound.
It may be that \mbox{\varMFA{}} requires more iterations than \fullMFA{} due to its partial E-steps, and we indeed observe more iterations in our empirical evaluations.
Importantly, however, our empirical results for \varMFA{} clearly show a substantial overall gain in computational efficiency compared to \fullMFA{}.
This gain presents itself as a different complexity scaling with $C$ (in contrast to, e.g., gains in terms of constant offsets).
Concretely, the joint evaluations per data point clearly scale {\em sublinearly} with $C$, with all scaling exponents consistently below $1/3$ (cubic root dependency, $\sqrt[3]{C}$).

For instance, for $C=100$, \fullMFA{} required approximately three times as many joint evaluations compared to \varMFA{} across all data sets.
However, as $C$ increased to $800$, the difference spanned an order of magnitude.
For even larger $C$, the reduction in joint evaluations for \varMFA{} compared to \fullMFA{} increases still further (we investigate such larger $C$ in the next sections).

When observing the training time (depicted as annotations in \cref{fig:scaling}), we discern marginal speed-ups of \varMFA{} over \fullMFA{} for $C=100$.
For $C=800$, \varMFA{} significantly outperformed \fullMFA{}, which shows not only the improved theoretical complexity scaling but also practical runtime reductions.
Analogous to the joint evaluations, the improvements in runtimes increase with larger values of~$C$ (cf.\ \Cref{sec:largescale}).

The results in \cref{fig:scaling} so far consider solely the computational costs, not the quality of the optimization.
To ensure that the observed improvements of \varMFA{} are not primarily due to a loss in quality, additional control experiments were conducted (see \cref{appendix:scaling-same-quality}).
When we compare the computational demand required to achieve the same optimization quality (in terms of NLL$_\mathrm{test}$), the speed-up of \varMFA{} over \fullMFA{} remains essentially consistent with \cref{fig:scaling}.
Only for EMNIST, where \varMFA{} showed the strongest speed-ups, a non-negligible (but small) fraction of the speed-up is due to a slight reduction in quality.

\subsection{Quality vs. Efficiency Analysis}
\label{sec:quality}

In \cref{sec:scalability}, it was shown that the number of joint evaluations per data point scales sublinearly w.r.t.~the number of components $C$ for \varMFA{}.
While those experiments focused on complexity scaling with~$C$, we will now systematically compare optimization quality and efficiency by evaluating different algorithms on these four benchmark data sets.

We evaluated different algorithms, including \fullMFA{} and \varMFA{}, by comparing wall-clock runtimes against optimization quality, measured by relative NLL values on the test sets.
To provide references, we also included NLL values after initialization to highlight the learning improvements of the algorithms.
Additionally, \varMFA{} and \fullMFA{} were compared to the \kmeansfa{} algorithm, which integrates \kmeans{} with factor analyzers (details in \cref{appendix:algorithms}).
This approach follows \citet{Richardson2018}, who used \kmeans{} and factor analysis for MFA initialization.

For the four data sets, we selected different numbers of components: $C=800$ for CIFAR-10, $C=1200$ for CelebA, $C=1800$ for SVHN, and $C=2000$ for EMNIST.
The hyperplane dimension was set to $H=5$ across all data sets.
For \varMFA{}, we evaluated the performance across nine settings of the variational hyperparameters (all combinations of $C' \in \{3,5,7\}$ and $G \in \{5,15,30\}$).
To initialize component means for the algorithms, we used the same initialization procedure.
The same applies for the variances for \varMFA{} and \fullMFA{}.

\begin{figure}[htbp]
  \centering
  \includegraphics[width=0.95\textwidth]{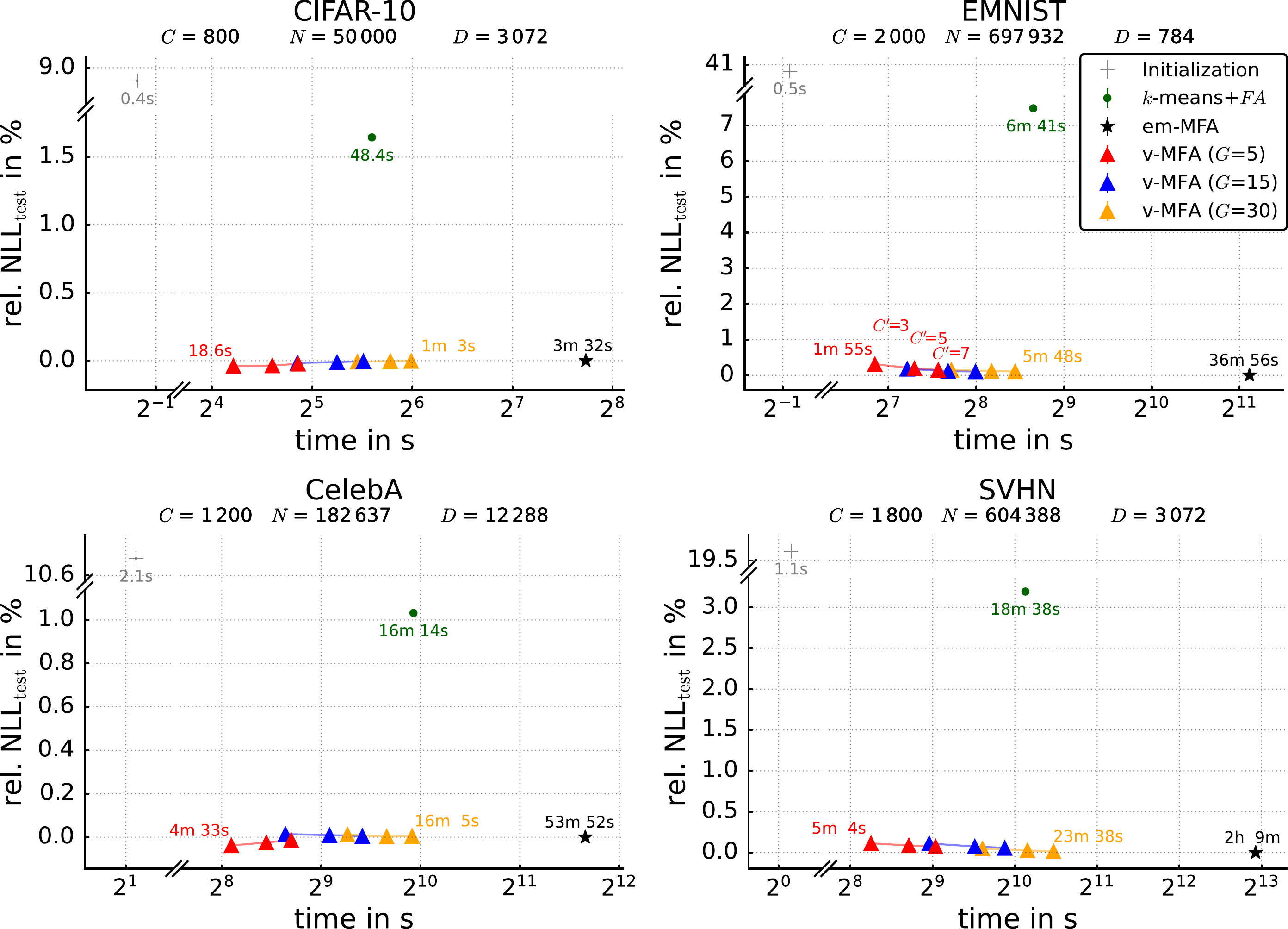}
  \caption{
    Effectiveness and efficiency in terms of negative log-likelihood (NLL) and runtime of \varMFA{} compared to \fullMFA{}, \kmeansfa{} and the parameters after initialization.
    Each of the four subfigures refers to experiments on one benchmark data set. The y-axes denote the relative NLL on the test set w.r.t.\ \fullMFA{} as baseline, given by \cref{eq:relnll}.
    For \varMFA{}, we present the performance across nine different settings of the hyperparameters $C'$ and $G$. For each configuration of $G \in \{5,15,30\}$, measurements for \varMFA{} (three red, blue and yellow triangles, respectively) refer to configurations with $C' \in \{3,5,7\}$. Settings with larger $C'$ lie to the right, as they required longer runtimes. Each measurement point is averaged over $40$ independent runs, with error bars indicating standard error of the mean (SEM) (note that the SEM values are smaller than the symbol sizes).
  }
  \label{fig:quality}
\end{figure}

\Cref{fig:quality} shows the measured NLL$_\mathrm{test}$ values vs.\ wall-clock runtimes for all algorithms.
As can be observed, \varMFA{} (with $C' = 3$ and $G = 5$) was the fastest among the algorithms evaluated on all four data sets (excluding Initialization). Across all settings of hyperparameters $C'$ and $G$, \varMFA{} consistently outperformed \fullMFA{}, achieving substantial speed-ups ranging from $3.3\times$ to $25.6\times$.
Additionally, \varMFA{} achieved faster runtimes than \kmeansfa{} for the majority of hyperparameter settings.

In terms of optimization quality, \fullMFA{} achieved the best (lowest) NLL values on the EMNIST and SVHN data sets, while \varMFA{} shows the best NLL values on the CIFAR-10 and CelebA data sets (with $C' = 3$ and $G = 5$).
The relative deviations of \varMFA{} from the baseline remained consistently below $0.32\%$, with most settings of the hyperparameters $C'$ and $G$ showing smaller deviations.
The \kmeansfa{} algorithm showed significantly higher NLL values, exhibiting a deviation from the baseline by approximately 1\% for data sets such as CelebA and CIFAR-10, and up to 3\% and 7\% for the SVHN and EMNIST data sets, respectively.
Further details on these results are provided in \cref{appendix:further_results}.

The performance of the \varMFA{} algorithm relative to \fullMFA{} depends, among other factors, primarily on two aspects: (1) how well the assumption of truncated posteriors fits the data, and (2) how effectively the variational EM algorithm optimizes the variational parameters.
To provide a more detailed analysis of the efficiency-quality trade-off between \varMFA{} and \fullMFA{}, \cref{appendix:v-mfa_quality} presents an ablation study.
This study examines the extent to which the observed differences in optimization performance between \varMFA{} and \fullMFA{} can be attributed to the use of truncation (with the optimal $\KKnopt$) versus the variational approximation employed to determine the $\KKn$ sets.

Additionally, we present control experiments in \cref{appendix:comparison} comparing \varMFA{}, \fullMFA{} and \mbox{\kmeansfa{}} with GMMs with diagonal covariances and previously studied large-scale MFA optimization \citep{Richardson2018} that used a stochastic gradient descent (SGD) approach.
To evaluate the robustness and optimization quality of the \varMFA{} algorithm under different initialization strategies, we also conducted an ablation study, as reported in \cref{appendix:abinit}.

\subsection{GMMs with up to Billions of Parameters}
\label{sec:largescale}
While previous numerical experiments were conducted on large-scale benchmarks, they still allowed for the application of \fullMFA{} (or \kmeansfa{}).
The sublinear scaling of \varMFA{} enables addressing much larger scale problems, however.
In this section, we therefore investigate significantly larger problems, which far exceed the problem sizes that can be addressed by \fullMFA{} or other GMM approaches (on current, reasonable compute resources).
Concretely, we used a data set of approximately 100M images, each sized $D = 32 \times 32 \times 3 = \num{3072}$. The used data set is a version of the YFCC100M data set, with $5\%$ of the samples randomly selected for testing and the remaining $95\%$ for training (see \cref{appendix:datasets}).
We then applied \varMFA{} with increasingly many components. We used $C=\num{500}$, \num{5000}, \num{50000}, and \num{500000}. Unlike in \cref{sec:scalability}, optimization by \varMFA{} used all training samples, regardless of the chosen value of $C$. The other hyperparameters were set as in \cref{sec:scalability}, specifically $H=5$, $C'=3$, and $G=15$.

Large numbers of components have consequences for auxiliary parts of the optimization. First, our default initialization method (AFK-MC$^2$) becomes impractical for the large $C$ values
due its quadratic computational complexity w.r.t.~$C$.
Thus, we employed a less sophisticated initialization approach.
For $\bmu_{1:C}$, we selected data points uniformly at random (without duplicates).
The remaining parameters were initialized as before.
Despite being less sophisticated, this initialization approach is expected to yield stable performance; as shown in \cref{appendix:abinit}, the \varMFA{} algorithm exhibits robustness to the choice of initialization method on the data sets considered in \cref{sec:scalability,sec:quality}.
Second, to further reduce initialization cost, the convergence threshold for the initialization procedure of the variational parameters, referred to as \warmup{}, was set to $\varepsilon = 10^{-3}$, a larger value than in the previous smaller-scale experiments (see \cref{sec:algorithmic_realization}).
And third, we observed that there were components containing no data points in the cases of $C = \num{50 000}$ and \num{500 000}.  These components were ignored, effectively reducing $C$, but the fraction of such `empty' components was (in all experiments) less than 0.2\% of all components.

\begin{figure}[htbp]
  \centering
  \includegraphics[width=0.8\textwidth]{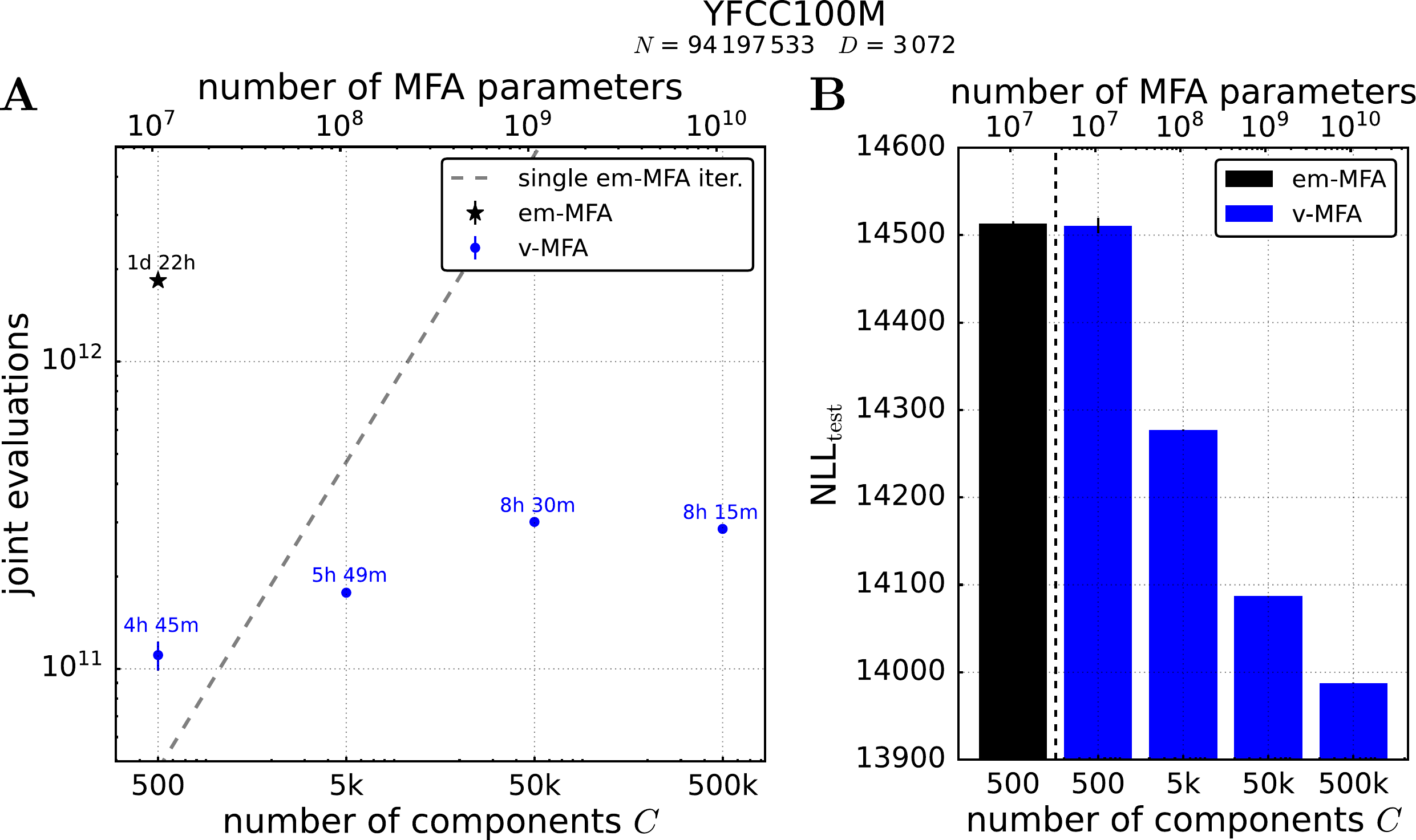}
  \caption{
    Evaluation of \varMFA{} and \fullMFA{} on the YFCC100M data set.
    \textbf{A} shows for different numbers of components $C$ the total number of joint evaluations required for one \varMFA{} optimization run. The number of \varMFA{} model parameters is shown at the top. For each $C$ we also provide the total runtime as annotation. Additionally, the dashed gray line represents the number of joint evaluations for a {\em single} iteration of \fullMFA{} (which is given by $NC$).
    \textbf{B} shows the values of the negative log-likelihood per data point on the test set for \fullMFA{} with $C=500$ and for \varMFA{} with different numbers of components~$C$.
    In both figures, each measurement is averaged over 3 independent runs and error bars denote the standard error of the mean (SEM).
  }
  \label{fig:large_scale}
\end{figure}

Using the described data and MFA optimization, we obtain results for the computational costs and NLL values, which are shown in \Cref{fig:large_scale}.
In the figure, we also express the size of an MFA representation in terms of the total number of optimized floating point parameters, which follows recent customs in studies on large-scale data representations \citep[][]{Dosovitskiy2021,Fang2023,Wortsman2024, KaplanEtAl2020, HenighanEtAl2020}
For an MFA representation, the number of parameters is determined by the number of components $C$, the hyperplane dimensionality $H$ and the data dimensionality $D$. Concretely, the number of MFA parameters is given by the number of parameters for the prior, means, factor loadings and variances, i.e.,
\begin{align}
  C + CD + CDH + CD = C(D(H+2)+1). \nonumber
\end{align}

In \cref{fig:large_scale}\textbf{A}, we observe that both the number of joint evaluations and training runtime increased only moderately as $C$ increased.
This behavior results from the sublinear scaling of \varMFA{}.
From $C = \num{50 000}$ to $C = \num{500 000}$, runtimes even slightly decreased, due to a small reduction in the number of iterations.
At the same time, increasing the number of components $C$ significantly reduced the test set NLL (\cref{fig:large_scale}\textbf{B}). For $C = 500$, the NLL values for \fullMFA{} and \varMFA{} were within the standard error of the mean, consistent with the findings in \cref{sec:quality}, where only minor NLL differences were observed between the two algorithms.
Scaling \fullMFA{} to component numbers significantly larger than $C = 500$ was computationally prohibitive.
Already for $C=500$, \fullMFA{} required approximately two days to train compared to approximately 5 hours for \varMFA{} (and \fullMFA{} required about $16 \times$ as many joint evaluations).
For $C=$ \num{5000} the runtime of \fullMFA{} would still be slower (by about an order of magnitude).
Additionally, \cref{fig:large_scale}\textbf{A} shows that \varMFA{} requires even fewer joint evaluations for the entire optimization procedure than \fullMFA{} does for a single iteration.

Our largest scale experiments for \varMFA{} demonstrate that with current hardware,\footnote{We used a node with considerable compute resources: \varMFA{} was trained utilizing all 64 cores of a single AMD Genoa EPYC 9554 CPU and 4 TB of RAM.} exceptionally large MFA models can be trained efficiently (see \cref{fig:large_scale}). Concretely, for $C=\num{500 000}$ an MFA model with $10.8$ billion parameters can be optimized in less than nine hours.

\subsection{Benchmarking on the Downstream Task of Image Denoising}
\label{sec:denoising}
Negative log-likelihood (NLL) as used in the previous subsections is the canonical measure to evaluate the ability of probabilistic generative models to learn data densities \citep[e.g.,][]{Theis2016}.
It can, however, be very instructive to add evaluations based on other measures that are obtained in specific contexts.

One strategy is to choose one of the many tasks that can be addressed by a trained generative model, e.g., denoising, deblurring, inpainting, outlier detection, compression and many more.
For a probabilistic generative model, any such task is a {\em downstream task} in the sense that a post-processing routine uses the learned data density in order to address the respective task.
Among all possible tasks, denoising is exceptionally widely used, and the task represents a long-standing benchmark of constantly high theoretical and practical interest.
Because of its importance, algorithms are often specialized only for this specific task.
In contrast, any model with focus on data density learning is itself much more general-purpose.
Hence, a benchmark comparing special purpose algorithms to general-purpose approaches such as generative models
can be expected to favor the specialists. Nevertheless, we here choose the specific task of image denoising and a comparison to specialist algorithms developed for this task.
An advantage of using image denoising as downstream benchmark is the ability to compare the effectiveness and efficiency of \varMFA{} to a broad range of other algorithms.
Our benchmarking will thus include comparisons to approaches such as filtering \citep[][]{DabovEtAl2007},
self-supervised deep neural networks \citep[DNNs;][]{BatsonEtAl2019,KrullEtAl2019,QuanEtAl2020,LequyerEtAl2022}, and approaches applying DNNs within probabilistic models \citep[we will use a variational autoencoder,][]{PrakashEtAl2021} or within inverse problem settings \citep[we will use the deep image prior approach,][]{UlyanovEtAl2018}.

Our comparison uses the recent benchmarking protocol proposed by \citet{LequyerEtAl2022}, which assesses a range of popular {\em blind zero-shot denoising} methods for images.
The blind zero-shot setting has recently become very popular \citep[e.g.,][]{LequyerEtAl2022,QingEtAl2025,SalwigEtAl2024,MousaviLucke2025} as it avoids the training on large data sets of `clean' (i.e., non-noisy) images. Such training on potentially large corpora is disadvantageous
because it is often computationally costly, can introduce biases, and hence and can lead to artifacts that are difficult to control \citep[for discussions see][]{LaineEtAl2021,ShocherEtAl2018}. Furthermore, clean data is often simply not available \citep[e.g., high resolution microscopy images, see][for a discussion]{SalwigEtAl2024}.

Following \citet{LequyerEtAl2022}, our evaluation data sets include two natural image data sets (BSD68 and Set12)
as well as a real-world microscopy data set containing inherently noisy images, referred to as {\em Confocal}.
BSD68 and Set12 contain grayscale natural images. Noisy versions of these images are generated by adding Gaussian noise with the standard deviations $\sigma$ as listed in \cref{tab:denoising_results}.
The Confocal data set is a subset of confocal microscopy images from the Fluorescence Microscopy Data set \citep[FMD;][]{ZhangEtAl2019} and contains only images in which the noise emerges from the image recording process (i.e., the noise is not artificially added but emerges naturally). To obtain (pseudo) ground-truth references for the Confocal data set, multiple captures of the same microscopy scene were averaged \citep[for more details, we refer to][]{ZhangEtAl2019}.

To investigate the scalability of the denoising algorithms in both performance and computational cost, we include in addition to the data used by \citet{LequyerEtAl2022}
one high resolution image of the Div2K data set \citep{AgustssonRadu2017}. The used image (with ID `\texttt{0010}') is a colored image of size $2040 \times 1644$ and was selected prior to any experiments. It will be denoted as `Tiger'.
As with BSD68 and Set12, we corrupt the `Tiger' image with synthetic Gaussian noise.
More details on all data sets are given in \cref{appendix:datasets}.

The original benchmark compared a range of DNNs, namely Noise2Self \citep[N2S;][]{BatsonEtAl2019}, Noise2Void \citep[N2V;][]{KrullEtAl2019}, Self2Self \citep[S2S;][]{QuanEtAl2020},
and Noise2Fast \citep[N2F;][]{LequyerEtAl2022} as well as the Deep Image Prior \citep[DIP;][]{UlyanovEtAl2018} approach which uses DNNs within an inverse problem setting.
Here we extend the original benchmark by additionally including the variational autoencoder (VAE)-based approach DivNoising \citep[DivN;][]{PrakashEtAl2021}, Pixel2Pixel \citep[P2P;][]{QingEtAl2025}, and the well-known filter-based method BM3D \citep[][]{DabovEtAl2007}.
Note that the version of DivNoising considered here and BM3D are not \emph{blind} zero-shot because they
require the value of the Gaussian noise as input parameter.
Hence, both algorithms are not applied to the Confocal data set, as the noise input parameter is unknown for these images.

The above extended list of image denoising algorithms is compared to \varMFA{}. For image denoising with \varMFA{}, we apply an elementary post-processing routine that makes use of the data distribution learned by \varMFA{}. Concretely, \varMFA{}
is trained on image patches as data points, and learns a corresponding data density model.
The learned data density is then used to estimate the non-noisy value of each image pixel.
Details on the post-processing routine are given in \cref{appendix:denoising_with_gmms}.
As \varMFA{} can, as a generative model, learn from noisy data (noisy patches in this case), it is well suited for blind zero-shot denoising.

For our comparison, we employed the same default hyperparameters for a given denoising algorithm across all data sets and noise levels
Details on these settings are given in \cref{appendix:algorithms}.
An exception was made for the learning rates of S2S and P2P on the Confocal data set because the default settings failed to produce meaningful results (the output images were uniformly gray).

To apply \varMFA{} to the benchmark data sets, we set the number of mixture components to $C=1000$. For the other hyperparameters, we employed the values previously used in \cref{sec:scalability,sec:largescale}.
Concretely, we set the dimensionality of the hyperplane to $H=5$ and the variational hyperparameters are chosen as $C'=3$ and $G=15$.
The patch size was fixed at $12 \times 12$ across all data sets, a size observed to perform well for image denoising in previous work \citep[cf.][]{DrefsEtAl2022}.

\begin{figure}[htbp]
  {\LARGE $\mathbf{A}$}

  \begin{flushright}
    \includegraphics[width=0.96\textwidth]{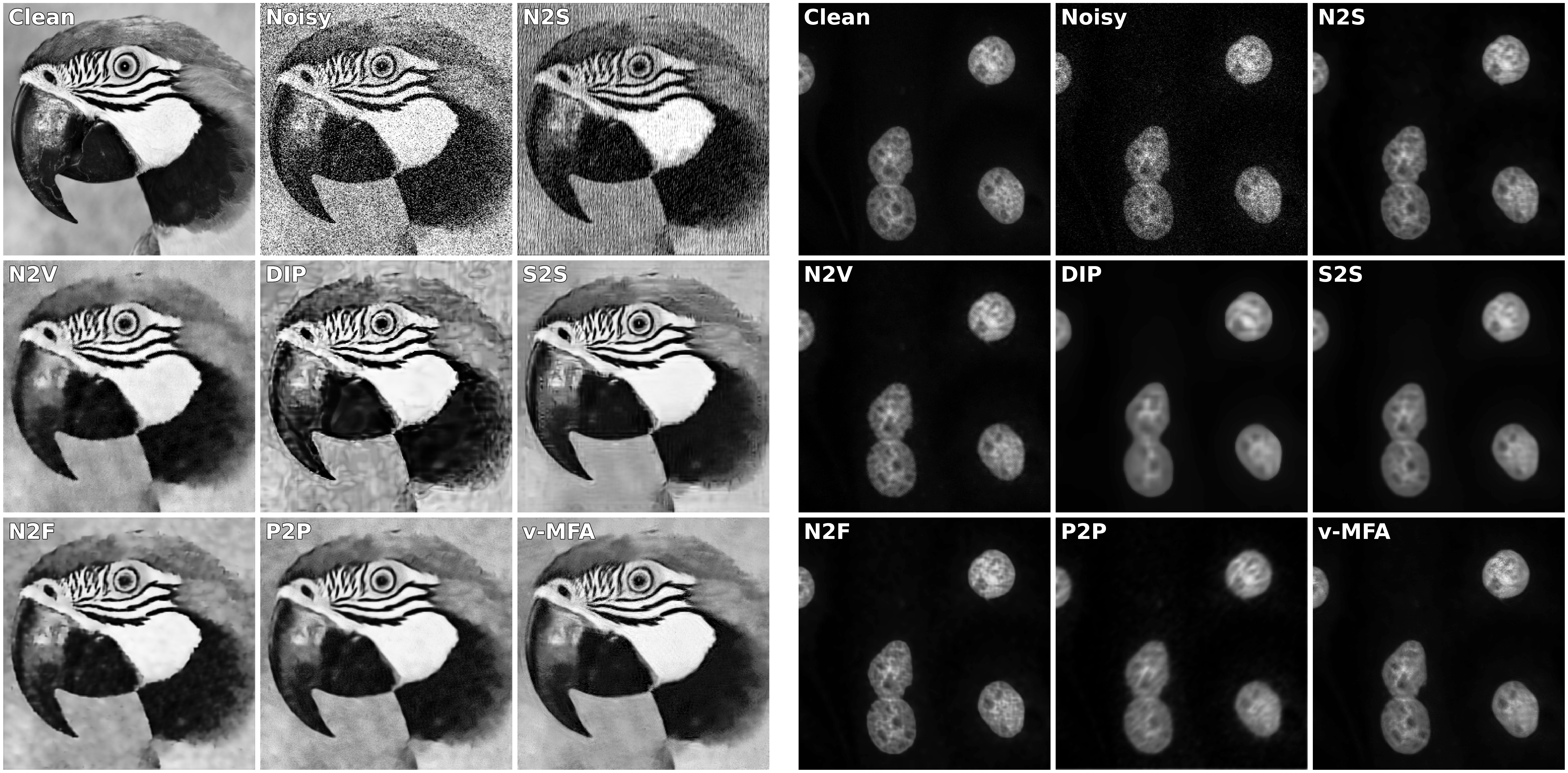}
  \end{flushright}
  \vspace{0.4em}
  \begin{minipage}[c]{0.47\textwidth}
    {\LARGE $\mathbf{B}$} \\
    \includegraphics[width=\textwidth]{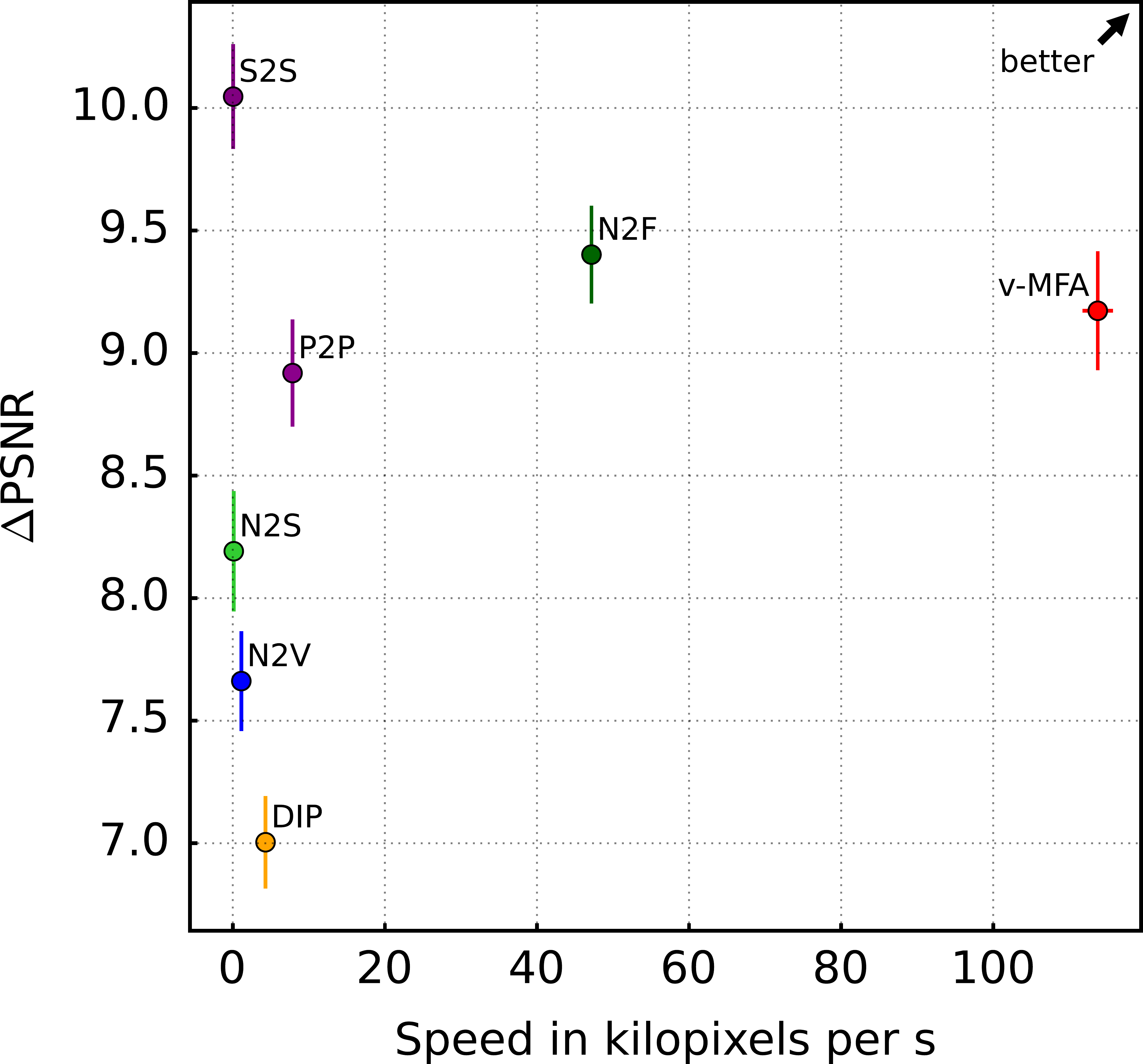}
  \end{minipage}
  \hspace{15pt}
  \begin{minipage}[c]{0.47\textwidth}
    \caption{ Denoising benchmark.
      $\mathbf{A}$ Results of different blind zero-shot denoising algorithms are shown for two images. On the left, denoising of the `parrot' image of the Set12 data set with noise level of $\sigma = 50$ is shown. On the right, denoising results of an image section of BPAE cells from Confocal data set are shown.
      $\mathbf{B}$ Denoising speed in kilopixels per second and PSNR gain is shown for the blind zero-shot denoising algorithms. The plot shows averaged over all data sets and noise levels (colored dots), and error bars denote the corresponding standard error of the mean (SEM).
    } \label{fig:denoising_main}
    \vspace{5em}
  \end{minipage}
\end{figure}

The zero-shot denoising algorithms are individually applied to each noisy image of a given data set, i.e., for each image, both training and inference are conducted exclusively on the image itself.
Training the algorithms on the entire image set would violate the assumptions of the (blind) zero-shot denoising setting.
The performance of each algorithm is evaluated using peak signal-to-noise ratio (PSNR), structural similarity index measure (SSIM) and execution time measured as wall-clock runtime.
For a given data set, we compute the PSNR for each image of the data set and then report the average PSNR. We proceed in the same way for SSIM and wall-clock runtime.
The denoising results are summarized in \cref{tab:denoising_results} and \cref{fig:denoising_main}.
To account for the varying PSNR values of the noisy images across data sets, the PSNR gain ($\Delta$PSNR), defined as the PSNR difference of the denoised and noisy image, is reported in \cref{fig:denoising_main}$\mathbf{A}$.
To normalize for differences in image size across data sets, we report processing speed in kilopixels per second \citep[following][]{LequyerEtAl2022}.
For an image of size $\mathsfit{W} \times \mathsfit{H} \times \mathsfit{C}$, the processing speed $v$
is computed as
\begin{align*}
  v = \frac{\mathsfit{W} \times \mathsfit{H} \times \mathsfit{C} \times 10^{-3}}{t},
\end{align*}
where $\mathsfit{C}$ is the number of color channels (1 for grayscale and 3 for RBG) and $t$ is the total runtime of a given algorithm to denoise this image.

When considering the results in \cref{tab:denoising_results} and \cref{fig:denoising_main}, it may first be observed that in the setting with artificially added noise,
BM3D often outperforms blind zero-shot denoising algorithms if it is provided with the true noise level.
In the standard blind zero-shot setting (when the noise level is unknown), the notable observation for our study is that \varMFA{} almost always ranks among the top three methods in terms of PSNR and SSIM values.
For this setting, the S2S approach exhibits the highest PSNR and SSIM values for all cases with artificially added Gaussian noise.
For the Confocal data set, which features non-artificial noise, we observed \varMFA{} to obtain the highest PSNR and SSIM values.

More central to our study is, however, performance in terms of computational efficiency.
Efficiency in terms of processing speed was also the main focus of the here used benchmark protocol of \citet{LequyerEtAl2022}.
By considering \cref{tab:denoising_results}, \varMFA{} can be observed to be the fastest blind zero-shot method by a considerable margin.
The approach is about twice as fast as N2F (requiring usually half the time), and notably
over 2000$\times$ faster than S2S (see \cref{fig:denoising_main}$\mathbf{B}$).
The only method faster than \varMFA{} is the non-blind classical filter-based BM3D approach.

\begin{table}[tb]
  \centering
  \caption{Quality and runtime of several zero-shot denoising algorithms, measured by PSNR, SSIM and average runtime per image in seconds. The reported errors denote standard deviations across images in the respective dataset.
    The best result within each row among the blind zero-shot algorithms are marked bold, while overall best results (including the non-blind algorithms BM3D and DivN) are underlined.}
  \label{tab:denoising_results}
  \resizebox{\linewidth}{!}{
    \begin{tabular}{l l l r@{$\,\pm\,$}l r@{$\,\pm\,$}l r@{$\,\pm\,$}l r@{$\,\pm\,$}l r@{$\,\pm\,$}l r@{$\,\pm\,$}l r@{$\,\pm\,$}l r@{$\,\pm\,$}l r@{$\,\pm\,$}l}
      \toprule
      \multicolumn{3}{c}{}                         & \multicolumn{4}{c}{non-blind zero-shot} & \multicolumn{14}{c}{blind zero-shot}                                                                                                                                                                                                                                                                                                                                                                                                                                                                                                    \\
      \cmidrule(lr){4-7} \cmidrule(lr){8-21}
      Data set                                     & $\sigma$                                &                                      & \multicolumn{2}{c}{BM3D}                & \multicolumn{2}{c}{DivN}      & \multicolumn{2}{c}{N2S}        & \multicolumn{2}{c}{N2V}      & \multicolumn{2}{c}{DIP}       & \multicolumn{2}{c}{S2S}                              & \multicolumn{2}{c}{N2F}      & \multicolumn{2}{c}{P2P}      & \multicolumn{2}{c}{v-MFA}                                                                                                                                                                         \\
      \midrule
      \multirow{6}{*}{Set12}                       & \multirow{3}{*}{25}                     & PSNR                                 & $29.99$                                 & $1.29$                        & $26.95$                        & $1.63$                       & $27.15$                       & $2.47$                                               & $27.35$                      & $1.55$                       & $27.16$                                  & $1.65$   & \underline{$\boldsymbol{30.03}$} & $1.13$    & $29.06$ & $1.16$ & $28.80$                          & $1.40$  & $29.15$             & $1.40$ \\
                                                   &                                         & SSIM                                 & \underline{$8.52$}                      & $0.30$                        & $7.80$                         & $0.52$                       & $7.37$                        & $1.28$                                               & $7.53$                       & $0.29$                       & $7.74$                                   & $0.53$   & $\boldsymbol{8.49}$              & $0.32$    & $8.22$  & $0.30$ & $8.11$                           & $0.31$  & $8.25$              & $0.33$ \\
                                                   &                                         & Time                                 & \underline{$0.81$}                      & $0.33$                        & $283.20$                       & $49.93$                      & $1139.62$                     & $708.82$                                             & $296.50$                     & $27.94$                      & $37.60$                                  & $1.19$   & $3262.46$                        & $2218.07$ & $3.32$  & $2.05$ & $16.94$                          & $10.93$ & $\boldsymbol{1.37}$ & $0.80$ \\
      \cmidrule(lr){2-21}
                                                   & \multirow{3}{*}{50}                     & PSNR                                 & \underline{$26.75$}                     & $1.27$                        & $23.81$                        & $3.23$                       & $23.66$                       & $3.15$                                               & $24.02$                      & $1.57$                       & $22.75$                                  & $0.91$   & $\boldsymbol{26.54}$             & $1.07$    & $25.89$ & $1.07$ & $25.72$                          & $0.90$  & $26.12$             & $1.19$ \\
                                                   &                                         & SSIM                                 & \underline{$7.68$}                      & $0.41$                        & $6.66$                         & $1.49$                       & $6.17$                        & $1.49$                                               & $5.86$                       & $0.67$                       & $5.48$                                   & $0.55$   & $\boldsymbol{7.42}$              & $0.37$    & $7.24$  & $0.42$ & $6.50$                           & $0.40$  & $7.03$              & $0.39$ \\
                                                   &                                         & Time                                 & \underline{$0.83$}                      & $0.35$                        & $197.02$                       & $69.79$                      & $1140.80$                     & $712.30$                                             & $294.17$                     & $27.80$                      & $37.91$                                  & $1.39$   & $3238.05$                        & $2197.54$ & $3.36$  & $2.86$ & $17.08$                          & $10.95$ & $\boldsymbol{1.23}$ & $0.73$ \\
      \midrule
      \multirow{6}{*}{BSD68}                       & \multirow{3}{*}{25}                     & PSNR                                 & \underline{$28.61$}                     & $2.52$                        & $25.11$                        & $3.25$                       & $26.72$                       & $3.23$                                               & $26.30$                      & $2.67$                       & $25.96$                                  & $2.84$   & $\boldsymbol{28.46}$             & $2.85$    & $28.11$ & $2.44$ & $27.26$                          & $2.89$  & $27.22$             & $3.33$ \\
                                                   &                                         & SSIM                                 & \underline{$8.04$}                      & $0.66$                        & $6.60$                         & $1.67$                       & $7.38$                        & $0.92$                                               & $7.06$                       & $0.70$                       & $7.13$                                   & $0.79$   & $\boldsymbol{7.97}$              & $0.72$    & $7.87$  & $0.61$ & $7.50$                           & $0.59$  & $7.54$              & $0.81$ \\
                                                   &                                         & Time                                 & \underline{$0.79$}                      & $0.02$                        & $224.38$                       & $94.58$                      & $1198.04$                     & $2.60$                                               & $920.73$                     & $535.47$                     & $39.68$                                  & $0.48$   & $3230.29$                        & $56.55$   & $3.67$  & $0.90$ & $20.97$                          & $0.19$  & $\boldsymbol{1.42}$ & $0.32$ \\
      \cmidrule(lr){2-21}
                                                   & \multirow{3}{*}{50}                     & PSNR                                 & $25.67$                                 & $2.60$                        & $20.72$                        & $4.89$                       & $24.49$                       & $2.26$                                               & $23.22$                      & $2.78$                       & $22.28$                                  & $1.62$   & \underline{$\boldsymbol{25.70}$} & $2.48$    & $25.31$ & $2.35$ & $24.84$                          & $2.10$  & $25.30$             & $2.66$ \\
                                                   &                                         & SSIM                                 & \underline{$6.89$}                      & $0.91$                        & $5.14$                         & $2.02$                       & $6.39$                        & $0.80$                                               & $5.16$                       & $0.77$                       & $5.16$                                   & $0.98$   & $\boldsymbol{6.87}$              & $0.86$    & $6.72$  & $0.80$ & $5.91$                           & $0.51$  & $6.51$              & $0.86$ \\
                                                   &                                         & Time                                 & \underline{$0.81$}                      & $0.03$                        & $157.14$                       & $96.25$                      & $1185.39$                     & $2.74$                                               & $935.96$                     & $552.42$                     & $39.99$                                  & $0.83$   & $3222.79$                        & $61.80$   & $3.36$  & $0.97$ & $21.14$                          & $0.33$  & $\boldsymbol{1.40}$ & $0.24$ \\
      \midrule
      \multirow{3}{*}{Confocal}                    & \multirow{3}{*}{--}                     & PSNR                                 & \multicolumn{2}{c}{--}                  & \multicolumn{2}{c}{--}        & $36.20$                        & $3.88$                       & $35.70$                       & $3.53$                                               & $34.53$                      & $3.89$                       & $36.51$                                  & $3.39$   & $36.60$                          & $3.84$    & $34.02$ & $2.33$ & \underline{$\boldsymbol{36.83}$} & $3.54$                                 \\
                                                   &                                         & SSIM                                 & \multicolumn{2}{c}{--}                  & \multicolumn{2}{c}{--}        & $9.26$                         & $0.38$                       & $9.18$                        & $0.45$                                               & $8.94$                       & $0.64$                       & $9.28$                                   & $0.38$   & $9.33$                           & $0.39$    & $9.04$  & $0.48$ & \underline{$\boldsymbol{9.34}$}  & $0.37$                                 \\
                                                   &                                         & Time                                 & \multicolumn{2}{c}{--}                  & \multicolumn{2}{c}{--}        & $1964.20$                      & $1.22$                       & $203.80$                      & $20.56$                                              & $41.93$                      & $3.54$                       & $5884.49$                                & $140.50$ & $8.42$                           & $0.99$    & $30.71$ & $2.24$ & \underline{$\boldsymbol{4.43}$}  & $1.63$                                 \\
      \midrule
      \multirow{6}{*}{\parbox{1cm}{Div2K `Tiger'}} & \multirow{3}{*}{25}                     & PSNR                                 & \multicolumn{2}{c}{\underline{$32.16$}} & \multicolumn{2}{c}{$29.00$}   & \multicolumn{2}{c}{$30.02$}    & \multicolumn{2}{c}{$29.59$}  & \multicolumn{2}{c}{$27.07$}   & \multicolumn{2}{c}{$\boldsymbol{31.95}$}             & \multicolumn{2}{c}{$29.66$}  & \multicolumn{2}{c}{$30.67$}  & \multicolumn{2}{c}{$29.19$}                                                                                                                                                                       \\
                                                   &                                         & SSIM                                 & \multicolumn{2}{c}{\underline{$8.60$}}  & \multicolumn{2}{c}{$7.57$}    & \multicolumn{2}{c}{$7.95$}     & \multicolumn{2}{c}{$7.63$}   & \multicolumn{2}{c}{$6.96$}    & \multicolumn{2}{c}{$\boldsymbol{8.56}$}              & \multicolumn{2}{c}{$7.79$}   & \multicolumn{2}{c}{$8.16$}   & \multicolumn{2}{c}{$7.83$}                                                                                                                                                                        \\
                                                   &                                         & Time                                 & \multicolumn{2}{c}{\underline{$29.46$}} & \multicolumn{2}{c}{$2904.16$} & \multicolumn{2}{c}{$72962.49$} & \multicolumn{2}{c}{$200.57$} & \multicolumn{2}{c}{$1652.69$} & \multicolumn{2}{c}{$89821.74$}                       & \multicolumn{2}{c}{$217.89$} & \multicolumn{2}{c}{$554.87$} & \multicolumn{2}{c}{$\boldsymbol{47.18}$}                                                                                                                                                          \\
      \cmidrule(lr){2-21}
                                                   & \multirow{3}{*}{50}                     & PSNR                                 & \multicolumn{2}{c}{$29.23$}             & \multicolumn{2}{c}{$26.70$}   & \multicolumn{2}{c}{$27.04$}    & \multicolumn{2}{c}{$26.35$}  & \multicolumn{2}{c}{$25.93$}   & \multicolumn{2}{c}{\underline{$\boldsymbol{29.56}$}} & \multicolumn{2}{c}{$27.14$}  & \multicolumn{2}{c}{$28.29$}  & \multicolumn{2}{c}{$28.03$}                                                                                                                                                                       \\
                                                   &                                         & SSIM                                 & \multicolumn{2}{c}{$7.81$}              & \multicolumn{2}{c}{$6.64$}    & \multicolumn{2}{c}{$6.83$}     & \multicolumn{2}{c}{$5.77$}   & \multicolumn{2}{c}{$6.60$}    & \multicolumn{2}{c}{\underline{$\boldsymbol{7.92}$}}  & \multicolumn{2}{c}{$6.92$}   & \multicolumn{2}{c}{$7.22$}   & \multicolumn{2}{c}{$7.41$}                                                                                                                                                                        \\
                                                   &                                         & Time                                 & \multicolumn{2}{c}{\underline{$29.28$}} & \multicolumn{2}{c}{$2773.43$} & \multicolumn{2}{c}{$73193.76$} & \multicolumn{2}{c}{$201.57$} & \multicolumn{2}{c}{$257.96$}  & \multicolumn{2}{c}{$90042.40$}                       & \multicolumn{2}{c}{$217.11$} & \multicolumn{2}{c}{$502.03$} & \multicolumn{2}{c}{$\boldsymbol{38.45}$}                                                                                                                                                          \\
      \bottomrule
    \end{tabular}
  }
\end{table}

\begin{figure}[htbp]
  {\LARGE $\mathbf{A}$}

  \begin{flushright}
    \includegraphics[width=0.96\textwidth]{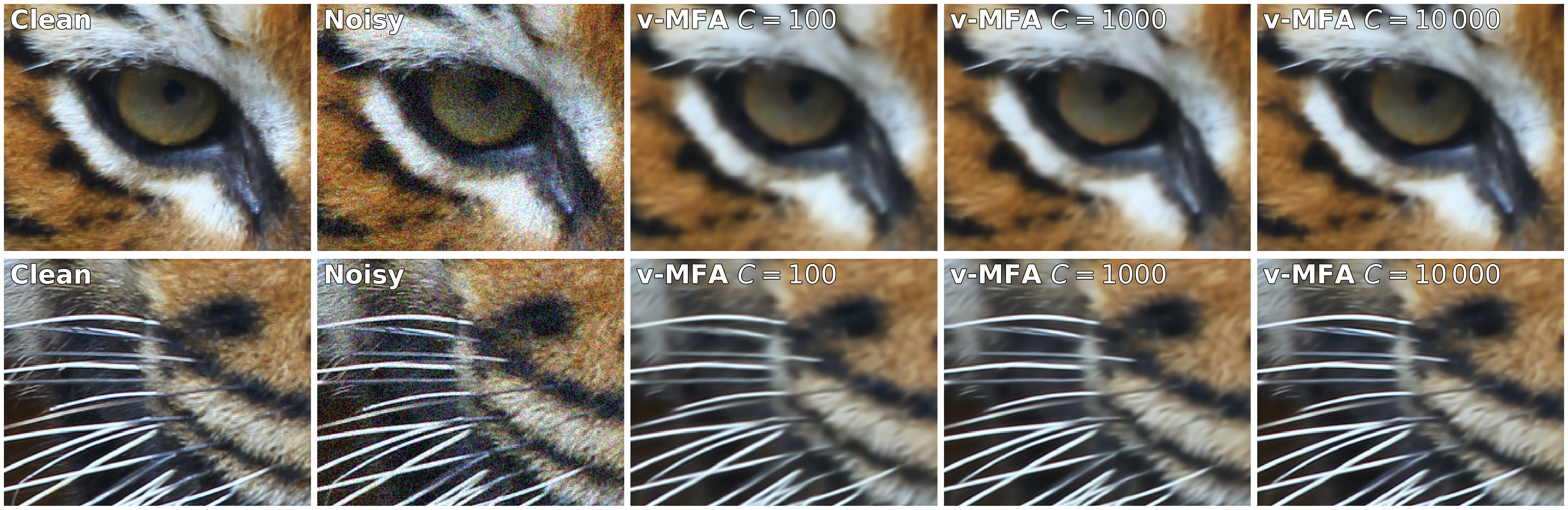}
  \end{flushright}
  \vspace{0.4em}
  \begin{minipage}[c]{0.40\textwidth}
    {\LARGE $\mathbf{B}$} \\
    \includegraphics[width=\textwidth]{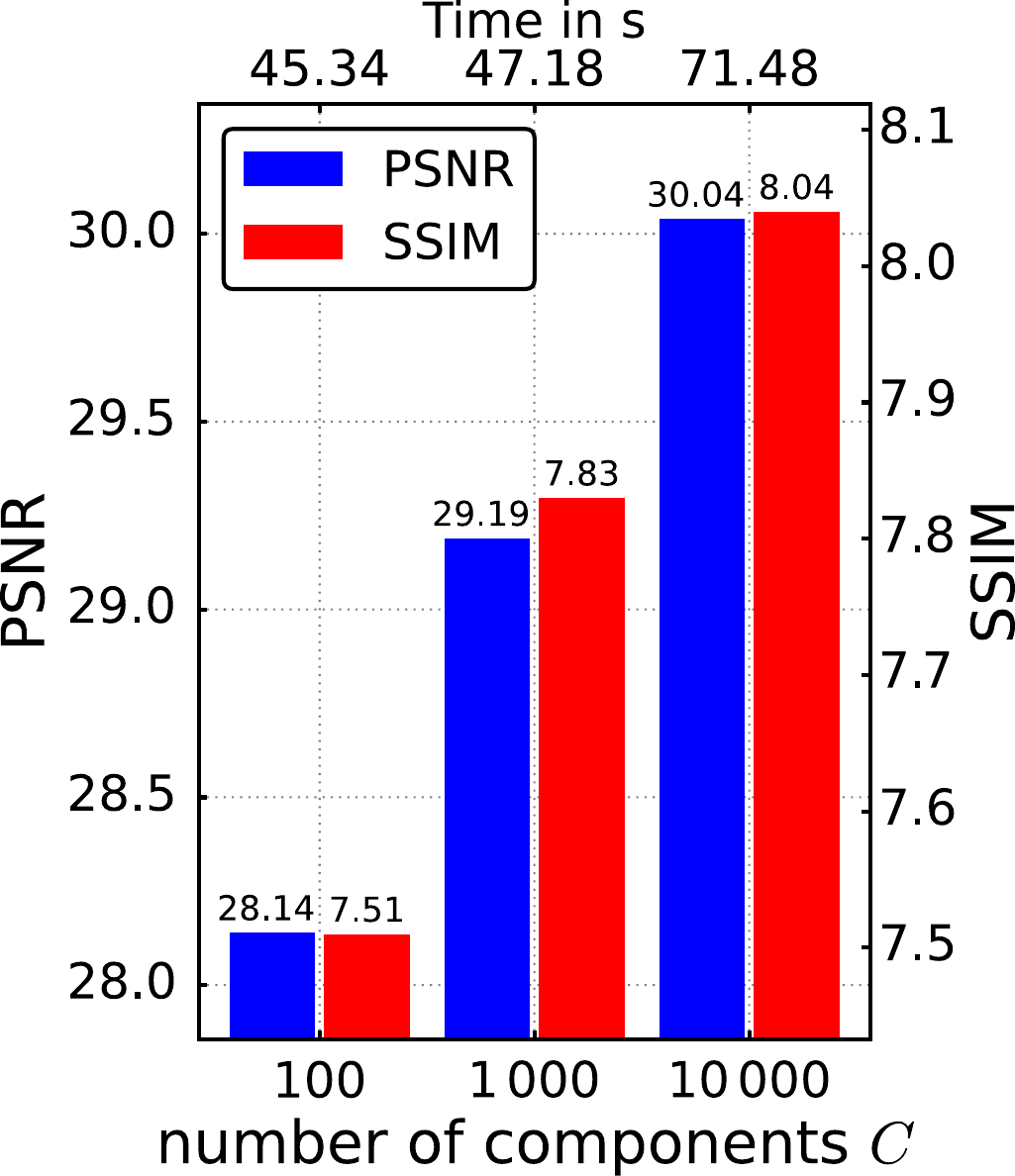}
  \end{minipage}
  \hspace{15pt}
  \begin{minipage}[c]{0.55\textwidth}
    \caption{The effect of scaling \varMFA{} for denoising.
      $\mathbf{A}$ shows a visual comparison for different numbers of components $C$ for two image sections of the Tiger image from Div2K with a noise level of $\sigma = 25$ (zoom in for details).
      $\mathbf{B}$ shows the denoising performance of \varMFA{} with different numbers of components $C$ on the Tiger image, measured by PSNR (left axis) and SSIM (right axis). The top axis indicates the corresponding total runtime in seconds.
    } \label{fig:denoising_scaling-2}
    \vspace{8.5em}
  \end{minipage}
\end{figure}

In \cref{tab:denoising_results} and \cref{fig:denoising_main}, we fixed the number of components to $C = 1000$.
However, the number of components $C$ (i.e., model parameters) is expected to influence the denoising performance of \varMFA{}.
From the high resolution `Tiger' image from Div2K, we can extract about 3.3M patches, which enables us to systematically study the effect of varying $C$.
Specifically, we evaluate $C \in \{\num{100}, \num{1000}, \num{10000}\}$, keeping all other hyperparameters fixed. The corresponding results are shown in \cref{fig:denoising_scaling-2}.
From \cref{fig:denoising_scaling-2}, it is evident that larger MFA representations lead to improved denoising performance in both PSNR and SSIM.
Larger models can better preserve fine structures, with edges appearing sharper, as illustrated in \cref{fig:denoising_scaling-2}$\mathbf{A}$.
At the same time, the runtime increases only marginally, owing to the sublinear scaling of \varMFA{}.
For a tenfold increase in model parameters (from $C=\num{1000}$ to $C=\num{10000}$), the denoising runtime is less than twice as long.

\newpage

In \cref{appendix:further_denoising_results}, we provide additional denoising results, including further visual comparisons,
additional noise levels as well as other GMM-based denoising algorithms.
This includes a comparison of \varMFA{} to \fullMFA{}, as well as a comparison in which the MFA model is replaced by a GMM with diagonal covariances.
Additionally, the effect of the hyperplane dimension $H$ on the denoising performance of \varMFA{} is investigated.

While \fullMFA{} yields marginal improvements in denoising performance, it comes at the cost of very substantially increased runtimes.
The results for the diagonal GMM show that the denoising performance of \varMFA{} is much stronger, which can be attributed to its flexible parametrization of correlations.
Furthermore, the results in \cref{tab:appendix_denosing-different-h} show that \varMFA{} can achieve even higher PSNR and SSIM values than those reported in \cref{tab:denoising_results} when different hyperplane dimensions $H$ are used.
In particular, larger values of $H$ can be beneficial when a sufficiently large number of patches is available.

\FloatBarrier
\section{Discussion}
\label{sec:discussion}

One of the most salient hallmarks of GMMs is their flexible parametrization of correlations within components.
Furthermore, large numbers of components $C$ allow GMMs to approximate potentially
any data density with increasing accuracy \citep[e.g.,][]{ZeeviMeir1997,LiBarron1999,MazyaSchmidt1996}, and approximation quality given $C$
decisively depends on the flexibility of GMMs.
However, the flexibility of general GMMs with large~$C$ comes with a very high computational demand.
For GMMs with flexible covariances we have here addressed the problem of high optimization costs.
Notably, the derived and evaluated algorithm, \varMFA{}, does not reduce runtimes by merely a fixed constant amount or percentage. Instead, and more substantially, we have empirically shown another complexity class for runtime scaling:
as the number of data points $N$ and components $C$ increase, the runtime complexity scales \emph{sublinearly} with $NC$.
Achieving such a sublinear scaling for MFA optimization was the main goal of the research here reported.

Possible future research directions are the treatment of generalizations of MFAs towards non-Gaussian observation spaces including observables of different types \citep{EmtiyazEtAl2010,Banerjee2005}, or extensions towards discrete time-sequence models \citep{Rabiner1989,ZhouCarin2015}.
Furthermore, future work may add other learning strategies \citep[e.g.,][]{ZhaoYu2008,Archambeau2008,ElidanFriedman2012,LinEtAl2019} to the approach presented here; and future work may explore the transferability of sublinear optimization strategies to further well-known algorithms such as different types and combinations of principal and independent component analysis \citep{Hyvarinen2015,DingHe2004}.
A crucial challenge for such future research will always be the derivation of efficient estimation of methods to define truncated distributions; and, furthermore, many models may (in contrast to GMMs and MFAs) not have closed form M-steps.

The concept of truncation, i.e., focussing posteriors by assigning probability mass only to the most relevant latent values, is likely transferable to other types of machine learning models.
Notably, truncated posteriors emerge when applying the concept of mixtures of experts \citep[][]{JacobsEtAl1991} to enhance the efficiency of DNNs. More concretely, gating in a standard routing approach for sparse mixtures of experts \citep[SMoE;][]{ShazeerEtAl2017,FedusEtAl2022} with \texttt{TopK} selection is equivalent to the use of truncated posteriors in a probabilistic formulation of gating \citep[see][and references therein for details]{NguyenNhatEtAl2025}.
Efficient search procedures like developed here for \varMFA{} could in principle enable more efficient routing, i.e., the computation of all router scores for gating could be avoided. Especially for large numbers of experts, sublinear scaling would be a very desirable property. Furthermore, efficient modeling of correlations as used for \varMFA{} (and for MFA in general) could enhance gating by exploiting its probabilistic formulation \citep[][]{NguyenNhatEtAl2025,XuEtAl1994}.

Regarding the here used models themselves, GMMs remain relatively elementary but they are also ubiquitous as the use of Gaussian is very common for standard metric data. Furthermore, non-Gaussian data typically makes the parametrization of correlations much more challenging.
Like GMMs, the here investigated MFA models also still remain relatively elementary data models.
Unlike general GMMs, MFAs can be regarded as additionally assuming the data to lie close to low-dimensional manifolds, which MFAs approximate piecewise by low-dimensional hyperplanes.
Otherwise, MFA models do not impose any additional assumptions or structures beyond GMMs.
Consequently, applications of \varMFA{} to tasks previously addressed using GMMs (or MFAs) should not be expected to show improvements in quality (if the same number of components is used).
However, for applications involving large GMMs, the \varMFA{} algorithm represents a {\em much} more efficient optimization algorithm.
At the same time, and except of the submanifold assumption, the optimized MFA models used by \varMFA{} retain the flexibility of general GMMs in terms of within-component correlations.

Our numerical analysis confirms that the optimization quality achieved by \varMFA{} is very similar to the optimization quality of conventional MFA optimization.
More importantly, however, the analysis also confirms the key advantage of \varMFA{}. Compared to conventional training, its substantially faster optimization times were observed
to result in speed-ups of an order of magnitude and more (see \cref{sec:scalability,sec:quality}). Furthermore, the relative speed-up compared to conventional training {\em increases}
with model size. For very large scales, \varMFA{} is therefore the by far most efficient and likely the only computationally feasible approach (see \cref{sec:largescale}).
As a consequence, \varMFA{} is (to the knowledge of the authors) currently the only approach allowing for the optimization of flexible GMMs with billions of parameters;
and \varMFA{} can perform such an optimization in less than nine hours on one state-of-the-art CPU.

The application of \varMFA{} to image denoising as a downstream task demonstrates that scalability translates into a method that is competitive in terms of denoising performance (see \cref{sec:denoising}). The observed high performance is particularly notable given that MFAs optimized by \varMFA{} are neither specifically designed for nor tuned to the task of image denoising.
Nevertheless, \varMFA{} consistently outperformed several DNN-based blind zero-shot methods across nearly all evaluated settings, especially under high noise conditions. On data with naturally induced noise \citep[the Confocal data set,][]{ZhangEtAl2019}, \varMFA{} even shows the highest performance. More importantly in our context, though, \varMFA{} achieves high denoising accuracy in substantially shorter runtimes than those observed for all compared DNN-based approaches. This makes it particularly well-suited for denoising pipelines where processing speed is critical, such as in real-time imaging applications \mbox{\citep[cf.][]{LequyerEtAl2022}}.

Also for data sets other than image data, processing speed is often crucial. Large amounts of high dimensional data are, for instance, a typical feature of auditory data and data from genetic or proteomic sequencing \citep[e.g.,][]{IHMC2005,ConroyEtAl2023,BennettEtAl2023}. Large-scale mixture models can be used in such context, e.g., for finding clusters, to model low-dimensional data structures or to enable downstream tasks such as error correction or missing data imputation.
For such and other types of data, applications of MFA models can be based on the here reported ability of \varMFA{} to learn data representations at scales that were previously only accessible by another class of machine learning models, namely DNNs.
For image data, the largest such neural network based models are reported to range from hundreds of millions to about one billion parameters (\citealp[][]{Dosovitskiy2021,Fang2023,Wortsman2024}; also compare \citealp{HsuEtAl2021,FreyEtAl2023}, for other domains).
Given that GMMs are universal density approximators, the novel ability to optimize giant GMMs with billions of parameters can make them to a valuable tool for future research.
This is especially relevant considering that large data sets and models are regarded as central in driving the whole field of artificial intelligence.
The interpretability of GMMs and MFAs may thus allow for contributing new insights in studies on inductive biases or on how performance can scale with large model sizes. Furthermore, the
here developed training principles and resulting sublinear scaling may be transferable to other (including deep) models.

\section*{Acknowledgements}

We thank Jan Warnken, Roy Friedman and Yair Weiss for valuable discussions and feedback.
We also thank Leonhard Kohl-Lörting for proofreading the manuscript.

\pagebreak
\appendix

\setcounter{figure}{0}
\setcounter{equation}{0}
\setcounter{table}{0}

\renewcommand*{\thefigure}{A\arabic{figure}}
\renewcommand*{\thetable}{A\arabic{table}}
\renewcommand*{\theequation}{A\arabic{equation}}

\section*{\LARGE Appendix}

\section{Supplementary Information on Methods}
\label{appendix:aintro}
We first provide some details about the method of truncated variational optimization.
Variational distributions do, in general, define a free energy which is a tractable lower bound on the log-likelihood \citep[see, e.g.,][]{NealHinton1998}. For a mixture model with $C$ components, the free energy results from the log-likelihood by applying Jensen's inequality as follows
\begin{align}
  \mathcal{L}(\bx_{1:N}; \bTheta) & = \sum_{n=1}^{N}\log \sum_{c=1}^{C} p(\bxn,c\,|\,\bTheta)
  = \sum_{n}\log\Big( \sum_{c} \qnct \frac{1}{\qnct} p(\bxn,c\,|\,\bTheta)  \Big) \nonumber                                              \\
                                  & \geq \sum_{n}\sum_{c} \qnct \log\left(  \frac{1}{\qnct} p(\bxn, c\,|\, \bTheta)  \right)   \nonumber \\
                                  & = \sum_{n} \Big( \sum_{c} \qnct \log  p(\bxn, c\,|\, \bTheta) \,+\,\HH\left[ q_n \right] \Big),
  \nonumber
\end{align}
where the expression for the MFA model inserted is provided by \cref{eq:FE}.
Truncated distributions are a family of variational distributions of the form given in \cref{eq:qnc}, with variational parameters $\bKK$ and $\tilde{\bTheta}$.
They allow for a reformulation of the free energy after each M-step that takes the form in \cref{eq:FEShort}.
Care has to be taken regarding the variational parameters $\tilde{\bTheta}$.
In general, the parameters can have values different from the model parameters $\bTheta$.
However, it can be shown that the optimal values of $\tilde{\bTheta}$ are equal to $\bTheta$ after each M-step.
See \citet[][]{Lucke2019}, Proposition 3, for a general derivation and see, e.g., \citet[][]{DrefsEtAl2022} and \citet[][]{HirschbergerEtAl2022} for expressions such as \cref{eq:FEShort} used for specific models.
For $\tilde{\bTheta} = \bTheta$, the particularly concise expression of the free energy in \cref{eq:FEShort} can be derived by inserting the truncated distribution in \cref{eq:qnc} into the free energy of \cref{eq:FE}.
As a technical remark, it can be verified \citep[see][Proposition 1]{Lucke2019} that the above derivation of the free energy can be generalized to variational distributions  $\qnct$ with exact (`hard') zeros (which is a property of truncated distributions).

In the following, we provide further details on the method used to avoid inversions of large covariance matrices in \cref{appendix:EfficientEvaluationMFAs}, the derivation of the M-step parameter updates in \cref{appendix:Mstep}, the proof of Proposition 1 in \cref{appendix:PropOne}, additional information on the estimation of component-to-component distances and the KL-divergence in \cref{appendix:comparison_to_KL}, as well as further details on the algorithmic complexity of the variational E-step in \cref{appendix:estep_complexity}.

\subsection{Avoiding Inversion of Large Covariance Matrices in MFAs}
\label{appendix:EfficientEvaluationMFAs}

Optimizing the MFA model using EM or variational EM requires repeatedly evaluating joint probabilities $p(c,\bxn \,|\, \bTheta)$ or, equivalently, log-joints.
Thus, it is essential to assess the efficiency of log-joint evaluations.
Given that
\begin{align}
  \log p(\bxn, c\,|\, \bTheta) & = \log p(c\,|\, \bTheta) + \log p(\bxn\,|\,c, \bTheta) \nonumber                                                                                   \\
                               & = \log \pic - \tfrac{D}{2} \log(2 \pi) - \tfrac{1}{2} \log |\bSigmac| - \tfrac{1}{2} (\bxn - \bmuc)^{\top} \bSigmac^{-1} (\bxn - \bmuc), \nonumber
\end{align}
the crucial aspects include the evaluation of the squared Mahalanobis distance
$(\bxn - \bmuc)^{\top} \bSigmac^{-1} (\bxn - \bmuc)$, which necessitates inverting
$\bSigmac$, and the determinant computation for $\log |\bSigmac|$.
Direct calculations involving $\bSigmac$ become computationally expensive for
high-dimensional data. To enhance computational efficiency, we adopt methods from \citet{Ghahramani1996}, \citet{McLachlanEtAl2003} and \citet{Richardson2018}.
Specifically, for the inversion of $\bSigmac$, we utilize Woodbury's matrix inversion lemma
\begin{align}
  \label{eq:sigma_inversion}
  \bSigmac^{-1} = \bDci - \bDci \bAc \left(\bIdentity +
  \bAcT\bDci\bAc\right)^{-1}\bAcT\bDci = \bDci - \bUc \bVc,
\end{align}
where we define
\begin{align}
  \bUc  \defeq \bDci\bAc			   \in \mathbb{R}^{D\times H} & ,
  \quad
  \bLc  \defeq \bIdentity + \bAcT \bDci \bAc = \bIdentity + \bUcT \bAc \in
  \mathbb{R}^{H\times H},  \nonumber                                                                                      \\
  \bVc  \defeq                                           & \bLci\bAcT\bDci  =  \bLci \bUcT	       \in \mathbb{R}^{H\times
    D}. \nonumber
\end{align}
Given that $\bDc$ is diagonal, its inversion is trivial. Furthermore,
the matrix $\bLc$ is of size $H \times H$ and can therefore be inverted more
efficiently than the $D \times D$ covariance matrix (especially for $H\ll{}D$).
As a quantification of complexity, it is evident from \cref{eq:sigma_inversion} that the squared Mahalanobis
distance, expressed as
\begin{align}
  \bvT \bSigmac^{-1} \bv = \bvT \bDci \bv - (\bvT \bUc) (\bVc \bv), \nonumber
\end{align}
can be evaluated with complexity $\OO(DH)$ for any $\bv \in \RR^{D}$.

Finally, the log-determinant is obtained using the matrix determinant lemma
\begin{align}
  \log |\bSigmac| = \log \left|\bAc \bAcT + \bDc\right| & = \log
  \left|\bIdentity + \bAcT\bDci \bAc\right| + \log \left|\bDc\right| = \log
  \left|\bLc\right| + \sum_{d} \log \sigma^2_{cd}. \nonumber
\end{align}
In conclusion, we can efficiently compute the log-joints for high-dimensional
data such as images, circumventing direct computations involving covariance
matrices of size \mbox{$D \times D$}.

The derivation for the parameter updates in the M-step that maintain efficiency for high~$D$ is provided in \cref{appendix:Mstep}.
Although there exist alternative optimization methods to EM, such as Alternating Expectation Conditional Maximization \citep[AECM,][]{McLachlanEtAl2003} and Expectation Conditional Maximization \citep[ECM,][]{ZhaoYu2008}, we keep in line with the EM approach.
AECM requires two cycles, involving two E-steps to update all model parameters, and the conditional M-step updates in AECM and ECM lead to higher computational complexities in $D$ or $H$.

\subsection{Derivation of the M-step Parameter Updates}
\label{appendix:Mstep}

In this section, we derive the parameter updates for the M-step of the MFA model, specified in \cref{eq:mfa} in the main text, following the derivations outlined in \citet{Ghahramani1996}.
Since our parameter updates deviate from \citet{Ghahramani1996} in certain aspects, e.g., using a variational approach with truncated posteriors or individual diagonal variances $\bD_c$ for each component, we present here the complete derivations for clarity and completeness.

In the following, we always denote the variational parameters $\tilde{\bTheta}$ with a tilde, e.g., $\tilde{\bmu}_c$, where $\bmu_c$ (without a tilde) corresponds to the model parameters $\bTheta$.
Recall that the MFA generative model is given by
\begin{align}
  p(c\,|\,\bTheta)          = \pic, \quad p(\bz\,|\,\bTheta)        = \mathcal{N}(\bz; \boldsymbol{0}, \bIdentity), \quad p(\bx\,|\,c, \bz,\bTheta)  = \mathcal{N}(\bx; \bAc\bz+\bmuc, \bDc). \nonumber
\end{align}
We start here by introducing truncated posteriors as variational distributions given by
\begin{align}
  q(c,\bz; \bxn,\KKn,\tilde{\bTheta}) & = \frac{1}{Z_n} p(c, \bz \,|\,\bxn,
  \tilde{\bTheta})\delta(c\in\KKn)   \nonumber                                                                           \\
                                      & = \frac{1}{Z_n} p(c \,|\,\bxn,\tilde{\bTheta}) p(\bz\,|\,c,\bxn,\tilde{\bTheta})
  \delta(c\in\KKn) \nonumber                                                                                             \\
                                      & = \qnct \qnczt,\label{eq:tv_q}
\end{align}
and define $\qnct  \defeq \frac{1}{Z_n} p(c \,|\,\bxn,\tilde{\bTheta}) \delta(c\in\KKn)$ and $\qnczt  \defeq p(\bz\,|\,c,\bxn,\tilde{\bTheta})$. $Z_n$ is a normalization constant given by
\begin{align}
  Z_n = \! \sum_{\tilde{c} \in \KKn} \int p(\ct, \bz \,|\,\bxn,
  \tilde{\bTheta}) d\bz = \! \sum_{\tilde{c} \in \KKn} p(c \,|\,\bxn,\tilde{\bTheta}) \int
  p(\bz\,|\,c,\bxn,\tilde{\bTheta}) d\bz = \sum_{\tilde{c} \in \KKn} p(c \,|\,\bxn,\tilde{\bTheta}). \nonumber
\end{align}
The distribution $\qnct$ mirrors \cref{eq:qnc} in the main text, consistent
with previous work employing truncated variational approximations for mixture
models \citep[e.g.,][]{HirschbergerEtAl2022,ExarchakisEtAl2022}.
We recognize that $\qnczt = p(\bz\,|\,c,\bxn,\tilde{\bTheta})$ conforms to a
Gaussian distribution $\mathcal{N}(\bz; \tilde{\bmu}_z, \tilde{\bSigma}_z)$, with
\begin{align}
  \label{eq:z_gauss_params}
  \tilde{\bSigma}_z  = \left(\bIdentity + \tbAcT \tbDci \tbAc \right)^{-1} = \tbLci,\quad
  \tilde{\bmu}_z     = \tilde{\bSigma}_z \tbAcT \tbDci (\bxn - \tilde{\bmu}_c) = \tbVc (\bxn - \tilde{\bmu}_c).
\end{align}
In the following, we introduce the variational free energy and reformulate it into a more convenient form for deriving the parameter update equations.
This free energy is given by
\begin{align}
  \FF(\bx_{1:N}; \bKK, \tilde{\bTheta}, \bTheta)
  = \sum_{n=1}^{N} \sum_{c=1}^{C}\int_{\RR^{H}} q(c,\bz\,|\,\bxn,\tilde{\bTheta})
  \log p(c, \bz, \bxn\,|\,\bTheta) \,d\bz + \mathcal{H}\left[q\right], \nonumber
\end{align}
where $\mathcal{H}\left[ q \right] = - \sum_n \mathbb{E}_{q}[\log q(c,\bz\,|\,\bxn,\tilde{\bTheta})]$ denotes the Shannon entropy.
Substituting \cref{eq:tv_q} into the free energy yields
\begin{align}
  \FF(\bx_{1:N}; \bKK, \tilde{\bTheta}, \bTheta)
   & =  \sum_{n,c} \qnct \int \qnczt \log p(c, \bz,
  \bxn\,|\,\bTheta) \,d\bz  + \mathcal{H}\left[q\right] \nonumber
  \\
   & = \sum_{n,c} \qnct \ \Ez{\log p(c, \bz, \bxn\,|\,\bTheta)} + \mathcal{H}\left[q\right]. \nonumber
\end{align}
The expectation $\Ez{\log p(c, \bz, \bxn\,|\,\bTheta)}$ can be expressed as
\begin{align}
   & \Ez{\log p(c, \bz, \bxn\,|\,\bTheta)} = \Ez{\log p(c\,|\,\bTheta) + \log
    p(\bz\,|\,\bTheta) + \log p(\bxn\,|\,c, \bz, \bTheta)} \nonumber
  \\
   & \qquad = \log \pic - \tfrac{D}{2} \log(2 \pi) + \tfrac{1}{2} \log |\bDci|
  - \tfrac{1}{2} \Ez{(\bxn - \bmuc - \bAc \bz)^{\top} \bDci (\bxn - \bmuc -
    \bAc
  \bz)} \nonumber                                                              \\
   & \qquad = \log \pic - \tfrac{D}{2} \log(2 \pi) + \tfrac{1}{2} \log |\bDci|
  - \tfrac{1}{2} \Ez{(\bxn - \hbAc \hbz)^{\top} \bDci (\bxn - \hbAc \hbz)}, \label{eq:expectation_logjoint}
\end{align}
where we combined $\bAc$ and $\bmuc$ by defining
\begin{align}
  \hbAc \defeq   \begin{bmatrix}
                   \bAc & \bmuc
                 \end{bmatrix},
  \qquad   \hbz \defeq   \begin{bmatrix}
                           \bz \\
                           1
                         \end{bmatrix}, \nonumber
\end{align}
as we have $\hbAc \hbz = \bAc \bz + \bmuc$.
By making use of the symmetry of $\bDci$, the cyclic property of the trace, and the
linearity of $\Ez{\cdot}$, the expectation in \cref{eq:expectation_logjoint} reformulates to
\begin{align}
  \Ez{(\bxn - \hbAc \hbz)^{\top} \bDci (\bxn - \hbAc \hbz)}
  = & \  \bxnT \bDci \bxn - 2 \bxnT \bDci \hbAc \Ez{\hbz} \nonumber \\
    & + \trace
  \left( \hbAcT \bDci \hbAc \Ez{\hbz \hbzT}\right), \nonumber
\end{align}

Inserting the aforementioned results into the free energy yields
\begin{align}
  \FF(\bx_{1:N}; \bKK, \tilde{\bTheta}, \bTheta)
   & = \sum_{n,c} \qnct	\Bigl(  \log \pic -\tfrac{D}{2} \log(2 \pi) + \tfrac{1}{2} \log
  |\bDci|
  - \tfrac{1}{2} \bxnT \bDci \bxn  \nonumber
  \\
   & + \bxnT \bDci \hbAc \hbmuz - \tfrac{1}{2} \trace \left( \hbAcT \bDci \hbAc \hEzz\right)
  \Bigl)
  + \ \mathcal{H}\left[ q \right], \label{eq:final_F}
\end{align}
where the expectation values $\Ez{\cdot}$ are given by
\begin{align}
  \hbmuz  = \begin{bmatrix}
              \tilde{\bmu}_z \\
              1
            \end{bmatrix}, \quad
  \hEzz   = \begin{bmatrix}
              \tilde{\bSigma}_z + \tilde{\bmu}_z \tilde{\bmu}_z^{\top} & \tilde{\bmu}_z \\
              \tilde{\bmu}_z^{\top}                                    & 1
            \end{bmatrix}. \nonumber
\end{align}

Based on this expression of the free energy, we derive the update equations by equating the partial derivatives of \cref{eq:final_F} w.r.t.\ all model parameters $\bTheta$ to zero, while keeping the variational parameters $\tilde{\bTheta}$ fixed.

\myparagraph{Updates of mixing proportions} Considering the constraint $\sum_{c} \pi_c = 1$, the mixing proportions
are determined by the well-known expression
\begin{align}
  \pi_c & = \frac{N_c}{N} \quad \text{with} \quad N_c = \sum_{n} \qnct. \nonumber
\end{align}

\myparagraph{Updates of means and factor loadings} For the derivative w.r.t.\ $\hbAc$ we obtain
\begin{align}
  \boldsymbol{0} = \dv{\FF}{\hbAc} = & \sum_{n} \qnct \bDci \bxn
  \hbmuzT - \sum_{n} \qnct \bDci \hbAc \hEzz, \nonumber
\end{align}
resulting in
\begin{alignat}{2}
                                              &  & \sum_{n} \qnct \bxn \hbmuzT & = \hbAc \sum_{n} \qnct \hEzz \nonumber \\
  \hspace{1cm} \Leftrightarrow   \hspace{2cm} &  & \bYc                        & = \hbAc\bEc     \nonumber              \\
  \hspace{1cm} \Leftrightarrow  \hspace{2cm}  &  & \hbAc                       & = \bYc \bEc^{-1}. \hspace{8cm}
  \label{eq:update_Ac}
\end{alignat}
where we introduced the following two matrices:
\begin{align}
  \bEc  = \sum_{n} \qnct \hEzz,        \quad
  \bYc  = \sum_{n} \qnct \bxn \hbmuzT. \nonumber
\end{align}
The factor loading matrix $\bAc$ is obtained from the first $H$ columns of $\hbAc$, while the mean $\bmu_c$ is given by the last column of $\hbAc$.

\myparagraph{Update of variance} Taking derivatives w.r.t.\ $\bDci$, we derive
\begin{align}
  \boldsymbol{0} & = \dv{\FF}{\bDci} = \frac{1}{2} \sum_{n} \qnct
  \left( \bDc - \bxn \bxnT + 2 \bxn \hbmuzT \hbAcT - \hbAc \hEzz  \hbAcT
  \right). \nonumber
\end{align}
It follows that
\begin{align}
  \bDc N_c = & \sum_{n} \qnct \bxn \bxnT - 2 \left(\sum_{n} \qnct
  \bxn \hbmuzT \right) \hbAcT \nonumber                                         \\
             & + \hbAc \left(\sum_{n} \qnct  \hEzz \right)
  \hbAcT  \nonumber                                                             \\
  =          & \sum_{n} \qnct \bxn \bxnT - 2 \bYc \hbAcT + \hbAc \bEc \nonumber
  \hbAcT.
\end{align}
Substituting $\bYc = \hbAc \bEc$ into the last term, we obtain
\begin{align}
  \bDc N_c & = \sum_{n} \qnct \bxn \bxnT - \bYc \hbAcT. \nonumber
\end{align}
Finally, by applying the diagonal constraint, each diagonal element $\sigma^2_{cd}$ of $\bDc$ can be computed as follows
\begin{align}
  \sigma^2_{cd} & = \frac{1}{N_c} \diag\left( \sum_{n} \qnct
  \bxn \bxnT - \bYc \hbAcT \right)_d \nonumber
  \\
                & = \frac{1}{N_c} \left( \sum_{n} \qnct \ x_{nd}^2 - \sum_h \left(\bYc \odot \hbAc\right)_{d,h} \right), \nonumber
\end{align}
where all non-diagonal elements are set to zero and $\hbAc$ are the updated parameters given by \cref{eq:update_Ac}.
Here, $\diag(\cdot)$ denotes a vector constructed by the main diagonal of a
given matrix and $\odot$ denotes element-wise multiplication.

Given that $\qnct$ is zero for $c \not\in \KKn$, the computational efficiency of the above equations can be improved by summing solely over non-zero $\qnct$, as elaborated in \cref{sec:varEM} in the main text.

\subsection{Relation Between Likelihood, Free Energy and Posterior Mass}
\label{appendix:error_bound}

Alongside contributions on improved parameter optimization, GMMs have also been studied theoretically, and notably regarding convergence guarantees.
The recovery of ground truth parameters, e.g., ground truth component centers, is important for the reliability of clustering results when using GMMs.
In contrast to most theoretical studies,
the MFA models we here consider allow for arbitrary covariance structure along low dimensional hyperplanes. Furthermore, the \varMFA{} optimization
considers modeling of potentially arbitrary data sets in $\RR^D$.
The setting considered here, therefore, makes it difficult to obtain rigorous theoretical results, e.g., on convergence guarantees.
For more controlled settings, far-reaching theoretical insights on the optimization of GMMs and in particular on EM-based algorithms for GMMs can be obtained.
For instance, works by \citet[][]{XuEtAl2016}, \citet[][]{DaskalakisEtAl2017} and \citet[][]{XuEtAl2024} even provide global convergence guarantees for EM-based GMM optimization.
However, assumptions such as GMMs constrained to two components \citep[][]{XuEtAl2016,DaskalakisEtAl2017} are required or strong assumptions such as Gaussianity of the data distribution  \citep[][]{XuEtAl2024} are used.
But also results on local convergence \citep[e.g.,][]{KwonCaramanis2020,SegolNadler2021} require significant assumptions.
Essentially all approaches (for global and local convergence) assume diagonal covariance matrices per component (e.g., all those just mentioned).
Furthermore, assumptions on the separation of components in the data are often central to obtain convergence guarantees \citep[e.g.,][and many more]{RegevVijayaraghavan2017,DiakonikolasEtAl2018,LiuLi2022}.
Studies of both global and local convergence usually consider the properties of either conventional EM or gradient EM based optimization, which is evaluating full posteriors for all data points or batches. Variational approximations differ from conventional EM or gradient EM, and it is less clear how variational EM can be treated (even in the case when taking aside the other assumptions that are not fulfilled in our setting).

In this context, it may, however, be useful to study in some more detail the relation between the log-likelihood objective for GMMs or MFAs (see \crefnobrackets{eq:loglikelihood}) and the here used variational objective in \cref{eq:FE}. It is, for instance, possible to exploit the specific form of the free energy objective that results from truncated variational distributions (cf. \crefnobrackets{eq:FEShort}).
Using \cref{eq:FEShort} (also see beginning of \cref{appendix:aintro}), we can derive the following relation for the difference between $\mathcal{L}(\bx_{1:N}; \bTheta)$ in \cref{eq:loglikelihood} and $\FF(\bKK,\bTheta)$  in \cref{eq:FE}:
\begin{align}
  \mathcal{L}(\bx_{1:N}; \bTheta) - & \FF(\bKK,\bTheta)  = \sum_{n} \log\Big( \sum_c p(c, \bxn\,|\,\bTheta) \Big) - \sum_{n} \log\!\Big(\sum_{\crampedclap{\;\; c\in\KKn}} p(c,\bxn\,|\,\bTheta) \Big) \nonumber \\
                                    & = \sum_{n} \log\left( \frac{\sum_c p(c, \bxn\,|\,\bTheta)} {\sum_{ c\in\KKn} p(c,\bxn\,|\,\bTheta)} \right)   \nonumber
  = \sum_{n} \log\left( 1 + \frac{\sum_{ c\notin\KKn} p(c, \bxn\,|\,\bTheta)} {\sum_{ c\in\KKn} p(c,\bxn\,|\,\bTheta)}\right).
\end{align}
The ratios of the sum of joint probabilities can be rewritten as a ratio of a sum of posterior probabilities. If we furthermore use $\log(1 + x)\leq{}x$, we obtain
\begin{align}
  \mathcal{L}(\bx_{1:N}; \bTheta) - \FF(\bKK,\bTheta) & = \sum_{n} \log\Big( 1 + \frac{\sum_{ c\notin\KKn} p(c\,|\,\bxn,\bTheta)} {\sum_{ c\in\KKn} p(c\,|\,\bxn,\bTheta)}\Big) \leq{} \sum_{n} \frac{\sum_{ c\notin\KKn} p(c\,|\,\bxn,\bTheta)} {\sum_{ c\in\KKn} p(c\,|\,\bxn,\bTheta)}. \nonumber
\end{align}
We, therefore, obtain an upper bound of $\mathcal{L}(\bx_{1:N}; \bTheta)$ in addition to the lower bound $\FF(\bKK,\bTheta)$. When additionally normalized by the number of data points $N$, we obtain
\begin{align}
  \frac{1}{N}\FF(\bKK,\bTheta) \leq{}  \frac{1}{N}\mathcal{L}(\bx_{1:N}; \bTheta)  \leq  \frac{1}{N}\FF(\bKK,\bTheta) + \frac{1}{N} \sum_{n} \frac{\sum_{ c\notin\KKn} p(c\,|\,\bxn,\bTheta)} {\sum_{ c\in\KKn} p(c\,|\,\bxn,\bTheta)}.
  \label{eq:ll_bounds}
\end{align}
The expression shows that the average of the ratio between non-captured vs.\,captured posterior mass directly determines how close the variational objective is to the log-likelihood. For well separated components, for instance, there is for each $n$ one component $c$ that will dominate the posterior values close to the global optimum.
If the $\KKn$ sets contain the corresponding component, the average ratio will be close to zero and both bounds will be very tight. But also for any sufficiently low numbers of dominating components, tight bounds are obtained. The bounds in \cref{eq:ll_bounds} may thus be used to directly link separability of components (or groups of components) to the relation between the variational objective and the studies (such as those discussed above) using conventional EM for the log-likelihood objective.

\subsection{Proof of Proposition 1}
\label{appendix:PropOne}

Let $\bKK$ be a given collection of index sets $\KK_{1:N}$. We now choose an arbitrary $n$ and replace an index $c\in\KKn$ by the index $\tilde{c}\not\in\KKn$. We denote the resulting index set by $\tKKn$, and the
collection of index sets $\KK_{1:N}$ with $\KKn$ replaced by $\tKKn$ we denote by $\tilde{\bKK}$.
Now, we obtain
\begin{alignat*}{2}
                  &  & \FF(\tilde{\bKK},\bTheta) \quad                                                                                                         & > \quad \FF(\bKK,\bTheta)                                                                                    \\
  \Leftrightarrow &  & \qquad \qquad \qquad \sum_{\substack{n'=1                                                                                                                                                                                                              \\ n'\neq{}n}}^{N} \log\!\Big(\sum_{\crampedclap{\;\; c'\in\KKnPrime}} p(c',\bx_{n'}\,|\,\bTheta) \Big)\,&+\,\log\!\Big(\sum_{\crampedclap{\;\; c'\in\tKKn}} p(c',\bxn\,|\,\bTheta) \Big) \\
                  &  & > \quad \sum_{\substack{n'=1                                                                                                                                                                                                                           \\ n'\neq{}n}}^{N} \log\!\Big(\sum_{\crampedclap{\;\; c'\in\KKnPrime}} p(c',\bx_{n'}\,|\,\bTheta) \Big)\,&+\,\log\!\Big(\sum_{\crampedclap{\;\; c'\in\KKn }}\ p(c',\bxn\,|\,\bTheta)  \Big)                                                   \\
  \Leftrightarrow &  & \log\!\Big(\sum_{\crampedclap{\;\; c'\in\tKKn}} p(c',\bxn\,|\,\bTheta) \Big)                                       \quad                & >  \quad\log\!\Big(\sum_{\crampedclap{\;\; c'\in\KKn }}\ p(c',\bxn\,|\,\bTheta) \Big)                        \\
  \Leftrightarrow &  & \sum_{\crampedclap{\;\; c'\in\tKKn\setminus\{\tilde{c}\}}}\ p(c',\bxn\,|\,\bTheta) \,+\, p(\tilde{c},\bxn\,|\,\bTheta)    \quad         & > \quad\sum_{\crampedclap{\;\; c'\in\KKn\setminus\{c\}}}\ p(c',\bxn\,|\,\bTheta) \,+\, p(c,\bxn\,|\,\bTheta) \\
  \Leftrightarrow &  & p(\tilde{c},\bxn\,|\,\bTheta)                                                                                                     \quad & >  \quad p(c,\bxn\,|\,\bTheta),
\end{alignat*}
where we have used strict concavity of the logarithm and $\tKKn\setminus\{\tilde{c}\} = \KKn\setminus\{c\}$.

With the same argumentation, the free energy decreases if $p(\ct,\bxn\,|\,\bTheta)<p(c,\bxn\,|\,\bTheta)$. \\
For $p(\ct,\bxn\,|\,\bTheta)=p(c,\bxn\,|\,\bTheta)$ the free
energy trivially remains unchanged. \\ \makebox[\linewidth][r]{\hspace*{1cm}$\blacksquare$}

\subsection{Estimation of Component-to-Component Distances and KL-Divergence}
\label{appendix:comparison_to_KL}

In this section, we further elaborate on the derivation of the KL-divergence approximation $\rel$, which we use to find components $\ct$ similar to component $c$ (\crefnobrackets{eq:deltaF} in the main text).
Afterwards, we compare $\rel$ to a previously suggested distance estimate in \cref{appendix:comparison_approx_analytical}.

We start by considering idealized assumptions to derive $\rel$ as an approximation of the KL-divergence. Importantly, however,
$\rel$ remains a valid expression for the estimation of similarity of components $\ct$ to $c$ also if the idealized
assumptions are not fulfilled (which we will elaborate on further below). The numerical experiments in \cref{sec:numerical_experiments} in the main text verify the validity of $\rel$
through the  effective and efficient performance of the v-MFA algorithm.

Let us first assume that v-MFA has already converged to a large extent, i.e.,
we assume that the GMM parameters have already reached values  that accurately represent the data components, and that the
search spaces $\GGn$ include the most likely components for their data points $\bxn$. Furthermore, we assume that the data is appropriately modeled by a GMM (i.e., approximately Gaussian components), and that the components are well separated.
Under these idealized conditions, we can assume that $\rel$ given by \cref{eq:deltaF} in the main text can approximate the KL-divergence relatively well for any components $\ct$ similar to $c$.

To illustrate this, we first consider the finite sample approximation of the KL-divergence (\crefnobrackets{eq:kldivergence} in the main text) using $M$ samples of $p(\bx\,|\,c, \bTheta)$, which becomes exact for $M\rightarrow\infty$:
\begin{align}
  \DKL \approx & \frac{1}{M} \sum_{m=1}^M \log \frac{p(\bxm\,|\,c, \bTheta)}{p(\bxm\,|\,\ct, \bTheta)}\,,\quad \text{where\ \ }\ \bxm\sim{}p(\bx\,|\,c, \bTheta). \label{eq:appRELOne}
\end{align}
Secondly, for well separated components, model parameters close to convergence and data that is well-modeled by Gaussian components, the
partitions $\II_c$ of \cref{eq:IIPartition_hard} in the main text will contain data points close to those that would be generated by $p(\bx\,|\,c, \bTheta)$.
Therefore, we can further approximate \cref{eq:appRELOne} by
\begin{align}
  \DKL \approx & \frac{1}{|\II_c|} \sum_{n\in\II_c} \log \frac{p(\bxn\,|\,c, \bTheta)}{p(\bxn\,|\,\ct, \bTheta)}\,.  \nonumber
\end{align}
Finally, we introduce the condition $\delta(\ct \in \GGn)$, which ensures that only components $\ct$ within the search spaces $\GGn$ are considered.
The condition is important for computational efficiency, as it avoids additional computations of $p(\bxn\,|\, \ct, \bTheta)$ that are not performed during the partial E-step.
A disadvantage is that the approximation quality may decrease as potentially fewer data points $\bxn$ are considered for the approximation.
However, if the algorithm has sufficiently converged, the components $\ct$ in the search spaces $\GGn$ of all data points with $n \in \II_c$ will be almost the same.
In that case, $\delta(\ct \in \GGn)$ will nearly always be equal to one, and we obtain
\begin{align}
  \DKL \,\approx\, & \frac{1}{N_{c\ct}} \sum_{n\in\II_c} \log \frac{p(\bxn\,|\,c, \bTheta)}{p(\bxn\,|\,\ct, \bTheta)} \delta(\ct \in \GGn) = \rel\,, \label{eq:appRELThree}
  \\
                   & \text{where} \quad N_{c\ct} = \sum_{n\in\II_c} \delta(\ct \in \GGn) \approx \sum_{n\in\II_c} 1 = |\II_c|. \nonumber
\end{align}
For sufficiently similar components $\ct$ in the sense of KL-divergences, the estimate $\rel$ can therefore provide an accurate approximation of the KL-divergence. However, as the components $\ct$ become increasingly dissimilar, the uncertainty regarding the accuracy of this approximation increases.

On the other hand, the estimate $\rel$ can not be expected to approximate the KL-divergence accurately if the components are strongly overlapping.
In such cases, the sets $\II_c$ may not represent their respective components well,
as each data point can only be assigned to a single component.
However, $\rel$ will by construction still result in small values as long as model parameters and search spaces $\GGn$ have sufficiently converged. In an extreme case of $p(\bxn\,|\,\ct, \bTheta)$ being almost equal to $p(\bxn\,|\,c, \bTheta)$, the summands in \cref{eq:appRELThree} will be close to zero.
As a consequence, similar components in the KL-divergence sense
will also be similar in a ranking based on $\rel$, although values of $\rel$ may differ from the values of the KL-divergences.

Finally, for components $\ct$ that are \emph{very} irrelevant to $c$, we set the values of $\rel$ to infinity, which would also not match the values of the KL-divergence. However, irrelevant components in the KL-divergence sense are anyway disregarded for the definition of sets $g_c$ in \cref{eq:GUpdate} in the main text.

To summarize, $\rel$ serves as an approximation for the KL-divergence between $p(\bxn\,|\,c, \bTheta)$ and $p(\bxn\,|\,\ct, \bTheta)$, which can be quite coarse.
However, as the values of $\rel$ are only used for ranking the similarity of components, coarse estimates are sufficient.
Most importantly, this approximation avoids the $\OO(C^2DH)$ scaling of computing the exact KL-divergences of all pairwise components.
Consequently, it preserves the overall computational complexity of the variational EM algorithm, which is critical for ensuring scalability w.r.t.\ the number of components (cf. \cref{alg:estep_complexity}).

\subsubsection{Comparison to Variational E-Steps of Previous GMM Optimizations}
\label{appendix:comparison_approx_analytical}

Previous work using variational E-steps \citep{HirschbergerEtAl2022,ExarchakisEtAl2022} has not considered GMMs with intra-component correlations.
Estimation of component-to-component distances has, however, been used before by \citet{HirschbergerEtAl2022}, who applied the estimate
\begin{align}
  \quad d^2_{c\ct} & := - \frac{1}{N_{c\ct}} \sum_{n\in\II_c} \log p(\ct, \bxn\,|\, \bTheta) \delta(\ct \in \GGn), \label{eq:euclidean_distance}
\end{align}
where $N_{c\ct}$ is defined as in \cref{eq:deltaF} in the main text. The estimate $d^2_{c\ct}$ was then used as a ranking to define sets $g_c$ analogously to \cref{eq:GUpdate} in the main text.

For isotropic components with equal variances and equal mixing proportions per component, it was argued that $d^2_{c\ct}$ will ultimately correspond to the Euclidean component-to-component distance (and $d^2_{c\ct}$ was originally defined for this isotropic case).
More concretely, it was shown (Appendix E of \citealp{HirschbergerEtAl2022}) that $d^2_{c\ct}$ results in approximately the same distance ranking of component pairs $(c,\ct)$ as the Euclidean component-to-component distance $||\bmu_c-\bmu_{\ct}||$. For diagonal covariance matrices, $d^2_{c\ct}$ was also related to KL-divergences.

From the perspective of the here used approach in \cref{eq:deltaF} in the main text, however, \cref{eq:euclidean_distance} can be considered as ignoring $p(\bxn\,|\,c, \bTheta)$ in the KL-divergence (\crefnobrackets{eq:kldivergence} in the main text). A further difference is a different consideration of the mixing proportions.
Estimate \cref{eq:euclidean_distance} has been shown to perform well for isotropic and diagonal components \citep{HirschbergerEtAl2022}, for which it was originally derived.
Here, our estimate is directly derived from KL-divergences, which implicitly accounts for different intra-component correlations.
Consequently, the more general derivation of $\rel$ (used for v-MFA) is a better match to these intra-component correlations, and hence can be expected to result in a more efficient learning algorithm.
If and how much the more general estimate \cref{eq:deltaF} in the main text is improving learning remains to be numerically investigated, which we provide in \cref{appendix:comparison_approx_numerical}.

\subsection{Algorithmic Complexity of the Variational E-step}
\label{appendix:estep_complexity}

To determine the algorithmic complexity of \cref{alg:estep} in the main text, we analyze each of its four blocks, assuming $N > C$. \Cref{alg:estep_complexity} outlines the steps in \cref{alg:estep} in the main text along with the complexity of each block (emphasized in bold) and the intermediate steps.

\myparagraph{Block 1}
This block begins with constructing the search space, where each search space can contain up to $S = C'G + 1$ elements.
Following this, the joints are computed, where each joint computation has a complexity of $\OO(DH)$ (see \cref{appendix:EfficientEvaluationMFAs}).
Next, selecting the $C'$ largest values among the computed joints to obtain $\KKn$ for a data point can be performed in $\OO(S)$ \citep{BlumEtAl1973}.
Since we need to evaluate up to $S$ joints per data point, the overall complexity of block 1 is $\OO(NSDH)$.

\myparagraph{Block 2}
In block 2, the data set is partitioned into $\II_{1:C}$.
To find the most likely component~$c_n$ within the $\KKn$ set of each data point, the component with the largest joint is selected, which has a complexity of $\OO(C')$.
Repeating this for all data points results in a total complexity of $\OO(NC')$.

\myparagraph{Block 3}
By construction, the union of all $\II_c$ contains all indices of the $N$ data points exactly once.
Therefore, in block 3, we loop over each search space of the data points exactly once, resulting in a complexity of $\OO(NS)$.
On average, $\II_c$ contains $N/C$ indices.
Thus, the average cost to iterate over all $n$ in $\II_c$ and the respective $\GGn$ is $\OO^{\dag}((N/C)S)$.
Similar to the update of $\KKn$ in block 1, the complexity of finding $G$ replacement candidates to obtain $\GGc$ is $\OO^{\dag}((N/C)S)$ on average.

\myparagraph{Block 4}
Normalizing the variational distributions requires calculating the normalization constant $Z_n$ by summing over all $C'$ indices in $\KKn$, followed by dividing by $Z_n$.
Consequently, the total computational complexity for all data points is $\OO(NC')$.

Summarizing the complexities of all four blocks, the overall algorithmic complexity of \cref{alg:estep_complexity} is $\OO(NSDH + NC' + NS + NC') = \OO(NSDH)$.

\begin{algorithm}[tb]
  \For(\COMMENT{$\boldsymbol{\OO(NSDH)}$}){$n=1:N$}{
    $\GGn =\bigcup_{c\in\KKn}\GGc$;  \COMMENT{$\OO(S)$}\algBreak
    $\GGn =\GGn \cup \{c\}$ with $c \sim \mathcal{U}\{1,C\}$;\COMMENT{$\OO(1)$} \algBreak \nl
    \For (\COMMENT{$\OO(SDH)$}) {$c\in\GGn$}{
      compute joint $p(c,\bxn\,|\,\bTheta)$; \COMMENT{$\OO(DH)$ }
    }
    $\KKn \!=\! \{c \mid p(c,\bxn\,|\,\bTheta)$ is among the $C'$ largest joints for all $c \in \GGn\}$;\COMMENT{$\OO(S)$}
  } \algline \\

  \For(\COMMENT{$\boldsymbol{\OO(NC')}$}){$n=1:N$}{
    $\con = \myargmax{c\in\KKn} \ p(c,\bxn\,|\,\bTheta)$;\COMMENT{$\OO(C')$}\\[-10pt]
    \nl\\[0pt]
    $\II_{\con} = \II_{\con} \cup \{n\}$; \COMMENT{$\OO(1)$}
  } \algline \\
  \For(\COMMENT{$\boldsymbol{\OO(NS)}$}){$c=1:C$}{
  \For(\COMMENT{$\OO^{\dag}((N/C)S)$}){$n\in\II_c$}{
    \For(\COMMENT{$\OO(S)$}){$\ct\in\GGn \setminus \{c\}$}{
  $\relt = \relt + \log p(\bxn\,|\,c,\bTheta) - \log p(\bxn\,|\,\ct,\bTheta)$;\COMMENT{$\OO(1)$}\\
  $N_{c\ct} = N_{c\ct} + 1$;\COMMENT{$\OO(1)$}\\
    [-10pt]
    \nl\\[-3pt]
  ${\cal D}_c = {\cal D}_c \cup \{\ct\}$; \COMMENT{$\OO(1)$}\\
    }
  }
  \For(\COMMENT{$\OO^{\dag}((N/C)S)$}){$\ct\in {\cal D}_c$}{
    $\rel = \relt / N_{c\ct}$; \COMMENT{$\OO(1)$}\\

  }
  $\GGc= \{\ct\ |\ \rel \,$ is among the $G-1$ smallest values for $\ct \in {\cal D}_c\} \cup \, \{c\}$ \COMMENT{$\OO^{\dag}((N/C)S)$} \\
  } \algline\\
  \For(\COMMENT{$\boldsymbol{\OO(NC')}$}){$n=1:N$}{
    \For(\COMMENT{$\OO(C')$}){$c\in\KKn$}{
      $Z_n = Z_n + p(c,\bxn\,|\,\bTheta)$; \COMMENT{$\OO(1)$}\\[-4pt] \nl\\[-12pt]
    }
    \For(\COMMENT{$\OO(C')$}){$c\in\KKn$}{
      $\qnc = p(c,\bxn\,|\,\bTheta)/Z_n$; \COMMENT{$\OO(1)$}
    }
  }
  \algorithmfootnote{$\OO^{\dag}$ denotes amortized or average complexity}

  \caption{Complexity of the Variational E-step}
  \label{alg:estep_complexity}
\end{algorithm}

\Cref{alg:estep_complexity} is analogous to the partial variational E-step described in \citet{HirschbergerEtAl2022}, with three differences. We will discuss these differences below, focusing on the complexity of the respective algorithms.
First, we refrain from using coresets. In \citet{HirschbergerEtAl2022}, the use of coresets reduces the complexity from $N$ to the coreset size $N'$, where $N > N'$. But coresets are not desirable in our setting, as discussed in \cref{sec:Introduction} in the main text.
Second, computing a joint defined by the MFA model requires $\OO(DH)$ operations, in contrast to the $\OO(D)$ operations required for the GMMs considered in \citet{HirschbergerEtAl2022}. The additional cost allows for a much more flexible modeling of correlations (i.e., of component shapes) than GMMs restricted to uncorrelated data per component. Nevertheless, a quadratic scaling with $D$ (as would be required for full covariance matrices) is avoided.
Third, the update of variational parameters via the estimation of component similarity is different (see \cref{sec:construction_Gn} in the main text, \cref{appendix:comparison_approx_analytical} and \cref{appendix:comparison_approx_numerical}).

\subsection{Postprocessing Routine for Denoising}
\label{appendix:denoising_with_gmms}

As part of our numerical experiments, we apply \varMFA{} and other variants of GMMs for the task of image denoising, i.e., we aim to remove noise from an image corrupted by
artificial (Gaussian) noise or non-artificial, `inherent' noise (see \cref{sec:denoising} in the main text and \cref{appendix:further_denoising_results}).
We follow the image denoising pipeline for generative models used in \citet{DrefsEtAl2022} \citep[also compare][]{SalwigEtAl2024}. We start by outlining the general procedure
before providing details specific for the generative models used here.

For a given image, we first extract all possible overlapping patches of a specified size.
Consider an image of size $\mathsfit{W} \times \mathsfit{H} \, (\times \, \mathsfit{C})$ and a patch size of $\mathsfit{P}_\mathsfit{W} \times \mathsfit{P}_\mathsfit{H} \, (\times \, \mathsfit{C})$, where $\mathsfit{C}$ denotes the number of color channels ($\mathsfit{C} = 3$ for RGB images, $\mathsfit{C} = 1$ for grayscale).
Using a sliding window, we obtain $N = (\mathsfit{W} - \mathsfit{P}_\mathsfit{W} + 1) \cdot (\mathsfit{H} - \mathsfit{P}_\mathsfit{H} + 1)$ noisy patches.
The extracted $N$ noisy patches  are the data points. Each patch is taken to be a flattened vector $\bx$ of length $D=\mathsfit{W} \times \mathsfit{H} \, (\times \, \mathsfit{C})$.
Note that the flattening can be done row-wise, column-wise, or in any consistent order as the used generative models (GMMs and MFAs) do not assume any structure (such as two-dimensional patch structures).
Still we will refer to $\bx$ as an image patch, as the vector contains the information of one patch.

\begin{figure}[bt]
  \centering
  \includegraphics[width=0.7\textwidth]{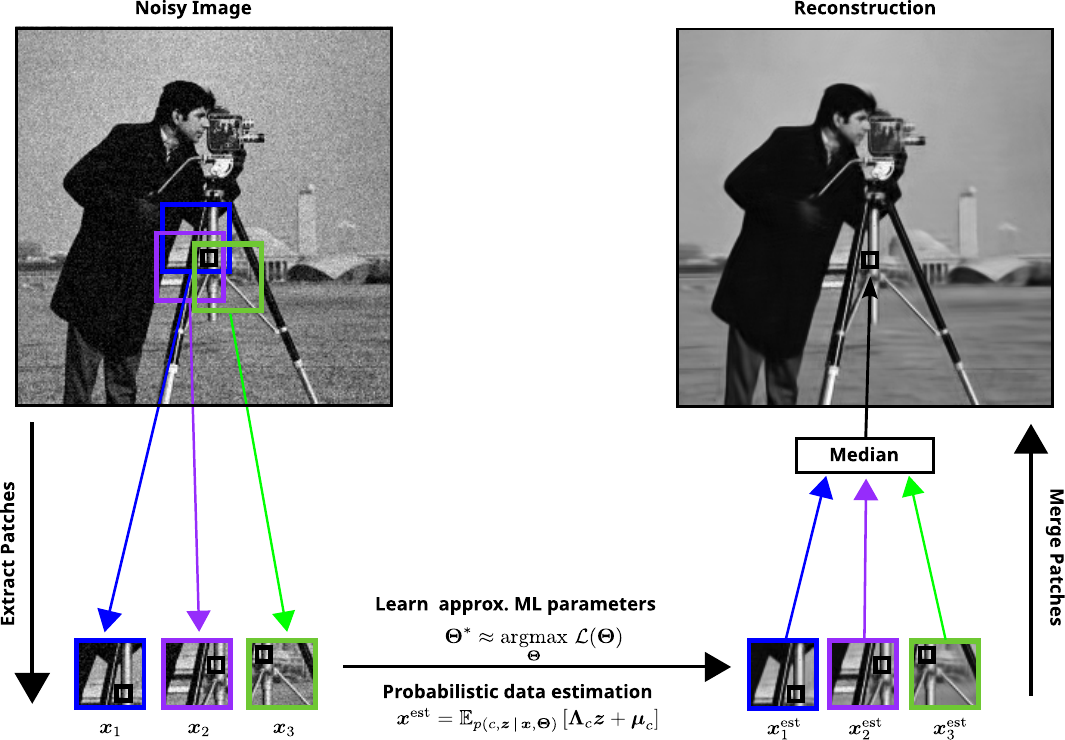}
  \caption{Illustration of patch-based blind zero-shot image denoising using probabilistic generative models.
    A noisy image serves as input (top left).
    Patches extracted from the noisy image (bottom left) are used to learn a probabilistic representation of the underlying image structure via an appropriate generative model (e.g., MFAs).
    Non-noisy patches are then estimated using this learned representation (bottom right).
    The final denoised image is obtained by aggregating pixel estimates from overlapping patches, using the median (top center and right).
    This approach requires no clean training data, and can be directly applied to a single noisy image.}
  \label{fig:denoising_pipeline}
\end{figure}

For each noisy patch $\bx$, our goal is to estimate a corresponding `clean' patch $\bxEst$.
Therefore, the data model is trained on the noisy patch set $\bx_{1:N}$ (see, e.g., \cref{sec:algorithmic_realization} and \cref{alg:varMFA}).
After inferring the model parameters, the learned representation is used to estimate the non-noisy patches.
Here we use a probabilistic data estimator \citep[compare][]{DrefsEtAl2022}, as described in detail below.
With clean versions of each extracted patch, the image is reconstructed by merging these patches.
As overlapping patches are used, each pixel in the image has multiple estimates.
The single final pixel value is determined by computing the median of all corresponding estimates for a given pixel. A schematic overview of this denoising pipeline for generative models is shown in \cref{fig:denoising_pipeline}.

In the following, we derive the probabilistic data estimator for MFAs and GMMs with isotropic or diagonal covariance matrices.

\myparagraph{MFAs}
For the MFA model, the denoised pixel values of a given image patch $\bx$ are obtained by
\begin{align}
  \label{zero-eq:deri2}
  \bxEst & = \EE{p(c, \bz \,|\, \bx,\bTheta)}{\bAc\bz+\bmuc} \nonumber     \\
         & = \sum_c \int p(c, \bz \,|\, \bx,\bTheta) (\bAc\bz+\bmuc) d\bz.
\end{align}

The joint posterior distribution $p(c, \bz \,|\, \bx,\bTheta)$ can be factorized as
\begin{align}
  p(c, \bz \,|\, \bx,\bTheta) = p(c \,|\, \bx,\bTheta) p(\bz \,|\, c, \bx, \bTheta).
\end{align}
Inserting this in \cref{zero-eq:deri2} yields
\begin{align}
  \label{zero-eq:deri3}
  \bxEst & =  \sum_c \int p(c \,|\, \bx, \bTheta) p(\bz \,|\, c, \bx,\bTheta) (\bAc\bz+\bmuc) d\bz \nonumber                                    \\
         & = \sum_c p(c \,|\, \bx,\bTheta) \bAc \int p(\bz \,|\, c, \bx, \bTheta)\, \bz \, d\bz + \sum_c p(c \,|\, \bx,\bTheta) \bmuc \nonumber \\
         & = \sum_c p(c \,|\, \bx,\bTheta) \bAc  \EE{p(\bz \,|\, c, \bx, \bTheta)}{\bz} + \sum_c p(c \,|\, \bx,\bTheta) \bmuc.
\end{align}
The posterior $p(\bz\,|\,c,\bx,\bTheta)$ conforms to a Gaussian distribution $\mathcal{N}(\bz; {\bmu}_z, {\bSigma}_z)$ with ${\bSigma}_z = \bLci$ and $\bmu_z = \bVc (\bx - \bmu_c)$ (compare \crefnobrackets{eq:z_gauss_params}).
Consequently, it holds that \mbox{$\EE{p(\bz \,|\, c, \bx,\bTheta)}{\bz} = \bmu_z$.}
Inserting this into \cref{zero-eq:deri3} yields
\begin{align}
  \label{zero-eq:deri4}
  \bxEst & =  \sum_c p(c \,|\, \bx,\bTheta) \bAc\,\bmu_z + \sum_c p(c \,|\, \bx,\bTheta) \bmuc \nonumber           \\
         & =  \sum_c p(c \,|\, \bx,\bTheta) \bAc\bVc (\bx - \bmu_c)+ \sum_c p(c \,|\, \bx,\bTheta) \bmuc \nonumber \\
         & =  \sum_c p(c \,|\, \bx,\bTheta) (\bAc\bVc (\bx - \bmu_c)+ \bmuc) \nonumber                             \\
         & = \EE{p(c \,|\, \bx,\bTheta)}{\bAc\bVc (\bx - \bmu_c)+ \bmuc}.
\end{align}

For an MFA model trained with \fullMFA{}, \cref{zero-eq:deri4} is used to reconstruct patches.
However, for an MFA model trained with \varMFA{}, the expectation can then be efficiently approximated by replacing the exact posteriors $p(c \,|\, \bxn,\bTheta)$ with the truncated posteriors $\qnc$ defined in \cref{eq:qnc}, as follows
\begin{align}
  \label{eq:de_mfa_approx}
  \bxEst & \approx \EE{\qnc}{\bAc\bVc(\bxn-\bmuc)+\bmuc}.
\end{align}
\myparagraph{Isotropic and diagonal GMMs}
For GMMs with isotropic or diagonal covariance matrices, the denoised pixel values of a given image patch $\bxn$ are obtained by
\begin{align}
  \bxEst & = \EE{p(c \,|\, \bxn,\bTheta)}{\bmu_c}.
\end{align}
For a GMM model trained with our variational approach, we efficiently approximate the posterior $p(c \,|\, \bxn,\bTheta)$ with the variational posterior $\qnc$:
\begin{align}
  \label{eq:de_gmm_approx}
  \bxEst & \approx \EE{\qnc}{\bmuc}.
\end{align}

\setcounter{figure}{0}
\setcounter{equation}{0}
\setcounter{table}{0}

\renewcommand*{\thefigure}{B\arabic{figure}}
\renewcommand*{\thetable}{B\arabic{table}}
\renewcommand*{\theequation}{B\arabic{equation}}

\setcounter{section}{1}
\section{Supplementary Information on Numerical Experiments and Control Experiments}

In the following, we provide further specifics on the implementation and execution of the algorithms, the used hardware and the data sets in \cref{appendix:further_details}, results of additional control experiments in \cref{appendix:control_experiments} and supplementary information on the quality analysis in \cref{appendix:further_results}.

\subsection{Further Details on Numerical Experiments}
\label{appendix:further_details}
Below, we present additional details regarding the implementation and execution of the algorithms as well as information about the used hardware and the data sets.

\subsubsection{Algorithms}
\label{appendix:algorithms}

The \fullMFA{} and the \varMFA{} algorithms are our own implementations.
To ensure a fair comparison, the primary motivation behind developing these algorithms was to optimize for execution speed and efficient memory usage.
The \torchmfa{} algorithm is a published PyTorch-based implementation of MFA that uses Stochastic Gradient Descent (SGD).
The \kmeansfa{} algorithm combines \kmeans{} with factor analysis.
While it was used as an initialization method in \citet{Richardson2018}, we employ it here as a comparison method to the other algorithms.

\myparagraph{\varMFA} The \varMFA{} algorithm trains the MFA model using the variational EM algorithm described in \cref{sec:varEM,sec:VariationalEStep,sec:construction_Gn,sec:algorithmic_realization} in the main text.
The implementation uses both Python and C++. The core functionality is written in C++, utilizing the open-source libraries Eigen \citep{eigen} and Boost \citep{boost}.
To facilitate data communication between Python and C++, pybind11 \citep{pybind11} is used.
The \varMFA{} source code is available on GitHub.\furl{https://github.com/variational-sublinear-clustering/vamm}
The implementations of \varMFA{}$^{\mathrm{Eucl.}}$, v-ISO and v-ISO$^{\mathrm{Eucl.}}$ used in \cref{appendix:control_experiments} as well as v-DIAG used in \cref{appendix:comparison,appendix:further_denoising_results} are also provided in this source code.
The source code of the denoising extension for the \varMFA{} algorithm is also available on GitHub.\furl{https://github.com/variational-sublinear-clustering/vammdx}

\myparagraph{\fullMFA} Our Python implementation of the \fullMFA{} algorithm optimizes the MFA model using exact inference via conventional EM (full posteriors). It primarily leverages the open-source library PyTorch \citep{pytorch}.
The \fullMFA{} source code is available on GitHub.\furl{https://github.com/variational-sublinear-clustering/emmi}
The implementation em-DIAG used in \cref{appendix:comparison,appendix:further_denoising_results} is also provided in this source code.
The source code of the denoising extension for the \fullMFA{} algorithm is also available on GitHub.\furl{https://github.com/variational-sublinear-clustering/emmidx}

\myparagraph{\kmeansfa{}} In this integrated methodology, the initial step entails the application of the \kmeans{} algorithm to the data set.
Upon convergence, the data is partitioned into Voronoi cells, with each data point $\bxn$ assigned to its closest cluster (component) $c$, i.e.,
\begin{align}
  \label{eq:voronoi_cells}
  \mathcal{P}_c = \{n \mid c=c_n\} \quad \text{with} \quad c_n = \myargmin{c'}\ ||\bxn - \bmu_{c'} ||^2.
\end{align}
A factor analyzer is then applied to each data subset $\mathcal{P}_c$, where each factor analyzer independently learns the parameters $\bmu_c$, $\bA_c$ and $\bD_c$ for the corresponding cluster $c$.
The priors~$\pi_c$ are determined based on the relative subset sizes, i.e.,
\begin{align}
  \pi_c = \frac{|\mathcal{P}_c|}{N}, \nonumber
\end{align}
where $N$ is the number of the total data points.
The \kmeansfa{} algorithm is realized by using the Scikit-learn \citep{sklearn} implementations (version $1.5.0$) of \kmeans{} and Factor Analyzer, using their default hyperparameters and settings.

\myparagraph{\torchmfa{}}
We used the official \torchmfa{} \citep{torchmfaCode} implementation,
an advancement of the stochastic gradient descent (SGD) based MFA used in \citet{Richardson2018}. Due to the SGD optimization, \torchmfa{} introduces additional parameters, such as learning rate and batch size. We set the batch size to 256, consistent with \citet{Richardson2018}, and determined the optimal learning rate to be $10^{-2}$ via a coarse grid search (prior to the experiments). We modified the function \texttt{sgd\_mfa\_train} in the \texttt{mfa.py} script to implement a convergence criterion similar to \cref{eq:convergence} in the main text. The original function does a fixed number of iterations. Unlike \fullMFA{} and \varMFA{}, where the free energy is computed for the entire data set during each E-step, \torchmfa{} updates parameters after each batch, making full data set evaluation impractical. Instead, we computed the free energy on a randomly selected subset of the data set after each epoch and used this for the convergence criterion in \cref{eq:convergence} in the main text. The subset size was set to $10 \times$ the batch size, and the computational time for this evaluation was excluded from the time measurement.

\myparagraph{AFK-MC$^2$} We provide an efficient implementation of the AFK-MC$^2$ seeding method \citep{BachemEtAl2016b} integrated into \varMFA{}. For all experiments, the Markov chain length was set to 10.

In the following, we describe the zero-shot denoising algorithms that we used for comparison with the \varMFA{} algorithm in \cref{sec:denoising}.
Note that for certain images, some algorithms failed to produce a meaningful denoised result, e.g., yielding almost uniformly gray outputs.
To address this issue, we use the PSNR and SSIM values of the noisy image whenever the corresponding denoised image performs worse in both metrics.

\myparagraph{Noise2Self} We used the official Noise2Self implementation \citep{N2SCode} and executed the algorithm with the same configuration as \citet{LequyerEtAl2022}, i.e., we increased the number of iterations from \num{500} to \num{20000}. To apply Noise2Self to RGB images, each color channel is processed independently.

\myparagraph{Noise2Void} We used the official Noise2Void implementation \citep{N2VCode}, following the \\ `\texttt{denoising2D\_RGB}' example, available in the respective GitHub repository. To be consistent with \citet{LequyerEtAl2022}, we used a patch size of $64 \times 64$, 100 epochs and 100 steps per epoch, a train batch size of 16 and a neighborhood radius of 5.

\myparagraph{Self2Self} We used the official Self2Self implementation \citep{S2SCode} and executed the algorithm with the configuration of the `\texttt{demo\_denoising.py}' script available in the respective GitHub repository.
On the Confocal data set, the default learning rate of $10^{-4}$ failed to yield meaningful results. Hence, on this data set, we lowered the learning rate to $10^{-5}$.

\myparagraph{Deep Image Prior}  We used the official Deep Image Prior implementation \citep{DIPCode} and executed the algorithm with the same configuration as \citet{LequyerEtAl2022}, including the same network (`skip-net') architecture  and a fixed number of 3000 iterations. For the parameter `\texttt{reg\_noise\_std}', we used a fixed value of $1 / 30$, independent of the noise level.

\myparagraph{Noise2Fast} We used the official Noise2Fast implementation \citep{N2FCode} and executed the algorithm following the `\texttt{N2F.py}' script for grayscale images and the `\texttt{N2F\_4D.py}' script for RGB images, both available in the respective GitHub repository.

\myparagraph{Pixel2Pixel} We used the official Pixel2Pixel implementation \citep{P2PCode} and executed the algorithm with the configuration of the `\texttt{Pixel2Pixel\_real.py}' script available in the respective GitHub repository.
On the Confocal data set, the default learning rate of $10^{-3}$ failed to yield meaningful results (the output image was uniformly gray).
Hence, on this data set, we lowered the learning rate to $10^{-4}$, which resulted in meaningful results for all but one images.

\myparagraph{DivNoising} 	We used the official DivNoising implementation \citep{DivNCode} and executed the algorithm with the configuration of the `\texttt{Convallaria}' and `\texttt{Mouse\_nuclei}' example available in the respective GitHub repository.
For certain images and noise levels in BSD68 and Set12, DivNoising produced almost uniformly gray outputs.
In such cases, we use the PSNR and SSIM values of the noisy image, as explained above.

\myparagraph{BM3D} We used the official BM3D Python software package \citep{BM3DCode} and executed the algorithm using the functions `\texttt{bm3d}' for grayscale images and `\texttt{bm3d\_rgb}' for RGB images.

\subsubsection{Hardware}
\label{appendix:hardware}

\Cref{tab:hardware} provides a detailed description of the hardware used for the various algorithms and experimental setups. To achieve high utilization, we executed four instances of \varMFA{}, \fullMFA{}, or \kmeansfa{} in parallel, with each instance using 16 of the 64 available cores, i.e., four independent runs were executed at the same time. The tool \texttt{GNU parallel} \citep{gnu_parallel} was employed for job scheduling.
Exceptions to this are the training of \varMFA{} and \fullMFA{} on YFCC100M and the denoising benchmarks in \cref{sec:largescale,sec:denoising} in the main text, where we executed only one instance at a time, using all 64 available cores.
We executed \torchmfa{} on an NVIDIA H100 GPU equipped with 96 GB of VRAM.
Running \torchmfa{} on the CPU configurations used by the other algorithms resulted in significantly longer execution times (GPU $\approx 1$ h  vs. CPU (16 cores) $>10$ h for a single epoch).
Similarly, for the denoising benchmarks in \cref{sec:denoising} in the main text, we executed Noise2Self, Noise2Void, Deep Image Prior, Self2SSelf, Noise2Fast, Pixel2Pixel and DivNoising on an NVIDIA L40s GPU equipped with 48 GB of VRAM.
BM3D was executed on the same CPU as \varMFA{}.

\begin{table}[tb]
  \centering
  \caption{Details on the utilized hardware for the different algorithms and experiments. The asterisk ($^\ast$) marks execution of four instances in parallel, with each instance using 16 of the 64 available cores, i.e., four independent runs were executed at the same time.}
  \label{tab:hardware}
  \begin{tabular}{cccc}
    \toprule
    Algorithm                                     & \multicolumn{3}{c}{Hardware}                                             \\
    \cmidrule(l{2pt}r{2pt}){2-4}
                                                  & Device                       & Model name           & Cores used         \\
    \midrule
    \varMFA{}, \fullMFA{}, \kmeansfa{}            & CPU                          & AMD Genoa EPYC 9554  & 16$^\ast$          \\
    \varMFA{}, \fullMFA{} (YFCC100M \& denoising) & CPU                          & AMD Genoa EPYC 9554  & 64$\phantom{\ast}$ \\
    \torchmfa{}                                   & GPU                          & NVIDIA H100 94GB SXM & all                \\
    N2S, N2V, DIP, S2S, N2F, P2P, DivN            & GPU                          & NVIDIA L40S 48GB     & all                \\
    BM3D                                          & CPU                          & AMD Genoa EPYC 9554  & 64$\phantom{\ast}$ \\
    \bottomrule
  \end{tabular}
\end{table}

\subsubsection{Data Sets and Preprocessing}
\label{appendix:datasets}

In the following, we provide additional information on the data sets used in the numerical experiments.
The main properties of the data sets are listed in \cref{tab:datasets}.
Unless otherwise stated, we use the official \emph{train} and \emph{test} splits, and the images are used in their original form without preprocessing.
Note that the label information was not used in any of the experiments.
The data sets listed in \cref{tab:datasets_denoising} are used for denoising benchmarks, where each image is processed individually by the denoising algorithms.

\myparagraph{CIFAR-10}
The CIFAR-10 data set \citep{Krizhevsky2009} contains of \num{60000} natural color images with size $32 \times 32$, divided into \num{50000} training images and \num{10000} test images.

\myparagraph{CelebA}
The original CelebA data set \citep{Liu2015} contains \num{202599} color images of faces sized at $178 \times 218$. We preprocessed the data to align, crop and resize the images to $64 \times 64$, following \citet{Richardson2018}.
Additionally, we merged the \emph{train} and \emph{valid} splits to obtain more training data.

\myparagraph{EMNIST}
The EMNIST data set \citep{Cohen2017} contains of small cropped grayscale images of handwritten digits and letters, sized at $28 \times 28$. The images are split into different groups. We used the \emph{byclass} split, which contains \num{697932} training images and \num{116323} test images.
To avoid zero variances at the image edges, standard Gaussian noise (with zero mean and standard deviation of 1) was added to the images.

\myparagraph{SVHN}
Similar to the digits in EMNIST, the SVHN data set \citep{Netzer2011} contains images of small cropped digits in the range from 0 to 9. But in contrast to EMNIST, these images are real, colored photographs capturing house numbers at a size of $32 \times 32$.
The SVHN data set is split into \num{73257} training images (\emph{train}), \num{26032} test images (\emph{test}) and \num{531131} additional training images (\emph{extra}). To obtain more training data, we merged the \emph{extra} and \emph{train} images.

\myparagraph{YFCC100M}
The original YFCC100M data set \citep{ThomeeEtAl2016} contains real-world images and videos.
We used all available images; some of the original \num{99 171 688} images were missing or corrupted and thus not included. The remaining \num{99 155 298} images of various sizes were center-cropped and resized with bilinear interpolation to $32 \times 32$.
We randomly selected a test split of \num{4 957 765} samples ($5\%$ of the data set) and used the remaining \num{94 197 533} samples for training.

The following data sets are used in the denoising benchmarks.

\myparagraph{Set12}
The Set12 data set contains of 12 grayscale natural images, with 7 of these 12 images of size $256 \times 256$ and the remaining 5 images of size $512 \times 512$.

\myparagraph{BSD68}
BSD68 \citep[][]{MartinEtAl2001} comprises 68 grayscale natural images with a resolution of $481 \times 321$ or  $321 \times 481$ pixels.

\myparagraph{Confocal}
This microscopy data set is the same subset of Confocal microscopy images from the Fluorescence Microscopy Data set \citep[FMD;][]{ZhangEtAl2019} as used by \citet{LequyerEtAl2022}. It contains 5 inherently noisy grayscale images of size $512 \times 512$.
To obtain (pseudo) ground-truth references, multiple captures of the same microscopy scene were averaged \citep[for more details, we refer to][]{ZhangEtAl2019}.

\myparagraph{Div2K}
As an example of a high resolution image, we include one image of the Div2K data set \citep{AgustssonRadu2017} in the denoising benchmark, namely the image with ID `\texttt{0010}'. This RGB image was chosen prior to any experiments and has a size of $2040 \times 1644$. We denote this image as `Tiger'.

\begin{table}[tb]
  \centering
  \caption{High-dimensional data sets used in our experiments in \cref{sec:scalability,sec:quality,sec:largescale}. The data dimension is given in pixel width $\times$ pixel height $\times$ color channels.}
  \label{tab:datasets}
  \begin{tabular}{@{}lS[table-format=8.0]S[table-format=7.0]rS[table-format=5.0]@{}}
    \toprule
    data set                   & {train samples} & {test samples} &
    \multicolumn{2}{c}{data dimension $D$}                                                             \\ \midrule
    CIFAR-10                   & 50000           & 10000          & $32
    \times 32 \times 3 = $     & 3072                                                                  \\
    CelebA                     & 182637          & 19962          & $
    64 \times 64 \times  3 = $ & 12288                                                                 \\
    EMNIST                     & 697932          & 116323         & $
    28 \times 28 \times 1 = $  & 784                                                                   \\
    SVHN                       & 604388          & 26032          & $
    32 \times 32 \times 3 = $  & 3072                                                                  \\
    YFCC100M                   & 94 197 533      & 4957765        & $ 32 \times 32 \times 3 = $ & 3072 \\
    \bottomrule
  \end{tabular}
\end{table}

\begin{table}[tb]
  \centering
  \caption{Data sets used in the denoising experiments in \cref{sec:denoising}. The data dimension is given in pixel width $\times$ pixel height $\times$ color channels.
  }
  \label{tab:datasets_denoising}
  \begin{tabular}{@{}lS[table-format=2.0]rS[table-format=8.0]@{}}
    \toprule
    data set                                 & {samples} &
    \multicolumn{2}{c}{data dimension $D$}                      \\ \midrule
    Set12 images \texttt{01} $-$ \texttt{07} & 7         & $256
    \times 256 \times 1 = $                  & 65536            \\
    Set12 images \texttt{08} $-$ \texttt{12} & 5         & $512
    \times 512 \times 1 = $                  & 262144           \\
    BSD68                                    & 68        & $
    321 \times 481 \times  1 = $             & 154401           \\
    Confocal                                 & 5         & $
    512 \times 512 \times 1 = $              & 262144           \\
    Div2K `Tiger'                            & 1         & $
    2040 \times 1644 \times 3 = $            & 10061280         \\
    \bottomrule
  \end{tabular}
\end{table}

\subsection{Numerical Control Experiments}
\label{appendix:control_experiments}

The following subsections report the results of additional control experiments, offering supporting or complementary insights to those discussed in \cref{sec:numerical_experiments} in the main text.

\subsubsection{Comparison of Component-to-Component Distance Estimates}
\label{appendix:comparison_approx_numerical}

In this section, we compare our v-MFA algorithm, which uses the proposed definition of $g_c$ based on the estimate $\rel$ in \cref{eq:deltaF} in the main text, with a variant of v-MFA (referred to as \varMFA{}$^{\mathrm{Eucl.}}$) that uses the definition of $g_c$ introduced in \citet{HirschbergerEtAl2022}, based on $d^2_{c\ct}$ in \cref{eq:euclidean_distance}.
In addition to the MFA model, we also consider isotropic GMMs (compare Equation~(5) in \citealp{HirschbergerEtAl2022}) and their optimization using variational EM with $g_c$ based on $\rel$ (referred to as v-ISO) and with $g_c$ based on $d^2_{c\ct}$ (referred to as v-ISO$^{\mathrm{Eucl.}}$). v-ISO$^{\mathrm{Eucl.}}$ is the algorithm used in previous work \citep{HirschbergerEtAl2022}.

We numerically compare v-MFA and \varMFA{}$^{\mathrm{Eucl.}}$ to investigate the differences of the here derived approach to the previous approach.
Additionally, we compare v-ISO to v-ISO$^{\mathrm{Eucl.}}$ to investigate potential advantages and disadvantages under the constraint of isotropic components.

The models are initialized identically and using the same settings as described in \cref{sec:quality} and \cref{fig:quality} in the main text.
Since both methods use partial E-Steps with the same computational complexity, we run the algorithms for a fixed number of iterations.
Concretely, we train \varMFA{} (or v-ISO) until convergence, and then train \varMFA{}$^{\mathrm{Eucl.}}$ (or v-ISO$^{\mathrm{Eucl.}}$) for the same number of iterations.

\begin{figure}[tb]
  \centering
  \includegraphics[width=0.8\textwidth]{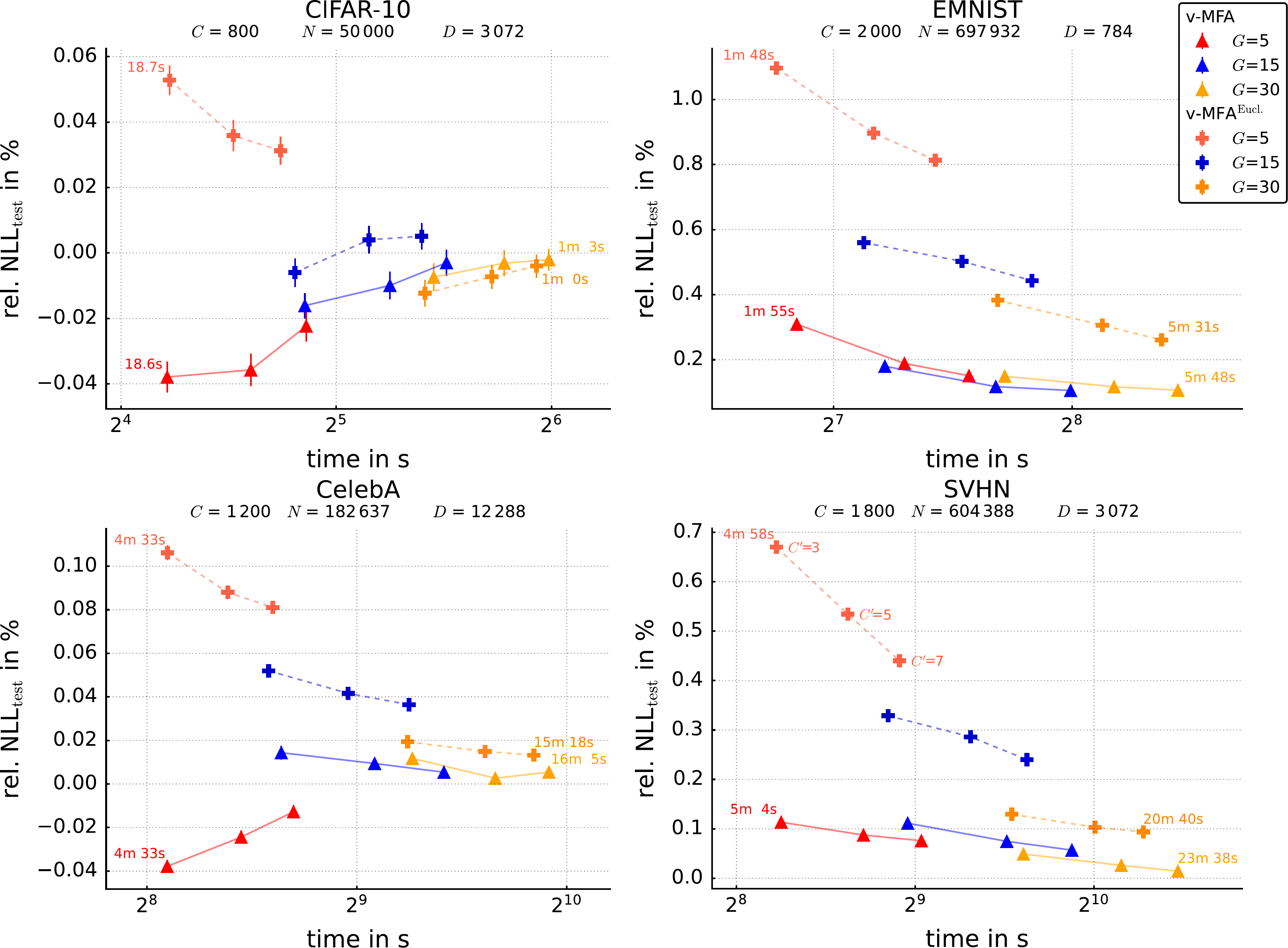}
  \caption{
    Effectiveness and efficiency of \varMFA{} and \varMFA{}$^{\mathrm{Eucl.}}$ in terms of NLL$_\mathrm{test}$ (relative to \fullMFA{}) and the runtime. Triangles refer to \varMFA{}, while pluses denote \varMFA{}$^{\mathrm{Eucl.}}$.
    Each of the four subfigures corresponds to experiments on one benchmark data set.
    The settings are similar to those in \cref{fig:quality} in the main text, but both algorithms are trained for the same number of iterations, as detailed in the text.
    For each configuration of $G \in \{5,15,30\}$, measurements refer to configurations with $C' \in \{3,5,7\}$ (three red, blue and orange triangles or pluses, respectively).
    Configurations with larger $C'$ lie to the right, due to their longer runtimes.
    Each measurement point is averaged over 40 independent runs and error bars denote the standard error of the mean (SEM).
  }
  \label{fig:compare_methods_MFA}
\end{figure}

\begin{figure}[tb]
  \centering
  \includegraphics[width=0.8\textwidth]{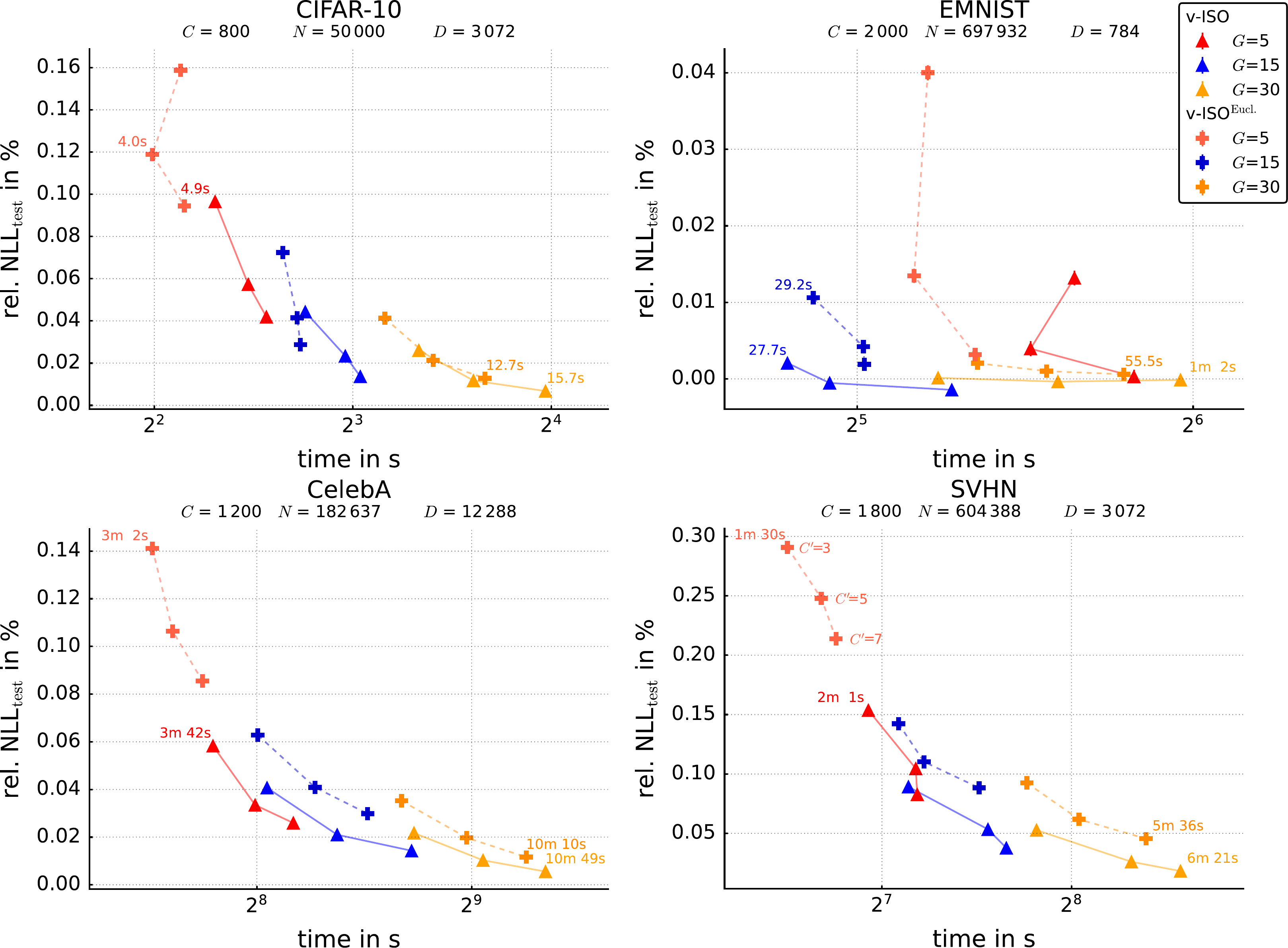}
  \caption{
    Effectiveness and efficiency of v-ISO and v-ISO$^{\mathrm{Eucl.}}$ in terms of NLL$_\mathrm{test}$ (relative to \fullMFA{}) and the runtime. Triangles refer to v-ISO, while pluses denote v-ISO$^{\mathrm{Eucl.}}$.
    Each of the four subfigures corresponds to experiments on one benchmark data set.
    The settings are similar to those in \cref{fig:quality} in the main text, but both algorithms are trained for the same number of iterations, as detailed in the text.
    For each configuration of $G \in \{5,15,30\}$, measurements refer to configurations with $C' \in \{3,5,7\}$ (three red, blue and orange triangles or pluses, respectively).
    Each measurement point is averaged over 40 independent runs and error bars denote the standard error of the mean (SEM).
  }
  \label{fig:compare_methods_iso}
\end{figure}

The comparison results for the MFA models and for the isotropic GMMs are presented in \cref{fig:compare_methods_MFA} and \cref{fig:compare_methods_iso}, respectively.
In the case of MFAs, the here derived \varMFA{} algorithm consistently performs better (or comparably within the error bars) in terms of  NLL$_\mathrm{test}$ values when compared to \varMFA{}$^{\mathrm{Eucl.}}$ (especially for smaller search spaces $\GGn$).
As the search space expands, the NLL$_\mathrm{test}$ values of both algorithms approach each other.

Also in the case of GMMs with isotropic components, v-ISO improves on NLL$_\mathrm{test}$ values compared to v-ISO$^{\mathrm{Eucl.}}$, albeit at a slightly higher computational cost, due to smaller intersections in the search spaces.
Notably, v-ISO$^{\mathrm{Eucl.}}$ only differs slightly from v-ISO in terms of NLL$_\mathrm{test}$ values for most of the data sets and settings of $C'$ and $G$ (see \cref{fig:compare_methods_iso}), which confirms the suitability $d^2_{c\ct}$ of \cref{eq:euclidean_distance} for the isotropic case, for which it was originally derived \citep{HirschbergerEtAl2022}.
In contrast, in the case of GMMs with flexible covariances (i.e.,\ for the MFA model), the significant improvements of \varMFA{} compared to \varMFA{}$^{\mathrm{Eucl.}}$ highlight the importance of the more general definition of $g_c$ using $\rel$ given by \cref{eq:deltaF} in the main text.

\subsubsection{Comparison to other GMM/MFA Approaches}
\label{appendix:comparison}

In experiments in \cref{sec:scalability} and \cref{sec:quality} in the main text, we showed the speed-up of our
variationally accelerated algorithm (\varMFA{}) compared to conventional EM optimization for MFA (\fullMFA{}).
Other variants of GMMs and MFAs and corresponding optimization methods do exist but often have focus on areas of application other than scalability.
For instance, the AECM \citep[][]{McLachlanEtAl2003} approach and the ECM approach \citep[][]{ZhaoYu2008} have been applied to MFA optimization.
Both approaches do not focus on scalability and, as discussed in \cref{appendix:EfficientEvaluationMFAs}, they exhibit a higher computational complexity compared to conventional EM (and we compare to the latter).
Methods such as \citet{HirschbergerEtAl2022} and \citet{ExarchakisEtAl2022} address scalability but are limited to isotropic or diagonal covariances.
A method by \citet{AsheriEtAl2021}
allows for more general covariance constraints, but their focus is on enabling larger numbers of components for the same amount of data. Their work does not focus on improving scalability compared to conventional GMM optimization.
The method in the literature most related to our approaches and our goal of scalability is thus \torchmfa{} which makes use of stochastic gradient decent integrated as part of the PyTorch framework.

The \torchmfa{} algorithm represents a further advancement of the TensorFlow implementation employed to produce results previously reported by the same group \citep{Richardson2018}  (for further details, see \cref{appendix:algorithms}).
The work by \citet{Richardson2018} documents (to the knowledge of the authors) the largest MFA models prior to this study.
However, at the scales here addressed, one can observe clear efficiency limits of \torchmfa{} (as well as its predecessor implementation), which we elaborate on further below.
When considering time measurements for the various settings, note that although actual runtimes have significant practical relevance, they are inherently dependent on implementation details and hardware.
We do not claim, nor do we expect, that \torchmfa{} is optimal in any specific sense, but rather representative of what a typical implementation might achieve.

In addition to \torchmfa{}, we also compare against GMMs constrained to diagonal covariance matrices. Like \torchmfa{}, \varMFA{} models intra-component correlations. A comparison with GMMs with diagonal
covariance matrices can thus be instructive about the benefits of modeling correlations. For our comparison, we concretely consider diagonal GMMs trained via conventional EM (denoted em-DIAG) and via our variational approach (denoted v-DIAG).

To assess the performance of \torchmfa{}, em-DIAG and v-DIAG against \varMFA{} and \fullMFA{}, we adopted the configurations of MFA hyperparameters used in an experiment on the CelebA data set \citep[see][]{Richardson2018}.
That is, the number of components and the hyperplane dimension were set to $C=1000$ and $H=10$ for all algorithms using MFAs.
We ran \torchmfa{} on a GPU with the configurations specified in \cref{appendix:further_details}. This is in contrast to the CPU configurations for the other algorithms, as discussed in \cref{appendix:hardware}.
The experiments in \citet{Richardson2018} were performed utilizing hierarchical training (for more details we refer to \citealp{Richardson2018}).
To ensure a fair comparison, we trained \torchmfa{} in our experiments directly on the complete data set, without hierarchical training.
In addition to these algorithms, we used the \kmeansfa{} algorithm in our experimental comparison.

The results of our comparison are shown in \cref{tab:comparison}.
As diagonal GMMs contain substantially fewer parameters than MFAs with the same number of components $C$, \cref{tab:comparison} also contains v-DIAG versions with larger $C$.
For $C = \num{6 000}$, v-DIAG contains approximately the same number of parameters as \varMFA{} with $C = \num{1 000}$. For $C = \num{10 000}$ and $C = \num{20 000}$, v-DIAG far exceeds the number of parameters of v-MFA.
The number of components for v-DIAG with $C>1000$ is indicated by superscripts on v-DIAG in \cref{tab:comparison}.

\begin{table}[tb]
  \centering
  \caption{Comparison of \fullMFA{}, \varMFA{}, \kmeansfa{},\torchmfa{}, em-DIAG with $C=1000$ and multiple v-DIAG variants with $C \in \{1000,6000,\num{10000},\num{20000}\}$ on CelebA. To ensure consistency with the other methods, we define an iteration for \torchmfa{} as completed when the model has processed the entire data set, rather than just one data batch. Results are averaged over 4 independent runs. All errors indicate the standard error of the mean (SEM), rounded up to the displayed precision. The best values for NLL$_\mathrm{test}$, the relative NLL$_\mathrm{test}$, total runtime including initialization (d=days, h=hours, m=minutes, s=seconds) and the number of joint or distance evaluations are marked bold *(initialization starts dominating the runtime for $C=\num{10000}$ and $\num{20000}$, e.g., the initialization alone already requires over 10m for $C=\num{20000}$). \label{tab:comparison}
  }
  \resizebox{0.9\linewidth}{!}{
    \begin{tabular}{@{}l r@{$\,\pm\,$}l r@{$\,\pm\,$}l r@{$\,\pm\,$}l r@{$\,\pm\,$}l c @{}}
      \toprule
                              & \multicolumn{2}{c}{NLL$_\mathrm{test}$} & \multicolumn{2}{c}{rel. NLL$_\mathrm{test}$} & \multicolumn{2}{c}{runtime} & \multicolumn{2}{c}{joint / distance eval.} & \multicolumn{1}{c}{iterations}                                                                                 \\
      \midrule
      \fullMFA{}              & 57571                                   & 3                                            & \multicolumn{2}{c}{0.00 \%} & 1h 16m                                     & 53.5s                          & $(4.66 $  & $ 0.05)\cdot 10^{9}$  & 25.5                                      \\
      \varMFA{}               & \bf{57565}                              & \bf{2}                                       & \bf{(-0.01}                 & \bf{0.01) \%}                              & 9m  2s                         & 11.6s     & $(2.47 $              & $ 0.05)\cdot 10^{8}$              & 35.0  \\
      \kmeansfa{}             & 58373                                   & 2                                            & (1.39                       & 0.00) \%                                   & 12m 12s                        & 2.8s      & $(1.83 $              & $ 0.01)\cdot 10^{10}$             & 106.7 \\
      \torchmfa{}             & 57868                                   & 13                                           & (0.52                       & 0.02) \%                                   & 1d 14h                         & 1h 35m    & $(6.8 $               & $ 0.3)\cdot 10^{9}$               & 38.5  \\[0.75em]
      em-DIAG                 & 61232                                   & 5                                            & (6.36                       & 0.01) \%                                   & 5m 13s                         & 0.5s      & $(2.56 $              & $ 0.01)\cdot 10^{9}$              & 14.0  \\
      v-DIAG                  & 61257                                   & 5                                            & (6.40                       & 0.01) \%                                   & \bf{ 1m  5s}                   & \bf{0.7s} & $\boldsymbol{(1.69 }$ & $\boldsymbol{ 0.02)\cdot 10^{8}}$ & 24.0  \\
      v-DIAG$^{6\mathrm{k}}$  & 60876                                   & 3                                            & (5.74                       & 0.01) \%                                   & 2m 38s                         & 0.6s      & $(3.87 $              & $ 0.02)\cdot 10^{8}$              & 27.5  \\
      v-DIAG$^{10\mathrm{k}}$ & 60864                                   & 4                                            & (5.72                       & 0.00) \%                                   & 4m 24s                         & 1.5s*     & $(7.12 $              & $ 0.02)\cdot 10^{8}$              & 28.0  \\
      v-DIAG$^{20\mathrm{k}}$ & 60930                                   & 1                                            & (5.83                       & 0.01) \%                                   & 12m 26s                        & 1.0s*     & $(2.21 $              & $ 0.01)\cdot 10^{9}$              & 27.0  \\
      \bottomrule
    \end{tabular}
  }
\end{table}

The findings show that \varMFA{} achieved the lowest NLL$_\mathrm{test}$ value in comparison to the other algorithms, with NLL values that are essentially equal to those of \fullMFA{}.
Among the MFA-based approaches, \varMFA{} was clearly the fastest algorithm, completing in approximately 9 minutes.
Following closely behind is \kmeansfa{}, with a runtime of approximately 12 minutes.
The \fullMFA{} algorithm exhibited considerably worse runtime performance, necessitating over 1 hour to complete.
Remarkably, the \torchmfa{} algorithm was the most time-consuming, requiring approximately one and a half days to complete, making it the slowest among the algorithms.

A possible reason for the long runtimes of \torchmfa{} could be related to the optimization procedure.
Unlike the other algorithms, which update model parameters once per iteration (M-step evaluated over the entire data set), \torchmfa{} updates parameters after each data batch.
Consequently, this results in approximately $\frac{N}{\text{batch size}} = \frac{182637}{256} \approx 713 \times$ as many (yet less computationally expensive) parameter updates.
In this context, we attempted to increase the batch size (to 512) in order to potentially reduce the runtime.
However, this was not possible due to memory constraints of the used GPU.
Another factor to consider is that the design objectives of the \fullMFA{} and \varMFA{} algorithms were mainly focused on optimizing speed, whereas speed might not have been the main objective for the \torchmfa{} algorithm.

In terms of the number of joint or distance evaluations, \varMFA{} also exhibited an order of magnitude fewer evaluations than \fullMFA{} and \torchmfa{}, which were within the same order of magnitude.
The greatest number of distance evaluations was performed by \kmeansfa{}, once more an order of magnitude higher than those of \fullMFA{} and \torchmfa{}.
This observation seems to be inconsistent with the runtimes of the algorithms.
However, a distance evaluation of \kmeans{} is significantly less computationally expensive than a joint evaluation of the other algorithms.

When comparing the MFA-based approaches to GMMs with diagonal covariance matrices, it can be observed that none of the diagonal GMM variants could reach NLL$_\mathrm{test}$ values as good as the MFA-based models.
Increasing the number of components improves performance to some extent, but even for $C=\num{10000}$, the NLL$_\mathrm{test}$ value remains significantly higher than those of the MFA-based approaches (although v-DIAG has more model parameters available in this case).
Note that further increasing the number of components does not improve performance, because v-DIAG starts to overfit the training data.
NLL$_\mathrm{test}$ values for v-DIAG with $C$ increased to $\num{20000}$ therefore get worse.
For this reason, using $C=\num{10000}$ achieves the best NLL$_\mathrm{test}$ value that we have observed for v-DIAG.

\subsubsection{Control Experiments for the \varMFA{} Scalability Analysis}
\label{appendix:scaling-same-quality}

The results in \cref{fig:scaling} in the main text do not account for the quality of \fullMFA{} and \varMFA{}.
However, it is important to consider the optimization quality also for the investigation of scaling behavior because
the improved scaling of \varMFA{} may be attributed to two key factors:
First, the variational approach reduces the computational complexity of the algorithm, which is the desired effect we aim to measure.
Second, some loss in quality can typically be expected for \varMFA{} as it represents an approximation of the conventional EM optimization of MFA.

Therefore, we conducted controls by repeating the experiments in \cref{sec:scalability} in the main text, now incorporating the different optimization qualities as measured by the NLL on the test data sets.
Concretely, we stopped the \fullMFA{} algorithm once it achieved the same NLL$_\mathrm{test}$ as was achieved after a full run of the \varMFA{} algorithm.

\begin{figure}[htb]
  \includegraphics[width=0.98\textwidth]{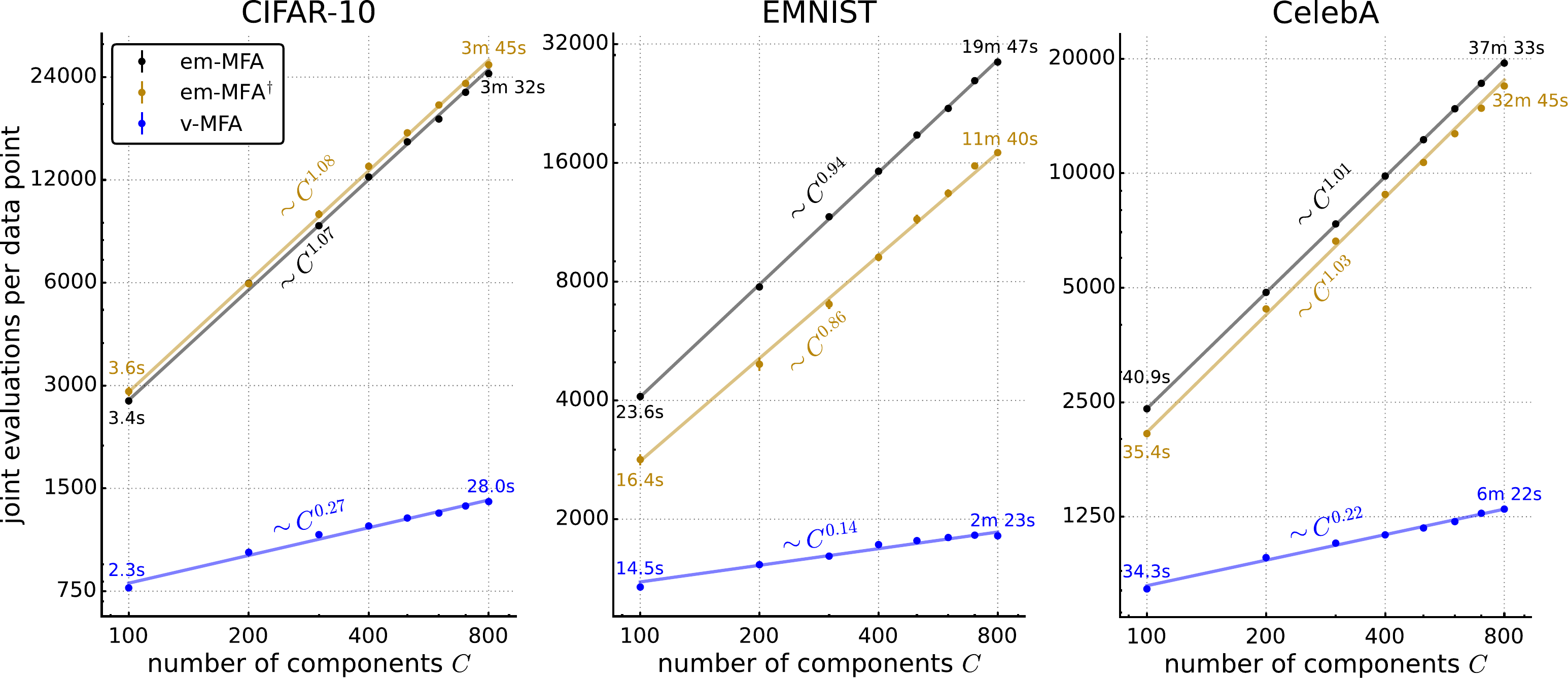}
  \vspace{0.8em} \\
  \begin{minipage}[c]{0.34\textwidth}
    \includegraphics[width=\textwidth]{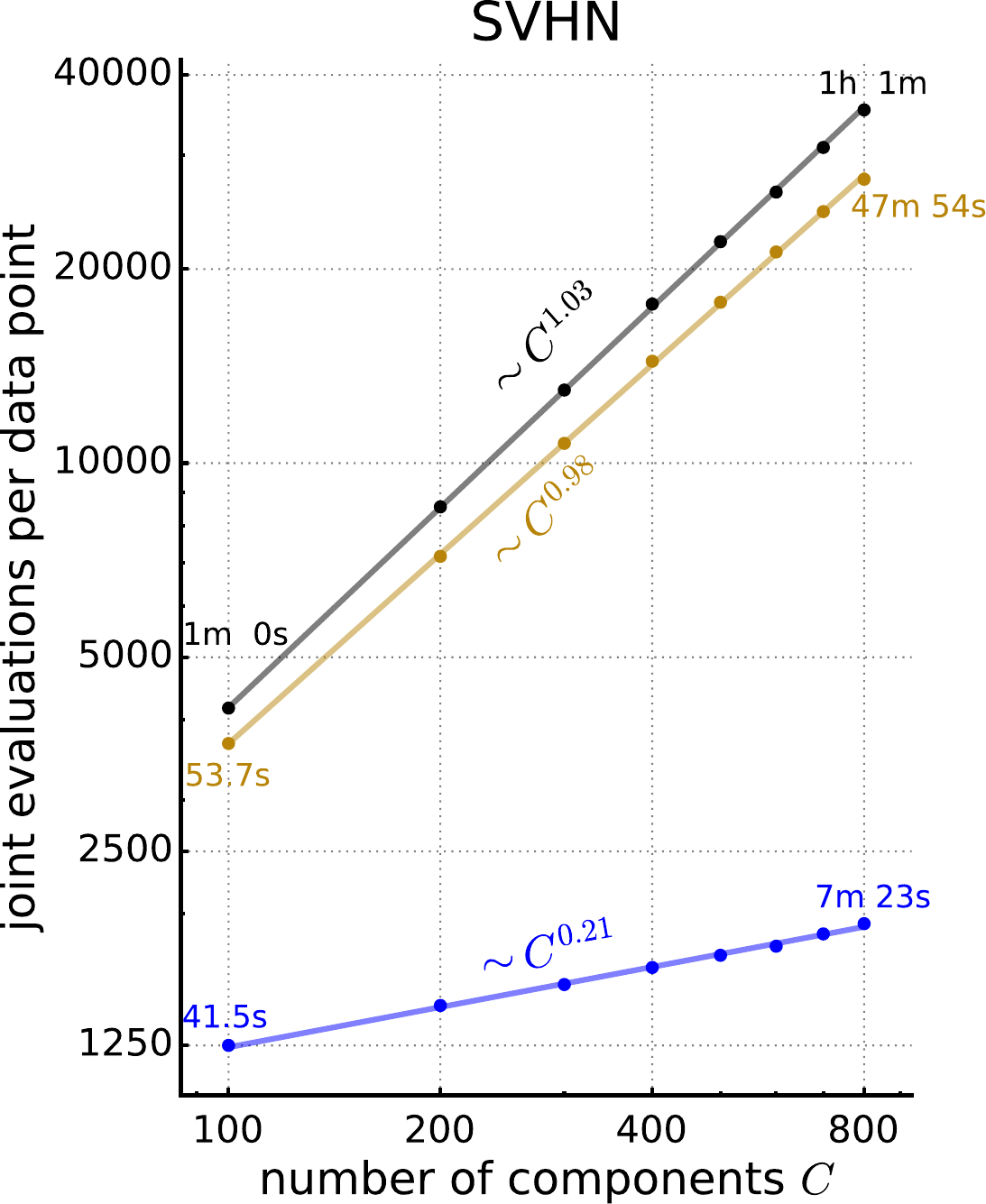}
  \end{minipage}
  \hspace{20pt}
  \begin{minipage}[c]{0.61\textwidth}
    \caption{
      Scaling behavior accounting for quality. Depicted is the number of joint evaluations per data point for increasingly large numbers of components~$C$. The values for \varMFA{} and \fullMFA{} are the same as in \cref{fig:scaling} in the main text. \fullMFAminus{} denotes the results for \fullMFA{} trained until it reaches approximately the same NLL$_\mathrm{test}$ as \varMFA{} (see text for details).
      Results are averaged over $40$ independent runs, with error bars indicating standard errors of the mean (SEM). Note that the SEM values are very small and barely visible in the plot.
      The legend in the first subfigure applies to all four subfigures. Lines represent regressions used to determine the scaling exponent of $C$ as in \cref{fig:scaling} in the main text, while annotations provide information on total training time.
    } \label{fig:scaling-same-quality}
  \end{minipage}
\end{figure}

As we only evaluate NLL$_\mathrm{test}$ values after a full iteration, values for \fullMFA{} and \varMFA{} never match precisely, however.
To address this, we proceeded as follows:
We trained the \fullMFA{} algorithm while monitoring the NLL$_\mathrm{test}$ after each iteration during training.
When the NLL$_\mathrm{test}$ of \fullMFA{} surpassed that of \varMFA{}, we stopped the algorithm.
We then used the model parameters from the \emph{previous} iteration of \fullMFA{} to ensure that the quality of \fullMFA{} was approximately the same as (but slightly worse than) the quality of \varMFA{}.
The results of this procedure are denoted as \fullMFAminus{} in \cref{fig:scaling-same-quality}, i.e., \fullMFAminus{} shows the speed-up achievable if \fullMFA{} is only required to reach
the final optimization quality of \varMFA{}. All other results shown in \cref{fig:scaling-same-quality} are taken from \cref{fig:scaling} in the main text.
For two of the four data sets, \fullMFAminus{} achieved a smaller scaling exponent than \fullMFA{}.
The number of iterations required to obtain the same NLL$_\mathrm{test}$ as \varMFA{} decreases slightly with increasingly large values of~$C$.
For both CIFAR-10 and CelebA, the scaling exponents of \fullMFAminus{} and \fullMFA{} are approximately the same.
On CIFAR-10, the NLL$_\mathrm{test}$ values of \varMFA{} are lower than those of \fullMFA{} for sufficiently large values of~$C$, as also observed in \cref{fig:quality} in the main text.
Consequently, the algorithm requires more iterations than previously to reach the desired quality.

Overall, the scaling exponent of \varMFA{} remains significantly lower than that of \fullMFAminus{} across all four data sets.
The control experiments thus show that the improved scaling behavior of \varMFA{} can be attributed to the sublinear complexity of the algorithm, while the impact of reduced quality contributes only a small fraction to the effect or is negligible.

\subsubsection{Ablation Study for Approximation Quality}
\label{appendix:v-mfa_quality}

The v-MFA algorithm is a variational approximation method that emphasizes computational efficiency, potentially at the expense of approximation quality (compare, e.g., \mbox{EMNIST} or SVHN in \cref{fig:quality}).
Its performance is influenced by: (1) the degree to which the assumption of truncated posteriors fits the underlying data, and (2) the effectiveness of the variational EM procedure in optimizing the variational parameters $\KKn$.

\begin{figure}[tb]
  \centering
  \includegraphics[width=0.8\textwidth]{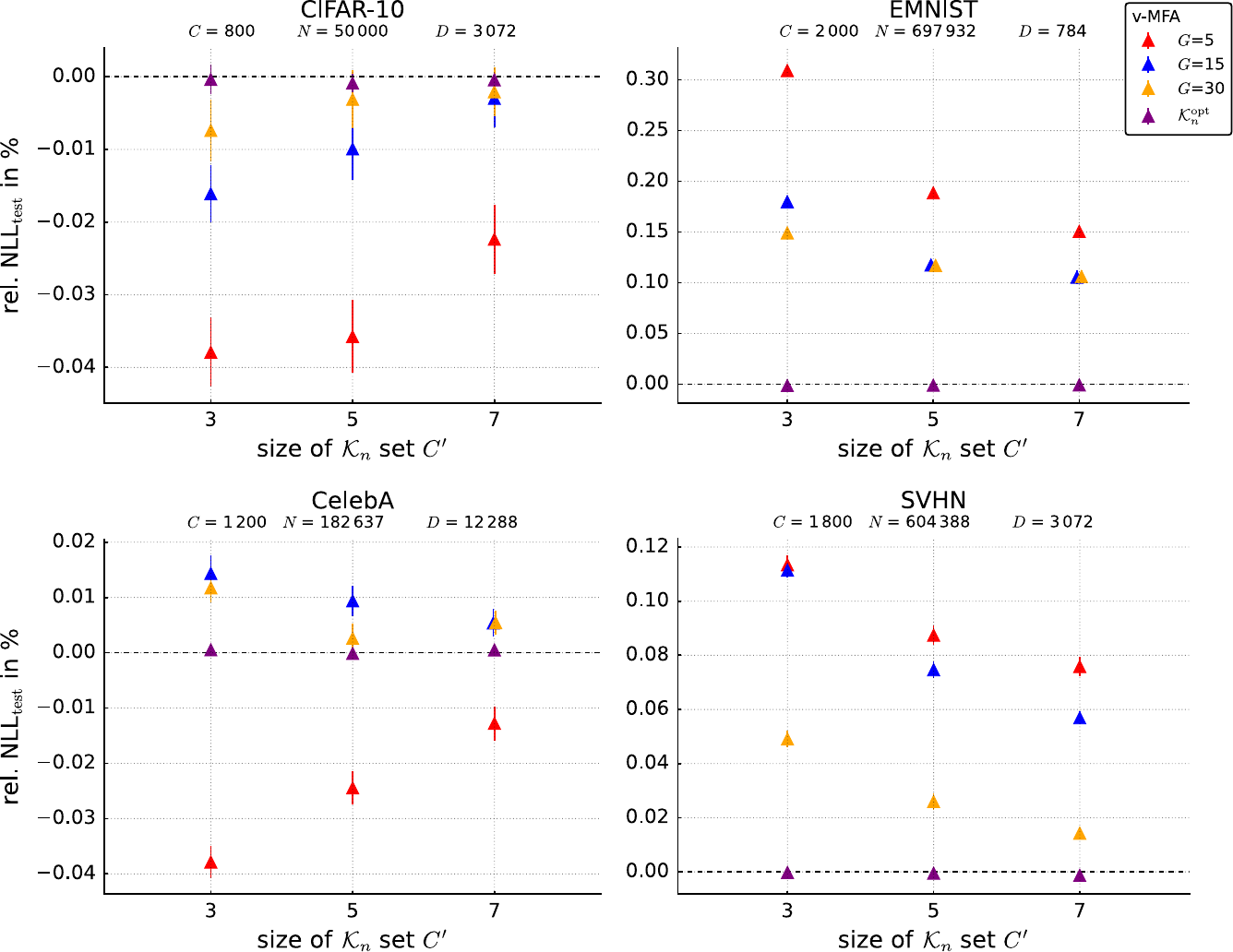}
  \caption{Effectiveness in terms of negative log-likelihood (NLL) of \varMFA{} for different configurations of $C'$ and $G$, as well as for a variant of \varMFA{} optimized using the optimal variational parameters $\KKnopt$ (as defined in \crefnobrackets{eq:Knopt}).
    Each of the four subfigures refers to experiments on one benchmark data set. The y-axes denote the relative NLL on the test set w.r.t.\ \fullMFA{} as baseline, given by \cref{eq:relnll}.
    Each measurement point is averaged over $40$ independent runs, with error bars indicating standard error of the mean (SEM) (note that the SEM values may be smaller than the symbol sizes).}
  \label{fig:var-quality}
\end{figure}

To disentangle the contributions of these two factors to the performance of \varMFA{}, we conducted an ablation study.
In addition to the experiments reported in \cref{sec:quality} in the main text, we evaluated \varMFA{} on the same data sets and the same values for $C'$. But instead of using the efficiently computed sets $\KKn$ of \varMFA{}, we used the optimal values $\KKnopt$, as defined in \cref{eq:Knopt}.
Finding $\KKnopt$ is computationally as demanding as an E-Step of \fullMFA{}. However, using $\KKnopt$ allows a quantification of the effect of truncation, without the influence of other approximations.
The residual difference in quality therefore reflects the accuracy of our approximation in estimating suitable variational parameters.
The results are presented in \cref{fig:var-quality}.

Considering the figure, it can be observed that when the \varMFA{} algorithm is evaluated using the optimal variational parameters $\KKnopt$, its performance is almost indistinguishable from that of \fullMFA{}, already for $C' = 3$.
Consequently, the effect of truncation is (almost) negligible, i.e., truncation itself does not account for the observed differences between \varMFA{} and \fullMFA{}.
\Cref{fig:var-quality} shows that the differences (at least for the studied data sets) are mainly due to the variational optimization procedure finding sets $\KKn$ different from the optimal sets $\KKnopt$.
This, in turn, leads to different model parameter optimizations due to the differences in the $\KKn$ sets.
Notably, the non-optimal $\KKn$ sets of \varMFA{} do not necessarily result in worse optimization results compared to \fullMFA{}. In some cases, e.g.,\ for CIFAR-10 and CelebA, \varMFA{} with $G = 5$ even results in slightly better
NLL$_\mathrm{test}$ values compared to the (much more computationally expensive) \fullMFA{} optimization.
This effect is likely due to alternative learning trajectories, e.g., avoiding overfitting or some local optima.

\subsubsection{Ablation Study for Different Initializations}
\label{appendix:abinit}
In the experiments reported in the main text, we consistently used the same default initialization strategy for the variational and model parameters (except in \cref{sec:largescale}).
This initialization strategy that leverages AFK-MC$^2$, is described in \cref{sec:algorithmic_realization}.
A crucial aspect of AFK-MC$^2$ is that it provides well-chosen seedings while maintaining low computational complexity.
To systematically analyze the impact of initialization on overall performance, we further investigate and compare AFK-MC$^2$ against alternative initialization strategies of varying sophistication. Specifically, we consider initialization using random noise, uniform sampling, \kmeans{}++, and the \kmeansfa{} algorithm, as detailed below.
For all here investigated initialization strategies, the sets $g_c$ are initialized using the default procedure and \warmup{} iterations are performed, as described in \cref{sec:algorithmic_realization}.

\myparagraph{Noise} The component means are initialized by sampling uniformly at random within the range of the observed data for each dimension, i.e.,
$\bmu_{cd} \sim  \mathcal{U}\{\mathrm{min}(x_{1:N,d}), \mathrm{max}(x_{1:N,d})\}$.
The factor loading matrices are initialized using the default strategy (cf. \cref{sec:algorithmic_realization}). The diagonal covariance matrices are set to the identity, i.e., $\bDc = \bIdentity$. The sets $\KK_{1:N}$ and $g_{1:C}$ are initialized by sampling indices uniformly from  $\{1,\ldots,C\}$ (ensuring uniqueness of the indices within each set).

\myparagraph{Uniform and \kmeans{}++} These initialization strategies closely follow that used in the main experiments (see \cref{sec:numerical_experiments}).
The factor loadings, diagonal matrix and $g_c$ set are initialized as described in \cref{sec:algorithmic_realization}.
The component means are initialized by sampling uniformly from the data points, i.e., $\bmu_{c} = \bxn \  \mathrm{with} \ n \sim  \mathcal{U}\{1, N\}$ (ensuring uniqueness of the indices), or by selecting data points via \kmeans{}++ seeding \citep{ArthurVassilvitskii2007}.
In both cases, the corresponding component index is inserted into the set $\KKn$.
Subsequently, $\mathcal{K}_{1:N}$ are populated with further indices by sampling uniformly from $\{1,\ldots,C\}$ (while ensuring uniqueness of the indices).
The Uniform strategy was also applied in \cref{sec:largescale}.

\myparagraph{\kmeansfa} This initialization strategy is the most sophisticated one. First, we run the \kmeansfa{} algorithm as described in \cref{appendix:algorithms}. The model parameters $\bTheta = \{\pi_{1:C}, \bA_{1:C}, \bmu_{1:C}, \bD_{1:C} \}$ are then initialized using the parameters obtained from the \kmeansfa{} algorithm.
For each set $\KKn$, we insert the index of the closest cluster (component) $c$ for the data point $\bxn$ as identified by the \kmeans{} algorithm (compare \crefnobrackets{eq:voronoi_cells}).
Subsequently, $\mathcal{K}_{1:N}$ are populated with additional indices  sampled uniformly from $\{1,\ldots,C\}$ (ensuring uniqueness of the indices).

Due to less informed initialization methods, we observed that certain components occasionally become empty during training, i.e., they contain no data points, causing their prior to become zero.
To maintain a fixed number of components (allowing for a fair comparison), we address this issue by splitting an existing component \citep[cf.][]{CaronEtAl2018, MonnierEtAl2020}.
Specifically, when a component $c$ becomes empty, we sample another component $c'$ from the prior distribution, such that larger components are more likely to be selected.
The parameters of the empty component $c$ are then re-initialized based on those of the sampled component $c'$, with additional small perturbations.
Finally, we insert $c$ into $g_{c'}$, which allows the $\KKn$ sets to be updated accordingly in the next E-Step.

\begin{figure}[tb]
  \centering
  \includegraphics[width=0.8\textwidth]{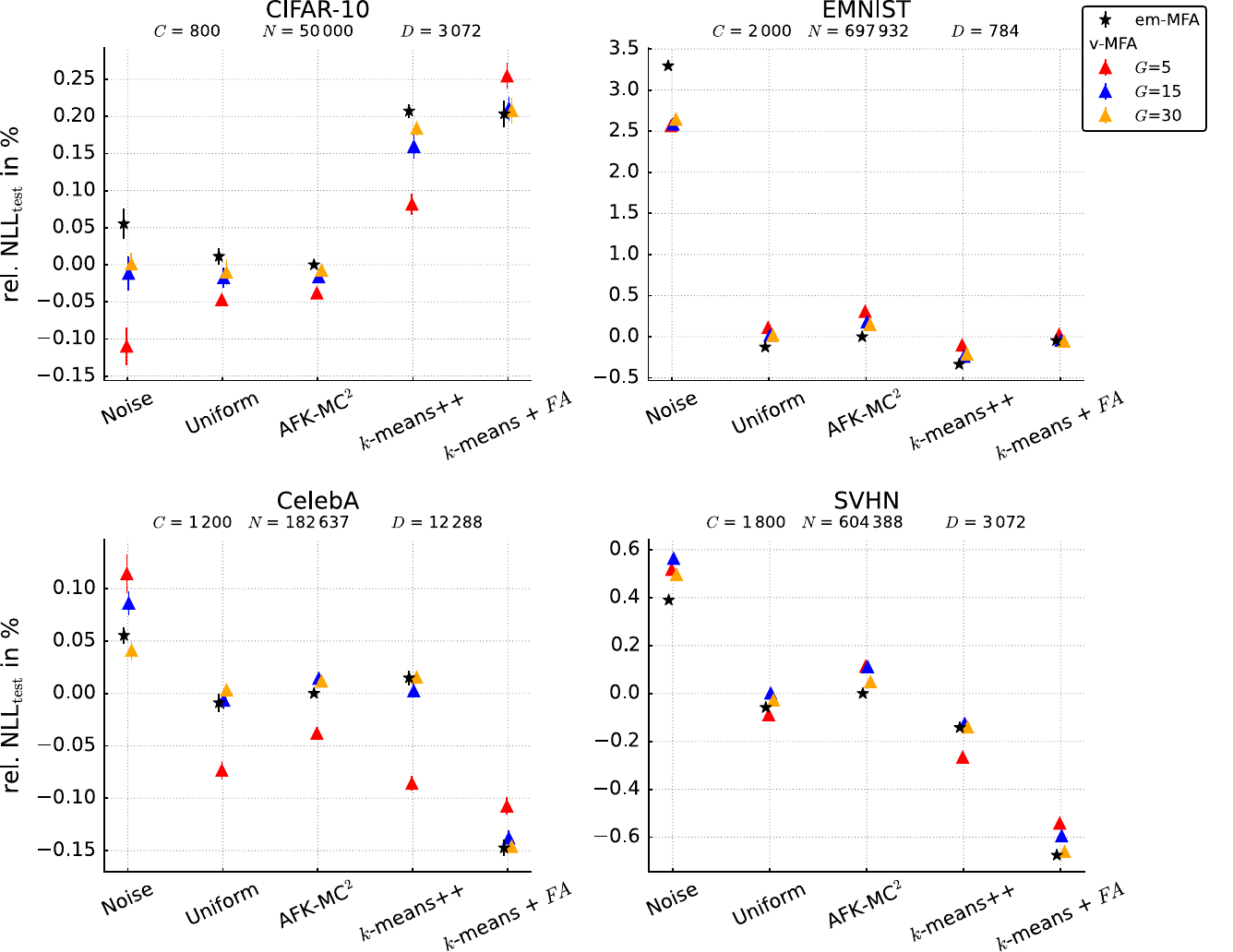}
  \caption{Comparison of different initialization strategies. Each of the four subfigures refers to experiments on one benchmark data set. The y-axes denote the relative NLL on the test set w.r.t. \fullMFA{} initialized which AFK-MC$^2$, given by \cref{eq:relnll}.
    Each measurement point is averaged over $40$ independent runs, with error bars indicating standard error of the mean (SEM) (note that the SEM values may be smaller than the symbol sizes).}
  \label{fig:initialization}
\end{figure}

The comparison of the different initialization strategies is presented in \cref{fig:initialization}.
The results indicate that more sophisticated methods, such as \kmeans{}++ or \kmeansfa{}, can yield improved NLL$_\mathrm{test}$ values for both \varMFA{} and \fullMFA{}, but not in all cases. However, this improvement comes at the expense of increased computational complexity. In particular, the \kmeansfa{} algorithm alone requires, for most settings, much more time than an entire \varMFA{} optimization with AFK-MC$^2$ seeding (see \cref{fig:quality} in the main text).

Using uniformly sampled data points as seeds for components (the `Uniform' method), as also employed in \cref{sec:largescale}, yields high final optimization quality that is similar to that obtained with AFK-MC$^2$ seeding, which may be taken as evidence for a substantial degree of robustness of the \varMFA{} algorithm w.r.t.\ initialization methods.
Only when initializing components with noise (the least informed method considered), performance decreases in almost all settings.
Finally, note that variations in NLL$_\mathrm{test}$ values across different initializations are comparable for both \varMFA{} and \fullMFA{}, i.e., the used variational optimization does not result in a noticeable increase of the sensitivity to initialization.

\subsection{Additional Information on the Quality Analysis}
\label{appendix:further_results}

Additional information and results about the experiments in \cref{sec:quality} in the main text are provided in \cref{tab:results}.
The table reports the relative NLL$_\mathrm{test}$, the average speed-up in terms of runtime and joint evaluations (or distance evaluations for \kmeansfa{}) compared to \fullMFA{} as well as the median number of iterations until convergence.
Additionally, for the \varMFA{} algorithm, the \warmup{} iterations are reported (note that the total number of iterations includes the \warmup{} iterations).

\begin{table}[tbp]
  \centering
  \caption{Additional information and results for the experiments in \cref{sec:quality} in the main text. Standard errors of the mean (SEM) for relative NLL$_\mathrm{test}$ were below $0.01\%$.
  }
  \label{tab:results}
  \resizebox{0.9\linewidth}{!}{
    \newcolumntype{?}[1]{}
    \begin{tabular}{c?{0.25pt}lrr?{0.25pt}r?{0.25pt}rr?{0.25pt}rr}
      \toprule
      \textbf{Data set} & \textbf{Algorithm} &      &     &                          & \multicolumn{2}{c?{0.25pt}}{Rel. speed-up} & \multicolumn{2}{c}{\#Iterations}                              \\
      \#Components      & Name               & $C'$ & $G$ & rel. NLL$_\mathrm{test}$ & \multicolumn{1}{l}{Time}                   & \multicolumn{1}{r?{0.25pt}}{Joints/Dist.} & \warmup{} & total \\
      \midrule
      CIFAR-10          & \fullMFA{}         & -    & -   & 0.0\%                    & 1.0$\times$                                & 1.0$\times$                               & -         & 31.0  \\$C=800$&\kmeansfa{}&-&-&1.64\%&4.4$\times$&0.5$\times$&-&77.6\\&\varMFA{}&3&5&-0.04\%&11.5$\times$&29.8$\times$&15.0&52.0\\&&3&15&-0.02\%&7.4$\times$&17.2$\times$&7.0&39.0\\&&3&30&-0.01\%&4.9$\times$&10.7$\times$&5.0&37.0\\&&5&5&-0.04\%&8.8$\times$&22.3$\times$&12.0&45.5\\&&5&15&-0.01\%&5.6$\times$&12.9$\times$&6.0&37.0\\&&5&30&-0.00\%&3.9$\times$&8.4$\times$&4.0&35.5\\&&7&5&-0.02\%&7.3$\times$&18.2$\times$&10.0&42.0\\&&7&15&-0.00\%&4.7$\times$&10.6$\times$&6.0&37.0\\&&7&30&-0.00\%&3.4$\times$&7.2$\times$&4.0&35.0\\
        \midrule
      EMNIST            & \fullMFA{}         & -    & -   & 0.0\%                    & 1.0$\times$                                & 1.0$\times$                               & -         & 29.0  \\$C=2000$&\kmeansfa{}&-&-&7.48\%&5.5$\times$&0.3$\times$&-&112.6\\&\varMFA{}&3&5&0.31\%&19.3$\times$&41.9$\times$&37.0&87.0\\&&3&15&0.18\%&14.9$\times$&30.1$\times$&16.0&48.0\\&&3&30&0.15\%&10.5$\times$&20.7$\times$&8.0&38.5\\&&5&5&0.19\%&14.1$\times$&30.6$\times$&31.0&75.0\\&&5&15&0.12\%&10.8$\times$&21.8$\times$&13.0&43.0\\&&5&30&0.12\%&7.7$\times$&15.0$\times$&8.0&37.0\\&&7&5&0.15\%&11.7$\times$&25.2$\times$&27.0&67.0\\&&7&15&0.11\%&8.7$\times$&17.4$\times$&11.0&40.0\\&&7&30&0.11\%&6.4$\times$&12.4$\times$&6.0&35.0\\
        \midrule
      CelebA            & \fullMFA{}         & -    & -   & 0.0\%                    & 1.0$\times$                                & 1.0$\times$                               & -         & 24.0  \\$C=1200$&\kmeansfa{}&-&-&1.03\%&3.3$\times$&0.2$\times$&-&108.6\\&\varMFA{}&3&5&-0.04\%&11.8$\times$&30.7$\times$&20.0&57.0\\&&3&15&0.01\%&8.1$\times$&20.3$\times$&8.0&35.0\\&&3&30&0.01\%&5.3$\times$&12.9$\times$&6.0&31.0\\&&5&5&-0.02\%&9.3$\times$&23.5$\times$&15.0&48.0\\&&5&15&0.01\%&6.0$\times$&14.7$\times$&6.0&32.0\\&&5&30&0.00\%&4.0$\times$&9.7$\times$&5.0&29.0\\&&7&5&-0.01\%&7.8$\times$&19.5$\times$&13.0&42.0\\&&7&15&0.01\%&4.7$\times$&11.6$\times$&6.0&30.5\\&&7&30&0.01\%&3.3$\times$&8.1$\times$&4.0&28.0\\
        \midrule
      SVHN              & \fullMFA{}         & -    & -   & 0.0\%                    & 1.0$\times$                                & 1.0$\times$                               & -         & 42.0  \\$C=1800$&\kmeansfa{}&-&-&3.19\%&7.0$\times$&0.4$\times$&-&110.3\\&\varMFA{}&3&5&0.11\%&25.6$\times$&61.9$\times$&17.0&78.0\\&&3&15&0.11\%&15.7$\times$&34.3$\times$&8.0&54.0\\&&3&30&0.05\%&10.0$\times$&21.2$\times$&6.0&50.0\\&&5&5&0.09\%&18.6$\times$&44.6$\times$&13.0&68.0\\&&5&15&0.07\%&10.7$\times$&23.3$\times$&7.0&51.0\\&&5&30&0.03\%&6.8$\times$&14.5$\times$&6.0&49.0\\&&7&5&0.08\%&14.9$\times$&35.4$\times$&11.0&63.0\\&&7&15&0.06\%&8.3$\times$&18.0$\times$&6.0&50.0\\&&7&30&0.01\%&5.5$\times$&11.6$\times$&5.0&47.0\\
      \bottomrule
    \end{tabular}
  }
\end{table}

\subsection{Additional Denoising Results}
\label{appendix:further_denoising_results}

\begin{figure}[tbp]
  \centering
  \includegraphics[width=\textwidth]{fig_Set12-noise50-grid_02.pdf}\\
  \includegraphics[width=\textwidth]{fig_Set12-noise50-grid_04.pdf}\\
  \includegraphics[width=\textwidth]{fig_Set12-noise50-grid_09.pdf}\\
  \includegraphics[width=\textwidth]{fig_Set12-noise50-grid_10.pdf}\\[0.25em]
  \includegraphics[width=\textwidth]{fig_BSD68-noise50-grid_11.pdf}\\
  \includegraphics[width=\textwidth]{fig_BSD68-noise50-grid_64.pdf}\\
  \includegraphics[width=\textwidth]{fig_BSD68-noise50-grid_14.pdf}\\
  \includegraphics[width=\textwidth]{fig_BSD68-noise50-grid_65.pdf}\\[0.25em]
  \includegraphics[width=\textwidth]{fig_Confocal-grid_mice.pdf}\\
  \includegraphics[width=\textwidth]{fig_Confocal-grid_fish.pdf}\\[0.25em]
  \includegraphics[width=\textwidth]{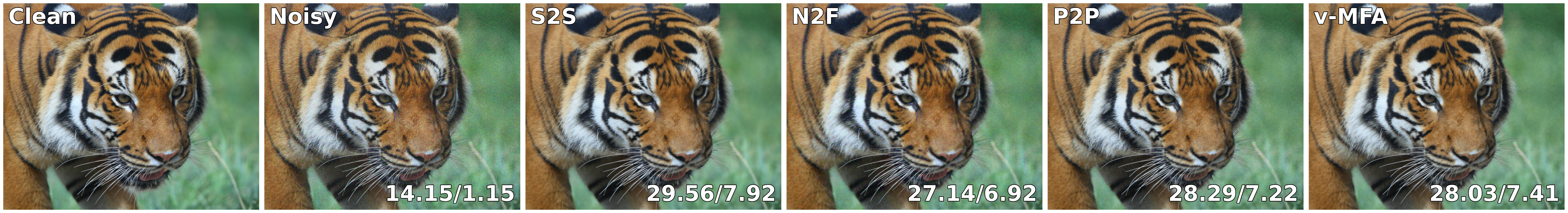}\\
  \vspace{-0.8em}
  \caption{
    Visual comparisons of several denoising algorithms applied to images from Set12, BSD68, Confocal and the `Tiger' image from Div2K (from top to bottom). All images were corrupted with additive Gaussian noise with a noise level of $\sigma = 50$, except for Confocal, which has inherent (natural) noise. PSNR and SSIM ($10\times$) values are displayed in the bottom right corner of the noisy and denoised images.
  }
  \label{fig:denoised_images_noise50}
\end{figure}

In the following, we show additional denoising results.
Visual comparisons of the denoised images produced by various blind zero-shot denoising algorithms are presented in \cref{fig:denoised_images_noise50}.
Further examples are available online.\furl{https://github.com/variational-sublinear-clustering/vammdx}

We also include other noise levels for Set12 and the `Tiger' image from Div2K and compare \varMFA{} against other variants of GMM-based denoising.
These include conventional EM training with \fullMFA{} instead of the variational optimization, as well as variants where the MFA model is replaced by a GMM with diagonal covariances (denoted em-DIAG and v-DIAG).
These results are presented in \cref{tab:appendix_denoising_results}.
Furthermore, we performed numerical experiments with different hyperplane dimensions $H$ of the MFA model (see \cref{tab:appendix_denosing-different-h}).

The results from \cref{tab:appendix_denoising_results} can be summarized as follows.
While \fullMFA{} shows in some settings marginal improvements in denoising performance over \varMFA{}, it comes at the cost of substantially increased runtimes (usually between one and two orders of magnitude longer runtimes). A similar behavior is observed for em-DIAG compared to v-DIAG but runtime differences are not as pronounced.
Regarding the comparison between the MFA model and diagonal GMMs, the MFA-based algorithms consistently show higher PSNR values (also compare \cref{tab:comparison}).

\begin{sidewaystable}[tb]
  \centering
  \caption{Accuracy and runtime of several zero-shot denoising algorithms, measured by PSNR, SSIM and average runtime per image in seconds. Errors denote standard deviation.
    The best result within each row among the blind zero-shot algorithms are marked bold, while overall best results (including the non-blind algorithms BM3D and DivN) are underlined.}
  \label{tab:appendix_denoising_results}
  \resizebox{\linewidth}{!}{
    \begin{tabular}{l l l r@{$\,\pm\,$}l r@{$\,\pm\,$}l r@{$\,\pm\,$}l r@{$\,\pm\,$}l r@{$\,\pm\,$}l r@{$\,\pm\,$}l r@{$\,\pm\,$}l r@{$\,\pm\,$}l r@{$\,\pm\,$}l r@{$\,\pm\,$}l r@{$\,\pm\,$}l r@{$\,\pm\,$}l}
      \toprule
      \multicolumn{3}{c}{}                         & \multicolumn{4}{c}{non-blind zero-shot} & \multicolumn{20}{c}{blind zero-shot}                                                                                                                                                                                                                                                                                                                                                                                                                                                                                                                                                                                                                                                                         \\
      \cmidrule(lr){4-7} \cmidrule(lr){8-27}
      Data set                                     & $\sigma$                                &                                      & \multicolumn{2}{c}{BM3D}                & \multicolumn{2}{c}{DivN}      & \multicolumn{2}{c}{N2S}        & \multicolumn{2}{c}{N2V}      & \multicolumn{2}{c}{DIP}       & \multicolumn{2}{c}{S2S}                              & \multicolumn{2}{c}{N2F}      & \multicolumn{2}{c}{P2P}      & \multicolumn{2}{c}{v-MFA}   & \multicolumn{2}{c}{em-MFA}    & \multicolumn{2}{c}{v-DIAG}                           & \multicolumn{2}{c}{em-DIAG}                                                                                                                                                                                                                       \\
      \midrule
      \multirow{12}{*}{Set12}
                                                   & \multirow{3}{*}{15}                     & PSNR                                 & \underline{$32.39$}                     & $1.13$                        & $28.35$                        & $2.05$                       & $28.45$                       & $2.86$                                               & $29.09$                      & $1.43$                       & $28.42$                     & $1.97$                        & $\boldsymbol{32.08}$                                 & $1.21$                      & $31.15$ & $1.34$ & $30.26$                         & $1.79$  & $30.59$                          & $1.73$  & $30.74$                         & $1.59$  & $26.91$                         & $2.07$ & $27.09$ & $1.98$ \\
                                                   &                                         & SSIM                                 & \underline{$8.96$}                      & $0.22$                        & $8.14$                         & $0.76$                       & $7.51$                        & $1.19$                                               & $8.27$                       & $0.27$                       & $8.00$                      & $0.78$                        & $\boldsymbol{8.89}$                                  & $0.26$                      & $8.72$  & $0.22$ & $8.66$                          & $0.25$  & $8.67$                           & $0.35$  & $8.67$                          & $0.33$  & $7.90$                          & $0.52$ & $7.94$  & $0.51$ \\
                                                   &                                         & Time                                 & $0.80$                                  & $0.33$                        & $298.69$                       & $34.37$                      & $1136.35$                     & $706.53$                                             & $296.43$                     & $24.83$                      & $38.06$                     & $1.26$                        & $3251.98$                                            & $2221.42$                   & $3.85$  & $2.22$ & $17.07$                         & $10.56$ & $1.50$                           & $1.09$  & $35.47$                         & $25.95$ & \underline{$\boldsymbol{0.46}$} & $0.23$ & $2.61$  & $1.55$ \\
      \cmidrule(lr){2-27}
                                                   & \multirow{3}{*}{25}                     & PSNR                                 & $29.99$                                 & $1.29$                        & $26.95$                        & $1.63$                       & $27.15$                       & $2.47$                                               & $27.35$                      & $1.55$                       & $27.16$                     & $1.65$                        & \underline{$\boldsymbol{30.03}$}                     & $1.13$                      & $29.06$ & $1.16$ & $28.80$                         & $1.40$  & $29.15$                          & $1.40$  & $29.19$                         & $1.30$  & $26.63$                         & $1.88$ & $26.73$ & $1.87$ \\
                                                   &                                         & SSIM                                 & \underline{$8.52$}                      & $0.30$                        & $7.80$                         & $0.52$                       & $7.37$                        & $1.28$                                               & $7.53$                       & $0.29$                       & $7.74$                      & $0.53$                        & $\boldsymbol{8.49}$                                  & $0.32$                      & $8.22$  & $0.30$ & $8.11$                          & $0.31$  & $8.25$                           & $0.33$  & $8.18$                          & $0.35$  & $7.78$                          & $0.50$ & $7.81$  & $0.48$ \\
                                                   &                                         & Time                                 & $0.81$                                  & $0.33$                        & $283.20$                       & $49.93$                      & $1139.62$                     & $708.82$                                             & $296.50$                     & $27.94$                      & $37.60$                     & $1.19$                        & $3262.46$                                            & $2218.07$                   & $3.32$  & $2.05$ & $16.94$                         & $10.93$ & $1.37$                           & $0.80$  & $36.06$                         & $23.60$ & \underline{$\boldsymbol{0.46}$} & $0.22$ & $2.07$  & $1.21$ \\
      \cmidrule(lr){2-27}
                                                   & \multirow{3}{*}{35}                     & PSNR                                 & $28.42$                                 & $1.31$                        & $25.37$                        & $2.76$                       & $25.08$                       & $2.87$                                               & $25.71$                      & $1.17$                       & $25.54$                     & $1.16$                        & \underline{$\boldsymbol{28.44}$}                     & $1.10$                      & $27.48$ & $1.08$ & $27.47$                         & $1.13$  & $27.81$                          & $1.27$  & $27.82$                         & $1.19$  & $26.16$                         & $1.76$ & $26.30$ & $1.70$ \\
                                                   &                                         & SSIM                                 & \underline{$8.15$}                      & $0.35$                        & $7.16$                         & $1.33$                       & $6.49$                        & $1.48$                                               & $6.71$                       & $0.56$                       & $7.05$                      & $0.52$                        & $\boldsymbol{8.09}$                                  & $0.33$                      & $7.76$  & $0.36$ & $7.49$                          & $0.34$  & $7.76$                           & $0.34$  & $7.62$                          & $0.39$  & $7.59$                          & $0.50$ & $7.62$  & $0.48$ \\
                                                   &                                         & Time                                 & $0.81$                                  & $0.33$                        & $245.19$                       & $66.98$                      & $1135.72$                     & $707.89$                                             & $299.21$                     & $26.24$                      & $37.75$                     & $1.47$                        & $3254.99$                                            & $2220.10$                   & $3.12$  & $2.29$ & $16.99$                         & $10.98$ & $1.38$                           & $0.90$  & $30.17$                         & $18.40$ & \underline{$\boldsymbol{0.47}$} & $0.18$ & $1.99$  & $1.09$ \\
      \cmidrule(lr){2-27}
                                                   & \multirow{3}{*}{50}                     & PSNR                                 & \underline{$26.75$}                     & $1.27$                        & $23.81$                        & $3.23$                       & $23.66$                       & $3.15$                                               & $24.02$                      & $1.57$                       & $22.75$                     & $0.91$                        & $\boldsymbol{26.54}$                                 & $1.07$                      & $25.89$ & $1.07$ & $25.72$                         & $0.90$  & $26.12$                          & $1.19$  & $26.05$                         & $1.14$  & $25.32$                         & $1.56$ & $25.44$ & $1.53$ \\
                                                   &                                         & SSIM                                 & \underline{$7.68$}                      & $0.41$                        & $6.66$                         & $1.49$                       & $6.17$                        & $1.49$                                               & $5.86$                       & $0.67$                       & $5.48$                      & $0.55$                        & $\boldsymbol{7.42}$                                  & $0.37$                      & $7.24$  & $0.42$ & $6.50$                          & $0.40$  & $7.03$                           & $0.39$  & $6.77$                          & $0.50$  & $7.25$                          & $0.51$ & $7.27$  & $0.47$ \\
                                                   &                                         & Time                                 & $0.83$                                  & $0.35$                        & $197.02$                       & $69.79$                      & $1140.80$                     & $712.30$                                             & $294.17$                     & $27.80$                      & $37.91$                     & $1.39$                        & $3238.05$                                            & $2197.54$                   & $3.36$  & $2.86$ & $17.08$                         & $10.95$ & $1.23$                           & $0.73$  & $34.93$                         & $23.30$ & \underline{$\boldsymbol{0.44}$} & $0.19$ & $1.73$  & $0.92$ \\
      \midrule
      \multirow{6}{*}{BSD68}                       & \multirow{3}{*}{25}                     & PSNR                                 & \underline{$28.61$}                     & $2.52$                        & $25.11$                        & $3.25$                       & $26.72$                       & $3.23$                                               & $26.30$                      & $2.67$                       & $25.96$                     & $2.84$                        & $\boldsymbol{28.46}$                                 & $2.85$                      & $28.11$ & $2.44$ & $27.26$                         & $2.89$  & $27.22$                          & $3.33$  & $27.29$                         & $3.23$  & $25.36$                         & $3.46$ & $25.48$ & $3.41$ \\
                                                   &                                         & SSIM                                 & \underline{$8.04$}                      & $0.66$                        & $6.60$                         & $1.67$                       & $7.38$                        & $0.92$                                               & $7.06$                       & $0.70$                       & $7.13$                      & $0.79$                        & $\boldsymbol{7.97}$                                  & $0.72$                      & $7.87$  & $0.61$ & $7.50$                          & $0.59$  & $7.54$                           & $0.81$  & $7.53$                          & $0.72$  & $6.89$                          & $1.05$ & $6.94$  & $1.03$ \\
                                                   &                                         & Time                                 & $0.79$                                  & $0.02$                        & $224.38$                       & $94.58$                      & $1198.04$                     & $2.60$                                               & $920.73$                     & $535.47$                     & $39.68$                     & $0.48$                        & $3230.29$                                            & $56.55$                     & $3.67$  & $0.90$ & $20.97$                         & $0.19$  & $1.42$                           & $0.32$  & $27.59$                         & $2.09$  & \underline{$\boldsymbol{0.44}$} & $0.05$ & $2.16$  & $0.25$ \\
      \cmidrule(lr){2-27}
                                                   & \multirow{3}{*}{50}                     & PSNR                                 & $25.67$                                 & $2.60$                        & $20.72$                        & $4.89$                       & $24.49$                       & $2.26$                                               & $23.22$                      & $2.78$                       & $22.28$                     & $1.62$                        & \underline{$\boldsymbol{25.70}$}                     & $2.48$                      & $25.31$ & $2.35$ & $24.84$                         & $2.10$  & $25.30$                          & $2.66$  & $25.22$                         & $2.42$  & $24.35$                         & $3.37$ & $24.43$ & $3.35$ \\
                                                   &                                         & SSIM                                 & \underline{$6.89$}                      & $0.91$                        & $5.14$                         & $2.02$                       & $6.39$                        & $0.80$                                               & $5.16$                       & $0.77$                       & $5.16$                      & $0.98$                        & $\boldsymbol{6.87}$                                  & $0.86$                      & $6.72$  & $0.80$ & $5.91$                          & $0.51$  & $6.51$                           & $0.86$  & $6.31$                          & $0.67$  & $6.32$                          & $1.18$ & $6.37$  & $1.15$ \\
                                                   &                                         & Time                                 & $0.81$                                  & $0.03$                        & $157.14$                       & $96.25$                      & $1185.39$                     & $2.74$                                               & $935.96$                     & $552.42$                     & $39.99$                     & $0.83$                        & $3222.79$                                            & $61.80$                     & $3.36$  & $0.97$ & $21.14$                         & $0.33$  & $1.40$                           & $0.24$  & $25.95$                         & $1.14$  & \underline{$\boldsymbol{0.46}$} & $0.07$ & $1.97$  & $0.28$ \\
      \midrule
      \multirow{3}{*}{Confocal}                    & \multirow{3}{*}{--}                     & PSNR                                 & \multicolumn{2}{c}{--}                  & \multicolumn{2}{c}{--}        & $36.20$                        & $3.88$                       & $35.70$                       & $3.53$                                               & $34.53$                      & $3.89$                       & $36.51$                     & $3.39$                        & $36.60$                                              & $3.84$                      & $34.02$ & $2.33$ & $36.83$                         & $3.54$  & \underline{$\boldsymbol{36.84}$} & $3.52$  & $35.64$                         & $3.52$  & $35.65$                         & $3.52$                    \\
                                                   &                                         & SSIM                                 & \multicolumn{2}{c}{--}                  & \multicolumn{2}{c}{--}        & $9.26$                         & $0.38$                       & $9.18$                        & $0.45$                                               & $8.94$                       & $0.64$                       & $9.28$                      & $0.38$                        & $9.33$                                               & $0.39$                      & $9.04$  & $0.48$ & \underline{$\boldsymbol{9.34}$} & $0.37$  & \underline{$\boldsymbol{9.34}$}  & $0.37$  & $9.14$                          & $0.51$  & $9.14$                          & $0.51$                    \\
                                                   &                                         & Time                                 & \multicolumn{2}{c}{--}                  & \multicolumn{2}{c}{--}        & $1964.20$                      & $1.22$                       & $203.80$                      & $20.56$                                              & $41.93$                      & $3.54$                       & $5884.49$                   & $140.50$                      & $8.42$                                               & $0.99$                      & $30.71$ & $2.24$ & $4.43$                          & $1.63$  & $89.62$                          & $37.24$ & \underline{$\boldsymbol{1.35}$} & $0.66$  & $8.94$                          & $3.86$                    \\
      \midrule
      \multirow{9}{*}{\parbox{1cm}{Div2K `Tiger'}} & \multirow{3}{*}{25}                     & PSNR                                 & \multicolumn{2}{c}{\underline{$32.16$}} & \multicolumn{2}{c}{$29.00$}   & \multicolumn{2}{c}{$30.02$}    & \multicolumn{2}{c}{$29.59$}  & \multicolumn{2}{c}{$27.07$}   & \multicolumn{2}{c}{$\boldsymbol{31.95}$}             & \multicolumn{2}{c}{$29.66$}  & \multicolumn{2}{c}{$30.67$}  & \multicolumn{2}{c}{$29.19$} & \multicolumn{2}{c}{$29.28$}   & \multicolumn{2}{c}{$26.56$}                          & \multicolumn{2}{c}{$26.60$}                                                                                                                                                                                                                       \\
                                                   &                                         & SSIM                                 & \multicolumn{2}{c}{\underline{$8.60$}}  & \multicolumn{2}{c}{$7.57$}    & \multicolumn{2}{c}{$7.95$}     & \multicolumn{2}{c}{$7.63$}   & \multicolumn{2}{c}{$6.96$}    & \multicolumn{2}{c}{$\boldsymbol{8.56}$}              & \multicolumn{2}{c}{$7.79$}   & \multicolumn{2}{c}{$8.16$}   & \multicolumn{2}{c}{$7.83$}  & \multicolumn{2}{c}{$7.86$}    & \multicolumn{2}{c}{$6.93$}                           & \multicolumn{2}{c}{$6.95$}                                                                                                                                                                                                                        \\
                                                   &                                         & Time                                 & \multicolumn{2}{c}{$29.46$}             & \multicolumn{2}{c}{$2904.16$} & \multicolumn{2}{c}{$72962.49$} & \multicolumn{2}{c}{$200.57$} & \multicolumn{2}{c}{$1652.69$} & \multicolumn{2}{c}{$89821.74$}                       & \multicolumn{2}{c}{$217.89$} & \multicolumn{2}{c}{$554.87$} & \multicolumn{2}{c}{$47.18$} & \multicolumn{2}{c}{$1513.44$} & \multicolumn{2}{c}{\underline{$\boldsymbol{13.39}$}} & \multicolumn{2}{c}{$86.46$}                                                                                                                                                                                                                       \\
      \cmidrule(lr){2-27}
                                                   & \multirow{3}{*}{50}                     & PSNR                                 & \multicolumn{2}{c}{$29.23$}             & \multicolumn{2}{c}{$26.70$}   & \multicolumn{2}{c}{$27.04$}    & \multicolumn{2}{c}{$26.35$}  & \multicolumn{2}{c}{$25.93$}   & \multicolumn{2}{c}{\underline{$\boldsymbol{29.56}$}} & \multicolumn{2}{c}{$27.14$}  & \multicolumn{2}{c}{$28.29$}  & \multicolumn{2}{c}{$28.03$} & \multicolumn{2}{c}{$28.06$}   & \multicolumn{2}{c}{$26.35$}                          & \multicolumn{2}{c}{$26.36$}                                                                                                                                                                                                                       \\
                                                   &                                         & SSIM                                 & \multicolumn{2}{c}{$7.81$}              & \multicolumn{2}{c}{$6.64$}    & \multicolumn{2}{c}{$6.83$}     & \multicolumn{2}{c}{$5.77$}   & \multicolumn{2}{c}{$6.60$}    & \multicolumn{2}{c}{\underline{$\boldsymbol{7.92}$}}  & \multicolumn{2}{c}{$6.92$}   & \multicolumn{2}{c}{$7.22$}   & \multicolumn{2}{c}{$7.41$}  & \multicolumn{2}{c}{$7.42$}    & \multicolumn{2}{c}{$6.85$}                           & \multicolumn{2}{c}{$6.88$}                                                                                                                                                                                                                        \\
                                                   &                                         & Time                                 & \multicolumn{2}{c}{$29.28$}             & \multicolumn{2}{c}{$2773.43$} & \multicolumn{2}{c}{$73193.76$} & \multicolumn{2}{c}{$201.57$} & \multicolumn{2}{c}{$257.96$}  & \multicolumn{2}{c}{$90042.40$}                       & \multicolumn{2}{c}{$217.11$} & \multicolumn{2}{c}{$502.03$} & \multicolumn{2}{c}{$38.45$} & \multicolumn{2}{c}{$1334.39$} & \multicolumn{2}{c}{\underline{$\boldsymbol{12.53}$}} & \multicolumn{2}{c}{$64.64$}                                                                                                                                                                                                                       \\
      \cmidrule(lr){2-27}
                                                   & \multirow{3}{*}{100}                    & PSNR                                 & \multicolumn{2}{c}{$26.43$}             & \multicolumn{2}{c}{$24.81$}   & \multicolumn{2}{c}{$24.51$}    & \multicolumn{2}{c}{$22.55$}  & \multicolumn{2}{c}{$25.14$}   & \multicolumn{2}{c}{\underline{$\boldsymbol{27.21}$}} & \multicolumn{2}{c}{$23.82$}  & \multicolumn{2}{c}{$25.31$}  & \multicolumn{2}{c}{$26.07$} & \multicolumn{2}{c}{$26.03$}   & \multicolumn{2}{c}{$25.43$}                          & \multicolumn{2}{c}{$25.40$}                                                                                                                                                                                                                       \\
                                                   &                                         & SSIM                                 & \multicolumn{2}{c}{$6.83$}              & \multicolumn{2}{c}{$5.92$}    & \multicolumn{2}{c}{$5.78$}     & \multicolumn{2}{c}{$3.37$}   & \multicolumn{2}{c}{$6.44$}    & \multicolumn{2}{c}{\underline{$\boldsymbol{7.14}$}}  & \multicolumn{2}{c}{$5.69$}   & \multicolumn{2}{c}{$5.57$}   & \multicolumn{2}{c}{$6.71$}  & \multicolumn{2}{c}{$6.72$}    & \multicolumn{2}{c}{$6.55$}                           & \multicolumn{2}{c}{$6.59$}                                                                                                                                                                                                                        \\
                                                   &                                         & Time                                 & \multicolumn{2}{c}{$30.07$}             & \multicolumn{2}{c}{$2235.74$} & \multicolumn{2}{c}{$73188.33$} & \multicolumn{2}{c}{$202.24$} & \multicolumn{2}{c}{$255.92$}  & \multicolumn{2}{c}{$89817.23$}                       & \multicolumn{2}{c}{$102.39$} & \multicolumn{2}{c}{$505.76$} & \multicolumn{2}{c}{$38.48$} & \multicolumn{2}{c}{$1187.01$} & \multicolumn{2}{c}{\underline{$\boldsymbol{12.60}$}} & \multicolumn{2}{c}{$59.22$}                                                                                                                                                                                                                       \\
      \bottomrule
    \end{tabular}
  }
\end{sidewaystable}

For the results in \cref{sec:denoising} and \cref{tab:appendix_denoising_results}, we employed a fixed set of hyperparameters across all data sets and noise levels. However, these hyperparameters are expected to influence denoising performance.
For the number of components, this is already evident from \cref{fig:denoising_scaling-2}, where we increased the number of components up to $C = \num{10000}$.
A low hyperplane dimensionality (we chose $H=5$ for the experiments in \cref{sec:denoising}) was observed to result in a favorable trade-off between optimization qualities and runtimes.
Here we further investigate the effect of the hyperplane dimensionality $H$. Concretely, we repeated the experiment from \cref{sec:denoising} for \varMFA{}, varying the hyperplane dimension over $H \in \{3, 5, 10, 20\}$. Increasing $H$ allows each component to model more intra-component correlations but model sizes and computational costs increase with higher $H$.

As can be observed by inspecting the results of \cref{tab:appendix_denosing-different-h}, increasing $H$ continuously increases performance for the high-resolution image of the Div2K data set. However, performance does not necessarily increase, and it decreases for the Set12 data set, for instance. This observation can be explained by limited data available for estimating the model parameters. For Set12, there is much less information available than for the Div2K image. If $H$ is increased, more model parameters have to be estimated, and data variations due to noise are being modeled with available parameters (a form of overfitting).
Finally, we observe that lower noise levels allow for higher $H$, which can also be explained by more limited amounts of information being available for higher noise levels.

\begin{table}[tb]
  \caption{The effect of different values of $H$ on \varMFA{} for denoising. Accuracy and runtime are measured by PSNR, SSIM and average runtime per image in seconds. Errors denote standard deviation.
    The best results within each row are marked bold.
  }
  \label{tab:appendix_denosing-different-h}
  \centering
  \begin{tabular}{l l l r@{$\,\pm\,$}l r@{$\,\pm\,$}l r@{$\,\pm\,$}l r@{$\,\pm\,$}l}
    \toprule
    \multicolumn{3}{c}{}                         & \multicolumn{8}{c}{v-MFA}                                                                                                                                                                                                             \\
    \cmidrule(lr){4-11}
    Data set                                     & $\sigma$                  &      & \multicolumn{2}{c}{$H=3$}                & \multicolumn{2}{c}{$H=5$}   & \multicolumn{2}{c}{$H=10$}  & \multicolumn{2}{c}{$H=20$}                                                                  \\
    \midrule
    \multirow{6}{*}{Set12}                       & \multirow{3}{*}{25}       & PSNR & $28.71$                                  & $1.64$                      & $\boldsymbol{29.15}$        & $1.40$                                   & $29.11$              & $1.14$ & $27.61$ & $1.81$ \\
                                                 &                           & SSIM & $8.22$                                   & $0.38$                      & $\boldsymbol{8.25}$         & $0.33$                                   & $8.05$               & $0.32$ & $7.29$  & $0.68$ \\
                                                 &                           & Time & $\boldsymbol{1.20}$                      & $0.81$                      & $1.28$                      & $0.91$                                   & $1.47$               & $0.93$ & $2.68$  & $1.49$ \\
    \cmidrule(lr){2-11}
                                                 & \multirow{3}{*}{50}       & PSNR & $\boldsymbol{26.19}$                     & $1.29$                      & $26.12$                     & $1.19$                                   & $25.26$              & $1.44$ & $23.16$ & $2.30$ \\
                                                 &                           & SSIM & $\boldsymbol{7.26}$                      & $0.38$                      & $7.03$                      & $0.39$                                   & $6.30$               & $0.68$ & $5.08$  & $1.07$ \\
                                                 &                           & Time & $\boldsymbol{1.19}$                      & $0.67$                      & $1.43$                      & $0.92$                                   & $1.62$               & $0.91$ & $3.04$  & $1.54$ \\
    \midrule
    \multirow{6}{*}{BSD68}                       & \multirow{3}{*}{25}       & PSNR & $26.78$                                  & $3.40$                      & $27.25$                     & $3.27$                                   & $\boldsymbol{27.77}$ & $2.86$ & $27.64$ & $2.01$ \\
                                                 &                           & SSIM & $7.30$                                   & $1.06$                      & $7.47$                      & $0.94$                                   & $\boldsymbol{7.62}$  & $0.70$ & $7.43$  & $0.48$ \\
                                                 &                           & Time & $\boldsymbol{1.22}$                      & $0.26$                      & $1.36$                      & $0.25$                                   & $1.63$               & $0.34$ & $2.77$  & $0.47$ \\
    \cmidrule(lr){2-11}
                                                 & \multirow{3}{*}{50}       & PSNR & $25.18$                                  & $2.82$                      & $\boldsymbol{25.30}$        & $2.66$                                   & $25.12$              & $2.19$ & $24.02$ & $1.46$ \\
                                                 &                           & SSIM & $\boldsymbol{6.54}$                      & $1.00$                      & $6.51$                      & $0.86$                                   & $6.20$               & $0.60$ & $5.34$  & $0.57$ \\
                                                 &                           & Time & $\boldsymbol{1.28}$                      & $0.22$                      & $1.48$                      & $0.31$                                   & $1.80$               & $0.34$ & $3.20$  & $0.56$ \\
    \midrule
    \multirow{3}{*}{Confocal}                    & \multirow{3}{*}{--}       & PSNR & $36.74$                                  & $3.60$                      & $\boldsymbol{36.83}$        & $3.54$                                   & $36.28$              & $3.27$ & $34.87$ & $3.13$ \\
                                                 &                           & SSIM & $9.31$                                   & $0.40$                      & $\boldsymbol{9.34}$         & $0.37$                                   & $\boldsymbol{9.34}$  & $0.36$ & $9.22$  & $0.40$ \\
                                                 &                           & Time & $\boldsymbol{4.61}$                      & $2.18$                      & $5.39$                      & $3.10$                                   & $5.66$               & $2.35$ & $9.59$  & $3.98$ \\
    \midrule
    \multirow{6}{*}{\parbox{1cm}{Div2K `Tiger'}} & \multirow{3}{*}{25}       & PSNR & \multicolumn{2}{c}{$28.42$}              & \multicolumn{2}{c}{$29.19$} & \multicolumn{2}{c}{$30.47$} & \multicolumn{2}{c}{$\boldsymbol{31.60}$}                                                    \\
                                                 &                           & SSIM & \multicolumn{2}{c}{$7.57$}               & \multicolumn{2}{c}{$7.83$}  & \multicolumn{2}{c}{$8.21$}  & \multicolumn{2}{c}{$\boldsymbol{8.49}$}                                                     \\
                                                 &                           & Time & \multicolumn{2}{c}{$\boldsymbol{43.73}$} & \multicolumn{2}{c}{$46.48$} & \multicolumn{2}{c}{$60.49$} & \multicolumn{2}{c}{$89.29$}                                                                 \\
    \cmidrule(lr){2-11}
                                                 & \multirow{3}{*}{50}       & PSNR & \multicolumn{2}{c}{$27.55$}              & \multicolumn{2}{c}{$28.03$} & \multicolumn{2}{c}{$28.65$} & \multicolumn{2}{c}{$\boldsymbol{29.13}$}                                                    \\
                                                 &                           & SSIM & \multicolumn{2}{c}{$7.24$}               & \multicolumn{2}{c}{$7.41$}  & \multicolumn{2}{c}{$7.61$}  & \multicolumn{2}{c}{$\boldsymbol{7.76}$}                                                     \\
                                                 &                           & Time & \multicolumn{2}{c}{$\boldsymbol{34.97}$} & \multicolumn{2}{c}{$38.65$} & \multicolumn{2}{c}{$47.82$} & \multicolumn{2}{c}{$74.24$}                                                                 \\
    \bottomrule
  \end{tabular}
\end{table}

\bibliographystyle{abbrvnat}
\bibliography{bibliography}

\end{document}